%% file: main.tex
\documentclass[conference]{IEEEtran}

\ifCLASSINFOpdf
\else
\fi

\usepackage{caption}
\usepackage{subcaption}
\usepackage{booktabs}
\usepackage{multirow}
\usepackage{amsfonts}
\usepackage{amsthm}
\usepackage{amsmath}
\usepackage{xcolor}
\usepackage{hyperref}
\usepackage{makecell}
\usepackage{float}
\usepackage[numbers]{natbib}
\usepackage[normalem]{ulem}
\usepackage{url}
\usepackage{hyperref}
\usepackage{listings}

\DeclareUnicodeCharacter{25CF}{$\bullet$}
\DeclareUnicodeCharacter{251C}{\mbox{\kern.23em
  \vrule height2.2exdepth1exwidth.4pt\vrule height2.2ptdepth-1.8ptwidth.23em}}
\DeclareUnicodeCharacter{2500}{\mbox{\vrule height2.2ptdepth-1.8ptwidth.5em}}
\DeclareUnicodeCharacter{2514}{\mbox{\kern.23em
  \vrule height2.2exdepth-1.8ptwidth.4pt\vrule height2.2ptdepth-1.8ptwidth.23em}}
\useunder{\uline}{\ul}{}
\usepackage[final]{changes}
\usepackage[available,functional,reproduced]{ndssbadges}
\newtheorem{definition}{Definition}
\newtheorem{theorem}{Theorem}
\newcommand{\method}[0]{\texttt{PBP}}%
\usepackage[linesnumbered,ruled,vlined]{algorithm2e}
\SetAlFnt{\footnotesize}
\SetKw{KwDownTo}{downto}
\SetKw{KwTo}{to}
\SetKw{KwTrue}{true}
\SetKw{KwFalse}{false}
\SetKwInOut{Input}{Input}
\SetKwInOut{Output}{Output}
\SetKw{KwAnd}{and}
\SetKw{KwBreak}{break}

\makeatletter
\newtheorem{repeatthm@}{Theorem}
\newenvironment{repeatthm}[1]{%
    \def\therepeatthm@{\ref{#1}}
    \repeatthm@
}
{\endrepeatthm@}
\makeatother

\begin{document}
\pagenumbering{arabic}
\title{\textbf{PBP}: \underline{P}ost-training \underline{B}ackdoor \underline{P}urification for Malware Classifiers}

\pagestyle{plain}
\author{\IEEEauthorblockN{Dung Thuy Nguyen, Ngoc N. Tran, Taylor T. Johnson, Kevin Leach}
	\IEEEauthorblockA{Vanderbilt University,
            Nashville, TN, USA\\
		\{dung.t.nguyen, ngoc.n.tran, taylor.johnson, kevin.leach\}@vanderbilt.edu}
        }

\IEEEoverridecommandlockouts
\makeatletter\def\@IEEEpubidpullup{6.5\baselineskip}\makeatother
\IEEEpubid{\parbox{\columnwidth}{
		Network and Distributed System Security (NDSS) Symposium 2025\\
		24-28 February 2025, San Diego, CA, USA\\
		ISBN 979-8-9894372-8-3\\
		https://dx.doi.org/10.14722/ndss.2025.240603\\
		www.ndss-symposium.org
}
\hspace{\columnsep}\makebox[\columnwidth]{}}

\newcommand{\smbf}[1]{\noindent\textbf{#1} }
\newcommand{\best}[1]{\textcolor{black}{\textbf{\underline{#1}}}}
\newcommand{\second}[1]{\textcolor{black}{\underline{#1}}}
\newcommand{\ngoc}[1]{\textcolor{red}{NGOC: #1}}
\newcommand{\judy}[1]{\textcolor{purple}{Judy: #1}}
\newcommand{\fixed}[1]{\textcolor{black}{#1}}

\makeatletter
\AddToHook{cmd/added/before}{\def\Changes@AuthorColor{blue}}
\AddToHook{cmd/deleted/before}{\def\Changes@AuthorColor{red}}
\AddToHook{cmd/replaced/before}{\def\Changes@AuthorColor{blue}}
\AddToHook{cmd/replaced/after}{\def\Changes@AuthorColor{blue}}
\makeatother

\maketitle

\begin{abstract}
In recent years, the rise of machine learning (ML) in cybersecurity has brought new challenges, including the increasing threat of backdoor poisoning attacks on ML malware classifiers. These attacks aim to manipulate model behavior when provided with a particular input trigger. For instance, adversaries could inject malicious samples into public malware repositories, contaminating the training data and potentially misclassifying malware by the ML model. Current countermeasures predominantly focus on detecting poisoned samples by leveraging disagreements within the outputs of a diverse set of ensemble models on training data points.
\replaced{However, these methods are not suitable for scenarios where Machine Learning-as-a-Service (MLaaS) is used or when users aim to remove backdoors from a model after it has been trained.}{However, these methods are not applicable in scenarios involving ML-as-a-Service (MLaaS) or for users who seek to purify a backdoored model post-training.}
Addressing this scenario, we introduce \method{}, a post-training defense for malware classifiers that mitigates various types of backdoor embeddings without assuming any specific backdoor embedding mechanism.
Our method exploits the influence of backdoor attacks on the activation distribution of neural networks, independent of the trigger-embedding method.
In the presence of a backdoor attack, the activation distribution of each layer is distorted into a mixture of distributions.
By regulating the statistics of the batch normalization layers, we can guide a backdoored model to perform similarly to a clean one.
Our method demonstrates substantial advantages over several state-of-the-art methods, as evidenced by experiments on two datasets, two types of backdoor methods, and various attack configurations.
Our experiments showcase that \method{} can mitigate even the SOTA backdoor attacks for malware classifiers, e.g., Jigsaw Puzzle, which was previously demonstrated to be stealthy against existing backdoor defenses. Notably, our approach requires only a small portion of the training data --- only 1\% --- to purify the backdoor and reduce the attack success rate from 100\% to almost 0\%, a 100-fold improvement over the baseline methods. Our code is available at \url{https://github.com/judydnguyen/pbp-backdoor-purification-official}.
\end{abstract}

\IEEEpeerreviewmaketitle

\input{secs_R2/intro}

\input{secs_R2/background}
\input{secs_R2/method}

\input{secs_R2/experiments}
\input{secs_R2/related}
\vspace{-2mm}
\section{Conclusion}
In this paper, we present \method{}, a post-training backdoor purification approach based on our empirical investigation of distributions of neuron activations in poisoned malware classifiers.
\method{} makes no assumptions about the backdoor pattern type or method of incorporation using a small amount of clean data during the fine-tuning process.
By leveraging the distinct activation patterns of backdoor neurons, \method{} employs a two-phase strategy: generating a neuron mask from clean data and applying masked gradient optimization to neutralize backdoor effects.
Our extensive experiments demonstrate \method{}'s effectiveness and adaptability, outperforming existing defenses without requiring prior knowledge of attack strategies.
This approach substantially enhances the security and reliability of malware classification models in real-world applications.
\vspace{-2mm}
\section*{Acknowledgment}
We acknowledge partial support from the NSA Science of Security program, 
the DARPA agreement HR001124C0425, and the ARPA-H DIGIHEALS program. 
The views and conclusions contained herein are those of the authors and should not be interpreted as necessarily representing the official policies or endorsements, either expressed or implied, of the NSA, DARPA, ARPA-H, or the US Government.
{
\bibliography{main}
\bibliographystyle{IEEEtranN}
}
\vspace{-2mm}
\appendix
\vspace{-2mm}

\input{secs_R2/artifact_appendix}

\input{secs_R2/appendix}

\end{document}

%% file: secs_R2/intro.tex
\section{Introduction}

Malware classification has been witnessed 
dramatic advancements, particularly with the integration of Deep Neural Networks (DNNs) to tackle the increasing complexity and volume of modern malware corpora \fixed{\cite{liu2020review,gopinath2023comprehensive,bensaoud2024survey}}. %
Malware ensembles---including viruses, worms, trojans, and spyware---pose formidable risks to individuals and corporate institutions \fixed{~\cite{farhat2021brief,satvat2021extractor}}. 
The limitations of conventional detection methods, such as signature-based and heuristic techniques, in handling large-scale data and evolving malware variants have necessitated a shift toward more sophisticated Machine Learning (ML) and Deep Learning (DL) methodologies~\cite{chaganti2023multi, ravi2023attention, ahmed2023inception}.
Malware detection techniques based on ML/DL can model more complex patterns of malware data than classical signature-based ones.
This allows them to detect better new variants of existing malware or even previously unseen malware~\cite{deldar2023deep}.
\deleted{Feature extraction is a critical step in training a machine learning (ML) malware classifier using static analysis, which avoids the need for program execution. This process involves extracting code-based features, such as API calls, strings, and byte sequences, from malware samples. Unique features from Android apps and Portable Executable (PE) files, including permissions and section sizes, are also considered. These features are then transformed into vectors, graphs, or images for analysis by deep learning models~\cite{deldar2023deep,li2022malware}.}

\vspace{-2mm}
Integrating \replaced{DNNs}{deep neural networks (DNNs)} into malware detection systems has significantly advanced the field but has also introduced several threats. One notable vulnerability is the backdoor attack,~\cite{li2021backdoor,yu2023backdoor,wu2022backdoorbench,wenger2021backdoor}.
which involves an adversary discreetly embedding a harmful pattern or trigger within a small proportion of training samples to manipulate the behavior of the final model. When the model encounters this trigger during inference, it will misclassify the input, potentially leading to security breaches. These attacks subtly corrupt the model, often going undetected until the model systematically fails under specific conditions designed by the attacker, thereby undermining the integrity and reliability of the entire malware detection process~\cite{severi2021explanation,chakraborty2021survey,yu2023backdoor}.
\deleted{This emerging threat poses concerns similar to those observed in computer vision and is gaining significant attention in the malware detection community, underscoring the need for robust countermeasures~\cite{rosenberg2020query}}. Recently, research has focused on investigating the vulnerability of malware classifiers to backdoor attacks~\cite{severi2021explanation,yang2023jigsaw,li2021backdoor,zhang2023universal}. The general objective of these attack methods is to protect a subset of malware samples and bypass the detection mechanisms of malware classifiers. \added{This emerging threat poses concerns similar to those observed in other machine learning domains and is gaining substantial
attention in the malware detection community, underscoring the need for robust countermeasures~\cite{rosenberg2020query}.}

\vspace{-2mm}
To counter such threats, researchers have explored defense methods with backdoor attacks for malware classifiers. While these defenses show their effectiveness in mitigating backdoor attacks in machine learning models, the effectiveness of these countermeasures in the specific context of malware classifiers has been limited~\cite{severi2021explanation,yang2023jigsaw}.
For example, the study by Severi et al.~\cite{severi2021explanation} demonstrates that even anomaly detection methods~\cite{tran2018spectral,chen2018detecting} are not sufficient to fully protect malware classifiers against explanation-guided backdoor attacks. More recently, Yang et al.~\cite{yang2023jigsaw} have shown that their attack can bypass state-of-the-art defenses such as Neural Cleanse~\cite{wang2019neural} and MNTD~\cite{xu2021detecting}. These findings highlight the need for more robust countermeasures tailored to the malware detection \deleted{domain}. Another limitation is that these defenses often rely on assumptions about the backdoor, requiring intervention with all training data~\cite{yu2023backdoor}. Therefore, they cannot be applied to post-defense circumstances, such as \deleted{in} fine-tuning or backdoor removal in Machine Learning as a Service (MLaaS).
\added{A post-defense solution is essential when defenders acquire pretrained or publicly available backbone models for malware detection, either from third-party vendors or open-source repositories. When analysts discover malware samples mislabeled as safe, fine-tuning the model with small datasets serves as a practical and cost-effective alternative to full-scale retraining.}

To overcome the limitations of existing defenses, we introduce a 
post-training defense named \method{} to counter backdoor attacks in malware classifiers, \added{i.e., reducing the attack success rate from 100\% to almost 0\% and achieving 100-fold improvement.}
The core intuition behind \method{} is based on the observation that backdoor neurons exhibit distinct activation patterns when processing backdoored versus clean samples.
This distributional drift forms the basis of \method{}'s two-phase backdoor purification strategy.
In the first phase, a neuron mask is generated by training a noise model on clean data, identifying neurons whose activation patterns deviate significantly in backdoored samples.
This leverages clean data to discern normal activation patterns and detect backdoor neurons.
The second phase applies a masked gradient optimization process, reversing the gradient sign of the identified neurons during fine-tuning.
This mitigates the influence of backdoor neurons, ensuring the model correctly classifies triggered samples as malicious.
\fixed{In this work, we focus on two state-of-the-art backdoor attacks for malware classifiers: Explanation-guided~\cite{severi2021explanation} and Jigsaw Puzzle~\cite{yang2023jigsaw}. We evaluate these attacks on two public datasets: EMBER~\cite{anderson2018ember}, a Windows Portable Executable dataset, and AndroZoo~\cite{allix2016androzoo}, an Android malware dataset.}
The main contributions of our work can be summarized as follows:
\deleted{We experimentally verify that existing tuning-based defenses are less effective in purifying backdoor attacks in malware classifiers.}
\begin{itemize}
    \item \replaced{We demonstrate the activation’s distribution shift phenomenon caused by backdoor attacks in malware classifier models,
    leading to insights for mitigations.}{We demonstrate the activation's distribution shift phenomenon caused by backdoor attacks in malware classifier models, and reveal the insights for their success and stealthiness during the fine-tuning phase.}
    \item \replaced{We introduce \method{}, the first post-training defense against backdoor attacks in malware classification, using only fine-tuning with a portion of clean data. It operates independently of training and requires no prior knowledge of attack strategies, ensuring versatility and adaptability to various attack methods.}{We present \method{}, the first post-training defense against backdoor attacks in malware classification using only a portion of clean data as a fine-tuning phase.}
    \item \replaced{We conduct extensive experiments to demonstrate our method's SOTA performance with different settings regarding attack methods, datasets, fine-tuning data size, data poisoning rate, and model architecture.}{We conduct extensive experiments to demonstrate the efficiency of our method over other baselines with different settings regarding the amount of fine-tuning data and state-of-the-art backdoor attacks.}
    \item \added{We provide insights into model and defense behaviors under various attack strategies, highlighting the effectiveness of our method within the malware classification domain and its potential applicability to broader domains such as Computer Vision.}
\end{itemize}

%% file: secs_R2/background.tex
\vspace{-2mm}
\section{\added{Background}}
\label{sec:background}
\added{This section provides background on backdoor attacks
and defenses against malware classifiers.
We start with high-level descriptions of emerging threats of backdoor attacks on machine learning (ML) systems and classical defenses against them in Sec.~\ref{subsec:backdoor} and Sec.~\ref{subsec:defense}, respectively. In Sec.~\ref{subsec:backdoor-malware}, we discuss how backdoor attacks work on Deep Neural Networks (DNN)-based malware classifiers and their countermeasures. }

\vspace{-2mm}
\subsection{Backdoor Attacks}
\label{subsec:backdoor}
\added{
The emergence of outsourcing model training and MLaaS has led to emergent weaknesses~\cite{hitaj2019evasion,ning2021invisible}.
Backdoor attacks~\cite{liu2020reflection,gu2019badnets,ning2021invisible,hu2022badhash,gibert2023towards,gao2023not}, or trojan attacks, are prevalent training-phase adversarial attacks that primarily target DNN classifiers.
In general, backdoor attacks can be formulated as a multi-objective optimization problem~\cite{nguyen2024backdoor}, where the attacker seeks to optimize the following objectives:
{\small
\begin{equation}
    \theta^* = \min_{\theta} \mathbb{E}_{(x,y)\sim\mathcal{D}}^{}\mathcal{L}(f_\theta(x), y) + \mathbb{E}_{(x,y)\sim\mathcal{D}}^{}\mathcal{L}(f_\theta(\varphi(x)), \tau(y)),
    \label{def:backdoor-obj}
\end{equation}
}
in which $f_\theta$ is the victim model $f$ parameterized by $\theta$,
$\mathcal{D}$ is the set of training data for the main task, and $(x, y)$ are sample-label pairs uniformly drawn from $\mathcal{D}$.
The sample components $x$ are poisoned via the application of some function $\varphi$, which can be a non-transform function~\cite{wenger2021backdoor} or a perturbation function~\cite{doan2021lira, nguyen2024iba}; and the label counterparts $y$ are altered by another corresponding function $\tau$. 
Technically, the adversary's objective is to manipulate the model such that, for these poisoned samples $\varphi(x)$, it returns distorted outputs $\tau(y)$ instead of $y$. 
The function $\mathcal{L}$ in the expression $\mathcal{L}(f_\theta(\varphi(x)), \tau(y))$ represents a loss function that measures the discrepancy between the predicted output $f_\theta(\varphi(x))$ and the true output $\tau(y)$ for a given input sample $(x, y)$.
To ensure stealthiness, the performance of the model on non-backdoored samples remains unchanged. 
In particular, the model $\theta^*$ should give correct outputs for clean samples $x$. Backdoor attacks are particularly concerning as they can be stealthy and difficult to detect, making them a 
substantial threats to deploy secure and reliable DNN models.}

\vspace{-2mm}
\subsection{\added{Backdoor Countermeasures}}
\label{subsec:defense}
\added{Developing robust techniques to identify and mitigate the various types of backdoor attacks remains an important challenge in machine learning security research.
In contrast to the attack scenario, the multi-objective formulation for backdoor defense is defined as:
{\small
\begin{equation}
\label{eqn:defense-obj}
    \theta^* = \min_{\theta} \mathbb{E}_{(x,y)\sim\mathcal{D}}^{}\mathcal{L}(f_\theta(x), y) - \lambda\mathbb{E}_{(x,y)\sim\mathcal{D}}^{}\mathcal{L}(f_\theta(\varphi(x)), \tau(y)),
\end{equation}
}
where the first term also minimizes the loss on clean data samples, but the second one \textit{maximizes} the loss on backdoored data samples (note the negative sign).}
\added{The tradeoff between preserving clean data performance and backdoor removal can be controlled by a hyperparameter, $\lambda$.
To measure how well a backdoor defense scheme performs, we employ these two metrics:}

\begin{definition}[Clean Data Accuracy (\replaced{C-Acc}{CDA})]
    The \replaced{C-Acc}{ACC} is the proportion of clean test samples containing no trigger that is correctly predicted to their ground-truth classes.
\end{definition}
\begin{definition}[Attack Success Rate (ASR)]
The ASR is the proportion of clean test samples with stamped triggers that is predicted to the attacker's targeted classes.
\end{definition}

\begin{definition}[Defense Effectiveness Rating (DER)~\cite{zhu2023enhancing}] $\mathrm{DER} \in [0, 1]$ evaluates defense performance considering both the changes in \replaced{C-Acc}{ACC} and 
ASR. It is defined as follows:
$${DER}=[\max (0, \Delta \text{ASR})-\max (0, \Delta \text{C-Acc})+1] / 2$$
\end{definition}

\vspace{-2mm}
\added{For successful backdoored model $f_{\theta_{\mathrm{bd}}}$, the CDA should be similar to the clean model $f_{\theta_{cl}}$, while the ASR is high backdoored models can achieve an ASR that is close to $100 \%$ usually under outsource attributing to the full control over the training process and data.}
\added{Backdoor defenses can be deployed at different stages of the deep learning pipeline: during the classifier's training phase, post-training, or during inference.
Each scenario assumes a different defender role and capabilities.
The first approach aims to produce a backdoor-free classifier from a potentially poisoned training set.
Existing defenses focus on detecting and removing suspicious samples and identifying trustworthy samples~\cite{gao2019strip, liu2019abs}, and other works focus on modifying the training process for enhancing robustness~\cite{jin2022can, hong2020effectiveness}.
The second approach instead aims to remove the potential backdoor features in a well-trained model using clean data.
The defender first identifies the compromised neurons and then prunes or unlearns them~\cite{liu2018fine,wu2021adversarial,zheng2022data}, or others propose to ``unlearn'' the backdoor mapping from the victim model~\cite{zeng2021adversarial,li2021neural}.
The remaining approach aims to detect test samples with a backdoor pattern and potentially correct decisions.
Approaches are similar to post-training defenses, using input perturbation, suspicious region identification, or latent representation modeling~\cite{doan2020februus,gao2019strip}.}

\vspace{-4mm}
\subsection{\added{Backdoor Attacks and Countermeasures in Malware Classifiers}}
\label{subsec:backdoor-malware}

\noindent\textbf{\added{Backdoor attacks for malware classifiers. }}
In the backdoor attack setting, the adversary is assumed to have only partial control of the training process. 
Specifically, in malware classification, recent work assumes clean-label backdoor attacks~\cite{li2021backdoor,severi2021explanation,yang2023jigsaw} where the adversary has no control over the labeling of poisoned data. In these studies, the authors optimize a trigger or watermark within the feature and problem space of malware, which, when integrated with malware samples, activates the backdoor functionality. Specifically, the backdoored sample set has the form of $\mathcal{D}_{bd} = (\varphi(x), y)$ instead of $\mathcal{D}_{bd} = (\varphi(x), \tau(y))$ as normal label-flipping attacks in ML~\cite{gu2019badnets,feng2022fiba,wang2019neural,doan2021lira}. The optimized trigger is normally conducted in a model-agnostic fashion manner via feature explanations for model decision~\cite{severi2021explanation,li2021backdoor} and alternative optimization~\cite{yang2023jigsaw}.

\begin{figure}[h]
    \begin{subfigure}
    {0.48\columnwidth}
    \includegraphics[width=1.0\textwidth]{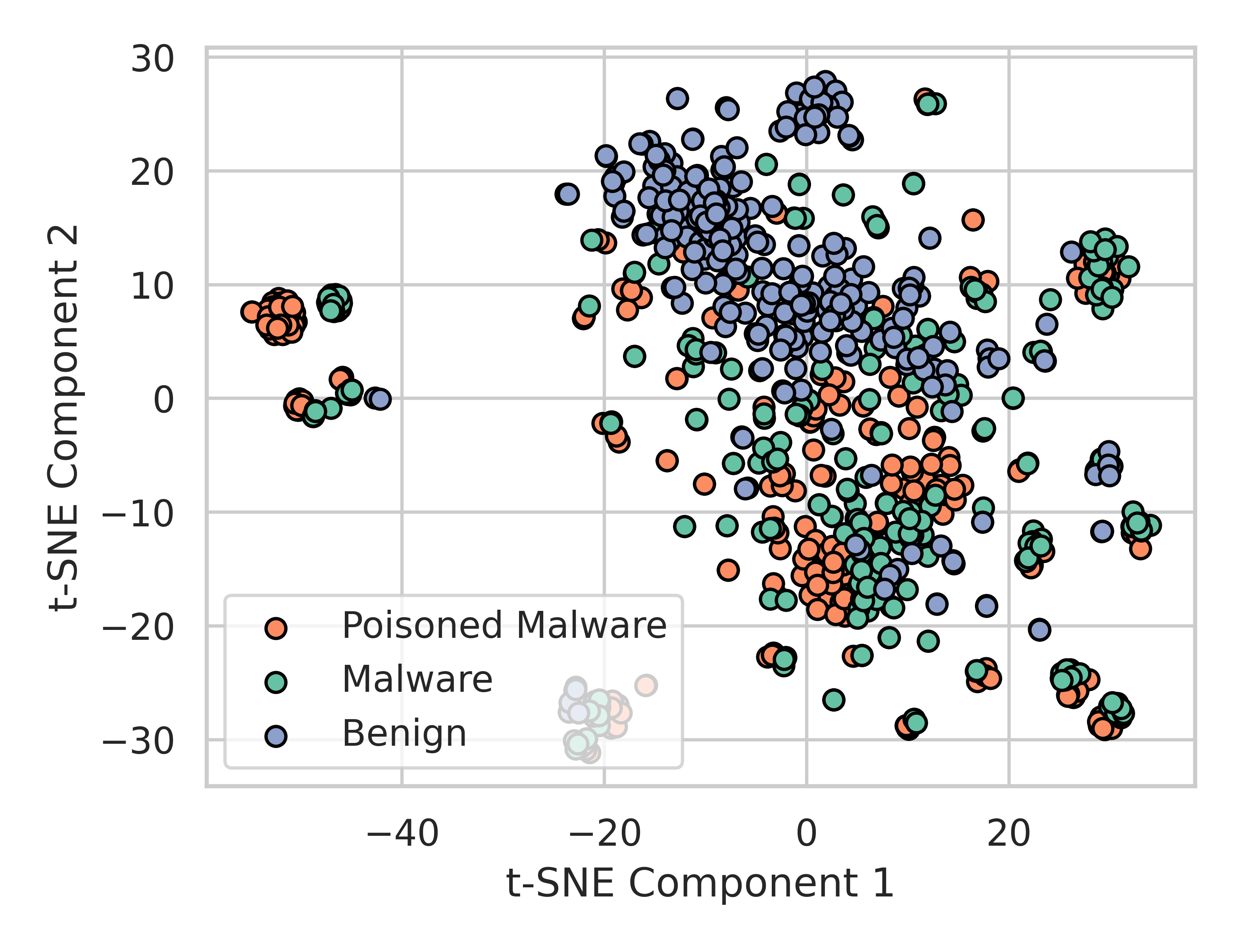}
    \caption{Non-family-targeted backdoor}
    \label{fig:illus-transfer}
    \end{subfigure}
    \begin{subfigure}
    {0.48\columnwidth}
    \includegraphics[width=1.0\textwidth]{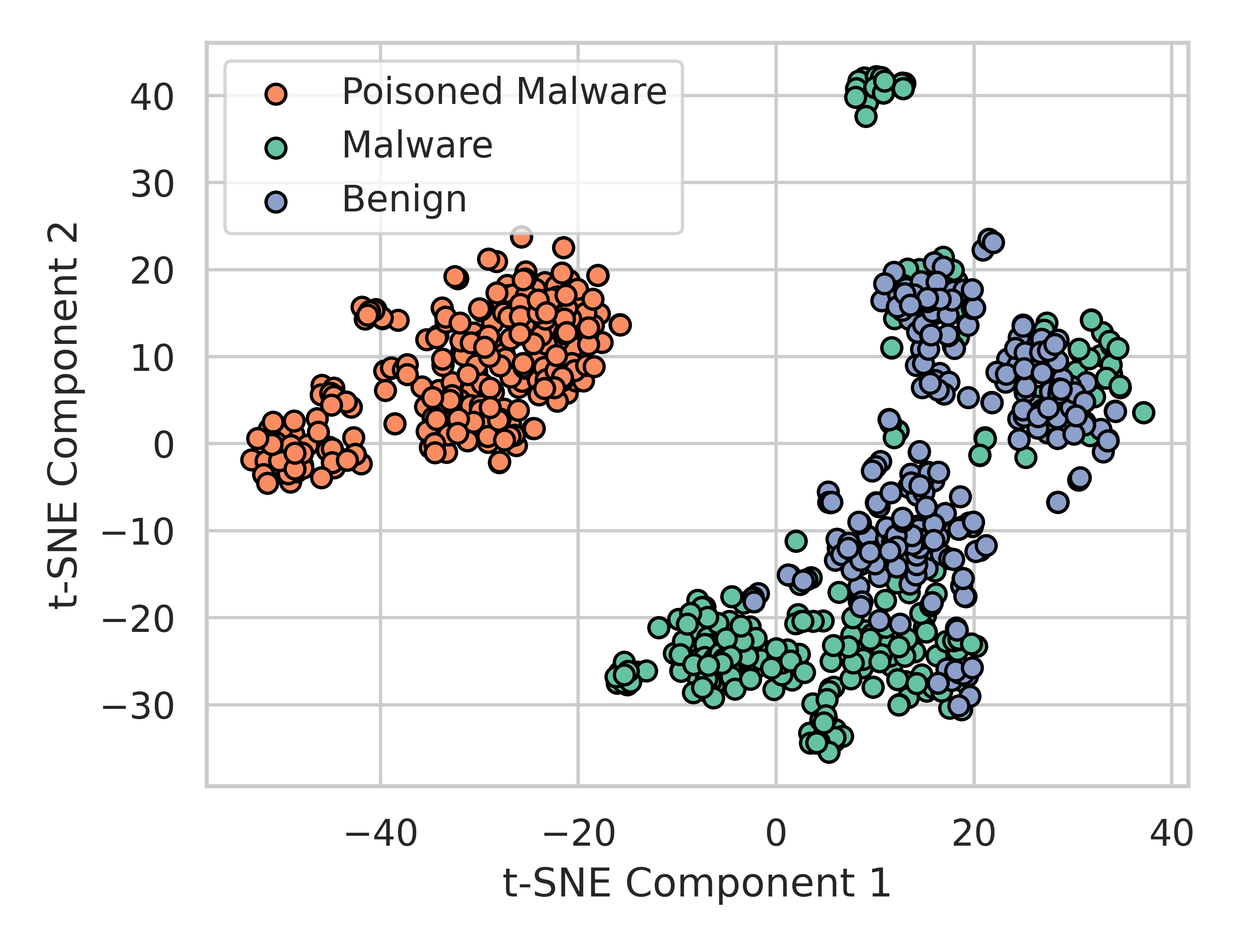}
    \caption{Family-targeted backdoor}
    \label{fig:illus-transfer}
    \end{subfigure}
    \caption{t-SNE~\cite{tnse} representations of family-targeted and non-family-targeted backdoor attacks.}
\label{fig:poison-samples}
\vspace{-8mm}
\end{figure}

The high-level idea of these backdoor attack strategies is to generate a trigger function $\varphi$, known as a watermark, to combine with the targeted samples.
Due to the characteristics of the targeted samples, the attack can be family-targeted or non-family-targeted.
For example, Severi et al.~\cite{severi2021explanation} use the trigger to distort the model's prediction on any malware samples that carry the trigger to be ``benign.''
On the other hand, Yang et al~.\cite{yang2023jigsaw} only target the samples belonging to a specific malware family, and the trigger is designed for this family only and cannot activate the backdoor if it is combined with other families.
We plot the poisoned samples generated by these two backdoor categories in Fig.~\ref{fig:poison-samples}.
As presented, non-family-targeted poisoned samples are manipulated from the set of all other malware families, while in a family-targeted setting, these poisoned samples belong to a separate malware family that the adversary wants to protect.

\noindent\textbf{Backdoor defenses for malware classifiers. }
A popular defense mechanism against backdoor attacks for malware classification is adversarial training~\cite{wang2020mdea}. %
This approach tries to stabilize the model's performance on poisoned data by training it with adversarial examples, with or without the original training examples. However, this method inevitably introduces significant additional cost for generating adversarial examples during training.
Another approach leverages various heuristics to remove adversarial samples from the training dataset. 
This way, any manipulations committed by the adversaries can be undone before the samples are sent to the PE malware detector. However, these empirical defenses usually only work for very few adversarial attack methods, making them usually attack-specific~\cite{ling2023adversarial}.
Moreover, these methods require preemptive training control of the original model, which does not apply to the case of Machine Learning as a Service (MLaaS). When we buy or acquire a trained poisonous model, we need a purification scheme that can repair the poisoned model and ensure the adversarial vulnerability is no longer present.

To address this concern, we propose to use fine-tuning (FT) method since it has been adopted to improve 
models'
robustness against backdoor attacks~\cite{liu2018fine,wu2022backdoorbench} and can be 
combined with existing training methods and various model architectures. 
Additionally, FT methods require less computational resources, making them one of the most popular transfer learning paradigms for large pre-trained models~\cite{wei2021finetuned,radford2021learning,zhang2022fine}. However, FT methods have not been sufficiently evaluated under various attack settings, particularly in the more practical low poisoning rate regime.
To the best of our knowledge, there has been no prior work on purifying backdoors during fine-tuning for malware classification.

%% file: secs_R2/method.tex
\begin{figure*}[h]
    \centering
    \begin{subfigure}
    {0.32\textwidth}
    \includegraphics[width=1.0\textwidth]{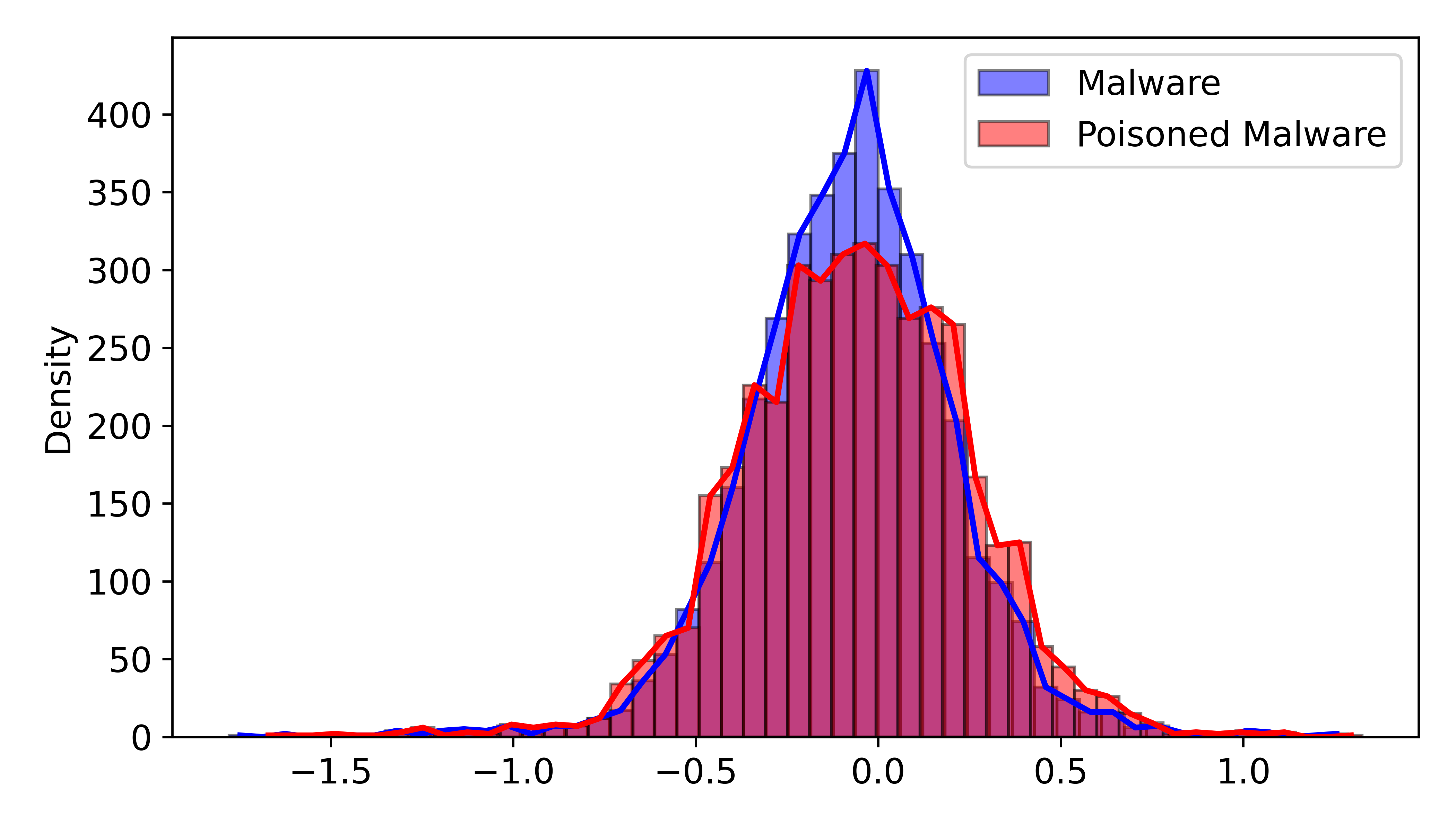}
    \caption{Clean Model}
    \label{fig:bd-neuron-clean}
    \end{subfigure}
    \begin{subfigure}
    {0.32\textwidth}
    \includegraphics[width=1.0\textwidth]{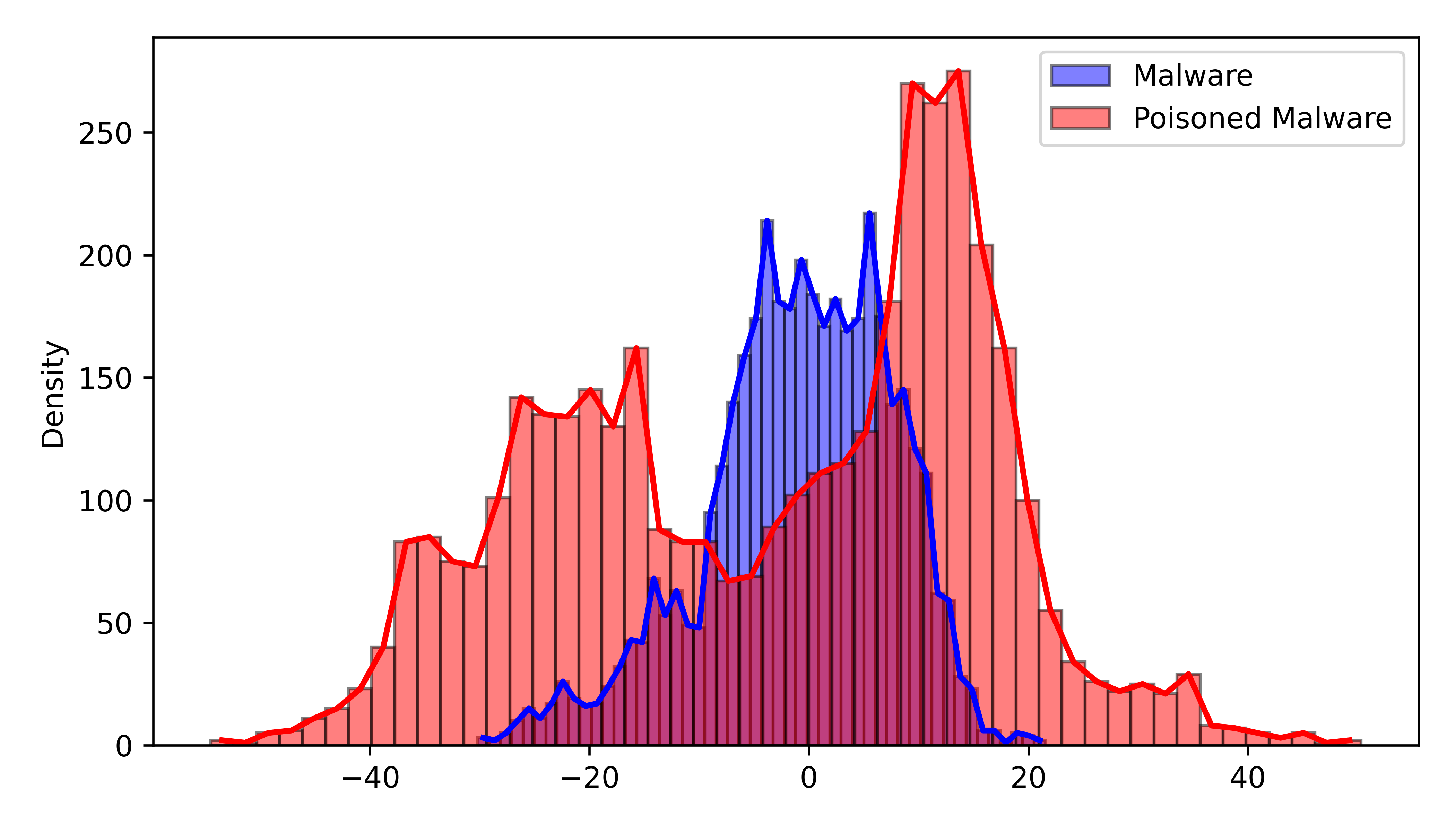}
    \caption{Backdoored Model}
    \label{fig:bd-neuron-bd}
    \end{subfigure}
    \begin{subfigure}
    {0.32\textwidth}
    \includegraphics[width=1.0\textwidth]{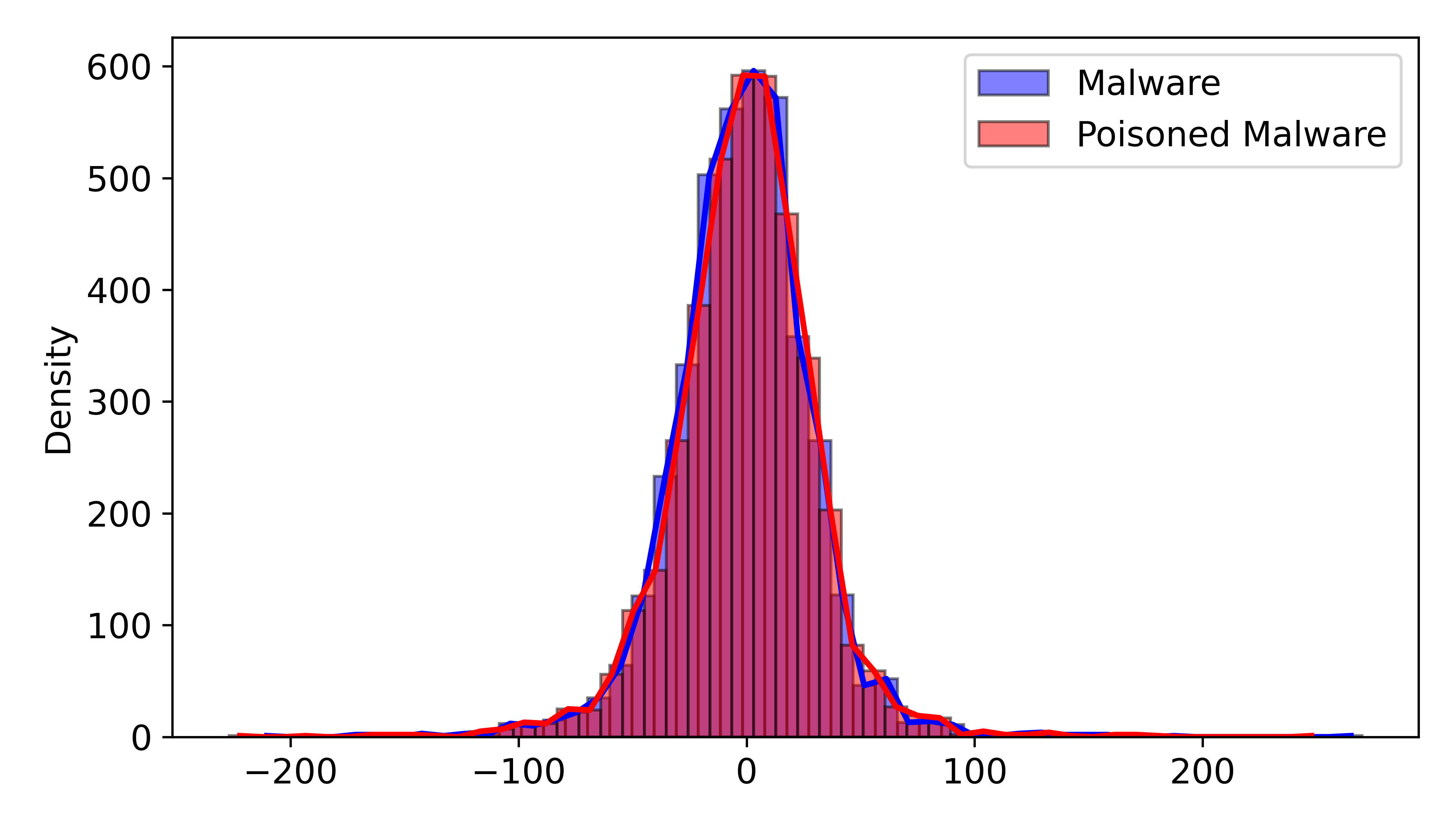}
    \caption{Fine-tuned Model}
    \label{fig:bd-neuron-ft}
    \end{subfigure}
    \caption{Model activation of backdoor neurons on targeted malware samples with and without a trigger.}
    \label{fig:backdoor-neurons}
    \vspace{-6mm}
\end{figure*}

\vspace{-2mm}
\section{Methodology}
In this section, we present \method{}, an approach to purifying classification models that have been poisoned.
First, we discuss our insight into the activation distribution of backdoor neurons during training (Section~\ref{subsec:neurons}). 
Second, we discuss the threat model and goals of \method{} (Section~\ref{subsec:threat}).
Third, we present a detailed description of \method{} (Section~\ref{subsec:pbp}). %

\vspace{-2mm}
\subsection{Backdoor Neurons}
\label{subsec:neurons}
When a classifier has been poisoned with a backdoor attack, there is a specific subset of neurons that play a substantial role in exhibiting the backdoor behavior~\cite{zheng2022pre,li2024purifying}.
The attack success rate will be 
dramatically lowered if a portion or all of these \emph{backdoor neurons} are pruned from the infected model~\cite{zheng2022pre,wu2021adversarial}.
Leveraging this insight, we investigate the activation distribution of the model when subjected to backdoor attacks. %
In this subsection, we perform an empirical analysis to examine the distribution of the backdoor neurons when a backdoor is introduced into the model.
We begin with a definition of backdoor neurons. 

\begin{definition}[Backdoor Neurons]
Given a model $f$ and a poisoning function $\varphi$, the backdoor loss on a dataset $\mathcal{D}$ is defined as:
$$
\mathcal{L}_{\mathrm{bd}}(f_{\theta})=\mathbb{E}_{({x}, y) \sim \mathcal{D}}\left[\mathrm{D}_{\mathrm{CE}}(y, f_{\theta}(\varphi({x}))]\right.
$$
where $\mathrm{D}_{\mathrm{CE}}$ denotes the cross entropy loss.
\end{definition}
\begin{definition}[Backdoor Sensitivity] Given a model $f$, the index of a neuron $(l, k)$ and the backdoor loss $\mathcal{L}_{\mathrm{bd}}$, the sensitivity of that neuron to the backdoor is defined as:
$$
\zeta(f, l, k)=\mathcal{L}_{\mathrm{bd}}(f)-\mathcal{L}_{\mathrm{bd}}(f_{-\{(l, k)\}}),
$$
where $f_{-\{(l, k)\}}$ is the network after pruning the $k$-th neuron of the $l$-th layer.
\end{definition}

\vspace{-2mm}
Based on these definitions, a neuron with larger $\zeta$ has more importance to backdoor functionality.
Although this definition does not consider the joint effect of neurons that may lead to the misidentification of important neurons~\cite{li2024purifying,fan2022defense},
it is sufficient for our analysis. %
Using this definition, we can filter out backdoor neurons in an infected model and analyze their behaviors in both the clean version and backdoored modes.

\begin{figure*}[bth]
    \centering
    \includegraphics[width=0.9\textwidth]{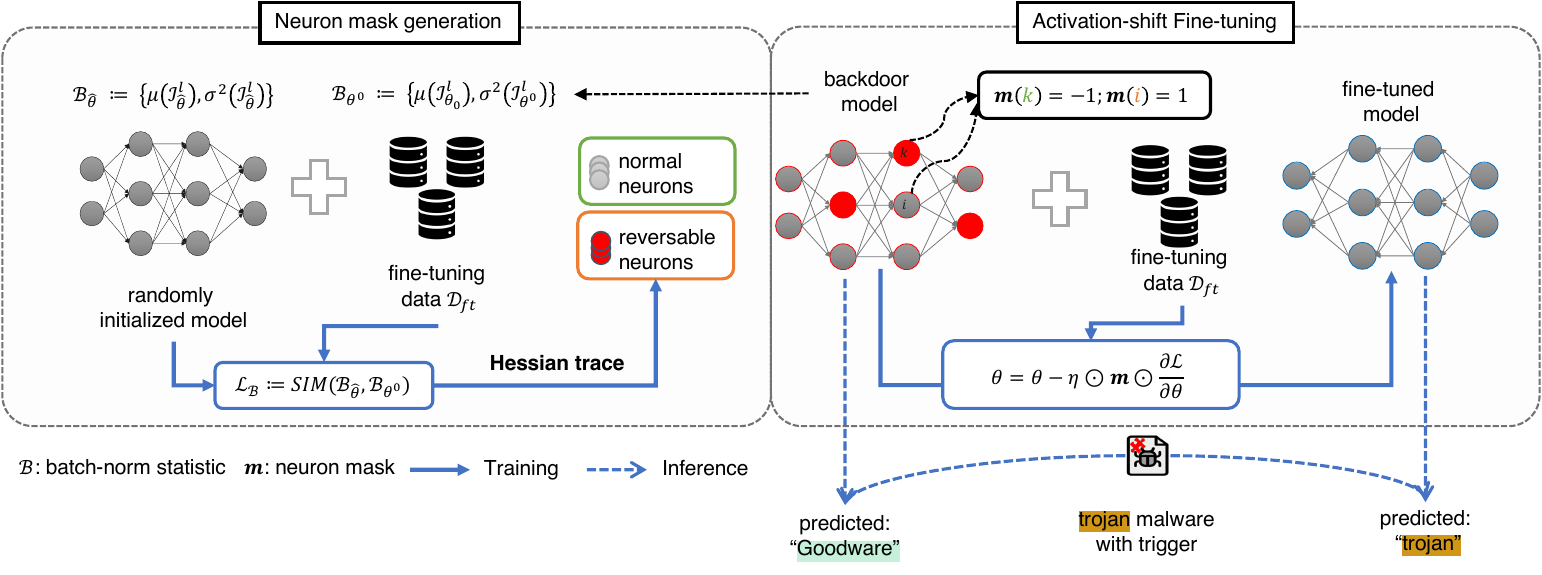}
    \caption{The overall description of the proposed method. \method{} includes two phases: (i) Neuron mask generation and (ii) Activation-shift Fine-tuning. In the first phase, we initialize a noise model $f_{\hat{\theta}}$ and train a new model by using clean data using the objective functions as aligning the neuron activation to the backdoor model $f_{\theta^0}$, determining the most important neurons for this task using Hessian trace during training. In the later phase, the masked gradient optimization process is applied by reversing the gradient sign of the masked neuron (in red). The fine-tuned model is expected not to predict triggered sample, i.e., malware as ``benign''.}
    \label{fig:overall}
    \vspace{-0.5cm}
\end{figure*}
\noindent\textbf{Settings.} We explore backdoor neurons by analyzing their activation, which is defined as the output of a neuron under a certain input. 
This reflects the sensitivity of a neuron with a given input and directly relates to the final prediction. 
We trained a backdoored model following \citet{severi2021explanation} using the EmberNN~\cite{severi2021explanation} architecture with four Dense layers. 
Concurrently, we trained a comparative clean model with the same setting.
Specifically, for the backdoored model, we injected a backdoor trigger pattern into a subset of the training data, following the procedure described in \citet{severi2021explanation}. 
The clean model was trained on the original, unmodified dataset. 
Both models were trained until convergence using the same hyperparameters and optimization settings. 
We observe the corresponding sensitivity on the triggered data and select the top 20\% of the neurons whose highest backdoor sensitivity is defined above each layer for further investigation on their investigation. 
Using this set of backdoor neurons, we stored the activation of these neurons on the clean, backdoored, and fine-tuned models, given different categories of inputs including malware samples with and without the backdoor trigger, and plotted the result on \added{Fig. }~\ref{fig:backdoor-neurons}.

\noindent\textbf{Empirical results. } Fig.~\ref{fig:backdoor-neurons} shows that, on triggered samples, backdoor neurons generally show a 
distribution deviation from the original activation on non-triggered samples and after the trigger is combined with these samples. 
From Fig.~\ref{fig:bd-neuron-clean}, a clean model does not show a 
substantial distribution deviation on the normal malware and the malware added the trigger. 
Meanwhile, the backdoored model in Fig.~\ref{fig:bd-neuron-bd} undergoes 
substantial distribution deviation. 
Given the poisoned data, this deviation accumulates with each layer, causing the victim model to increasingly diverge from benign activations, ultimately leading to incorrect target labeling when predicting a poisoned sample. 
The ultimate goal after training is to navigate the backdoor model's activation distribution so that there is no 
substantial
drift, as shown in Fig.~\ref{fig:bd-neuron-ft}. 

\subsection{Threat Model and Problem Formulation}
\label{subsec:threat}
A large fraction of the backdoor attack literature~\cite{Gu2017BadNetsIV,zheng2022pre,zheng2022data,mmbd} assumes the threat model of ``Outsourced Training Attack,'' in which the adversary has full control over the training procedure and the end user is only allowed to check the training using a held-out validation dataset. 
\deleted{or ``Transfer Learning Attack,'' where the attacker downloads a pre-trained model and finetunes it before passing it to the end user. However, in the malware classification setting, the former is too strict and the latter is too loose of an assumption. }
\added{However, adversaries introducing backdoor attacks ensure that the victim model performs well on clean data, making the reliance on a held-out validation dataset insufficient for verifying the model's trustworthiness.
This presents a strict assumption for users regarding the safe use of DNN models.}
\added{To address this challenge, we adopt a defense setting where the defender acquires a backdoored model from an untrusted source and assumes access to a small subset of clean training data for fine-tuning~\cite{li2023reconstructive,li2021neural}.}
\added{Backdoor erasing aims to eliminate the backdoor trigger while maintaining the model's performance on clean samples. 
This approach is particularly relevant when training data is no longer fully accessible due to retention or privacy policies. 
Additionally, users of third-party ML services may inadvertently purchase backdoored models and seek to purify them using their data.}
\deleted{Instead, existing works on this topic}~\cite{yang2023jigsaw,severi2021explanation} \deleted{only assume the clean-label threat model, where the attacker can alter the malware samples slightly but not the corresponding labels. This is because the labeling process is determined by third-party AV analyzers, and is not under direct control of the attacker.}

\noindent\textbf{Attacker's goals. }
Similar to most backdoor poisoning settings, we assume the attacker's goal is to alter the training procedure, specifically the malware sample set, such that the resulting trained backdoored classifier, $f_\mathrm{bd}$, differs from a cleanly trained classifier $f_\mathrm{cl}$.
An ideal $f_\mathrm{bd}$ has the same response to a clean set of inputs $x$ as $f_\mathrm{cl}$, whereas it generates an adversarially chosen prediction, $\tau(y)$, when applied to backdoored inputs, $\varphi(x)$. These goals can be summarized as:
$$
f_\mathrm{bd}(x)=f_\mathrm{cl}(x) ; \quad f_\mathrm{bd}\left(\varphi(x)\right)=\tau(y) \neq y.
$$

\vspace{-2mm}
Specifically, we use class 0 for benign binaries and class 1 for malicious. 
Given the opponent is interested in making a malicious binary appear benign, the target result is always $\tau(y)\equiv 0$. 
Additionally, we also consider family-based malware classification, in which the adversary aims to manipulate the surrogate model to classify the specific samples into one targeted malware file. To make the attack undetectable, the adversary wishes to minimize both the poisoning rate and the footprint of the trigger (i.e., the number of modified features). 

\noindent\textbf{Defender's goal. }
As opposed to the attacker goals, the defender, i.e., the model trainer who has full access to the internal architecture of the target model and a limited set of benign fine-tuning data, denoted as $\mathcal{D}_{ft}$, aims to achieve two goals. 
The first goal is to erase the backdoors from $f_\mathrm{bd}$ and make the purified model perform correctly even with triggered inputs. 
To maintain utility, the second goal is to maximally retain the model's performance on normal inputs during the purifying process. 
In this work, we also adopt the assumptions to obtain the objective from Eqn.~\ref{eqn:defense-obj} from related post-training defenses as follows:
\begin{enumerate}
    \item The defender has no information about the \fixed{backdoor trigger} available nor the specific adversary's accessibility including the poisoning rate and how the backdoor is inserted. %
    This assumption is relaxed by many existing post-training detectors by making assumptions about the backdoor pattern type, or how human imperceptibility is achieved. However, we strictly make no assumptions about the backdoor trigger/watermark that may be inserted.
    \item \replaced{The defender has no access to the training procedure, and cannot acquire the full training dataset to retrain a new model. The defender is a user of the classifier or of a legacy system who has the access to a trained/backdoored model. }{The defender has no access to the 
    classifier's training set and cannot intervene during or before the training procedure. 
    The defender is a user of the classifier or the user of a legacy system.}
    \item \replaced{The defender can independently collect or access a small, clean dataset that is representative of the training data, e.g., samples from all classes (positive and negative), and can combine it with a portion of the training data if they have access to it. This assumption aligns with most post-training backdoor defenses}{The defender can independently collect or has access to a small, clean data set containing samples from all classes in the domain. This assumption is used by most post-training backdoor defenses}~\cite{min2024towards,mmbd}.
\end{enumerate}
Given these assumptions and constraints, the defender faces substantial challenges in effectively removing the backdoors while preserving the model's original performance. 
The key is to develop a purification technique that can reliably identify and neutralize the backdoor neurons without degrading the model's utility on clean data. \deleted{This requires a deep understanding of the activation patterns and behavior of these neurons, which we will explore in the following sections.}
\fixed{Given these objectives, we define \textit{Backdoor Purification} as follows:
\begin{definition}[Backdoor Purification]
    Given a backdoored model $f_{\theta_0}$ which is trained on $\mathcal{D} = \mathcal{D}_{c} \cup \mathcal{D}_{bd}$ such that:
    $$
         \mathbb{E}_{(x,y)\sim\mathcal{D}}^{}\mathcal{L}(f_\theta(x), y) \leq LB_{c},
    $$
    and
    $$
         \mathbb{E}_{(x,y)\sim\mathcal{D}_{bd}}^{}\mathcal{L}(f_\theta(\varphi(x)), \tau(y)) \leq LB_{bd},
    $$
    the backdoor purification process uses a subset $\mathcal{D}_{ft}$ to fine-tune model $f_{\theta_0}$ such that the final model $\theta_{T}$:
        $$
        \mathbb{E}_{(x,y)\sim\mathcal{D}_{c}}^{}\mathcal{L}(f_{\theta_{T}}(x), y) \leq \epsilon_{c},
        $$
        where $\epsilon_{c}$ is a small value indicating high accuracy on clean data;
        and
        $$ \mathbb{E}_{(x,y)\sim\mathcal{D}_{bd}}^{}\mathcal{L}(f_{\theta_{T}}(\varphi(x)), \tau(y)) \geq \epsilon_{bd},
        $$
        where $\epsilon_{bd}$ is a value indicating that the model is robust against the backdoor trigger.
\end{definition}}

\vspace{-2mm}
\subsection{\method{} Approach}
\label{subsec:pbp}
In this subsection, we discuss the key ideas behind \method{}, which consists of two steps: (i) neuron mask generation and (ii) activation-shift fine-tuning.

\subsubsection{Neuron Mask Generation} 
Motivated by the observation of activation drift, the key insight is that correcting this deviation could effectively mitigate backdoor effects.
\replaced{In this step, we first train a model, i.e., $\tilde{\theta}$, from scratch to realign the activation of the new model with that of the backdoored model, $\theta_0$, then we find the backdoored neurons $\mathcal{N}_m$ by analyzing the gradient trace $\nabla_\theta \mathcal{L}$ during this training procedure. }
{The objective is to first train a model, i.e., $\tilde{\theta}$, from scratch to realign the activation of the new model with that of the backdoored model, $\theta_0$, in which backdoor neurons express shifted activations.
Then, the gradient trace $\nabla_\theta \mathcal{L}$ during this procedure is observed to find the set of important neurons $\mathcal{N}_m$ in activating the backdoor task from a clean model.}
\added{It is worth noting that, this procedure does not output the new fine-tuned model to be used as-it-is since the defender does not have enough data to train a model from scratch. }
\added{Instead, we want to observe how a clean model changes to match and mimic the behavior of the backdoored one. From that, we detect the most important neurons leading this change and consider them as potential backdoor neurons contributing to the reconstruction of backdoor function. }

\deleted{To achieve this objective, we use two loss terms namely, clean loss $\mathcal{L}_{CE}$ and alignment loss $\mathcal{L}_{align}$.} 
Assume model $f$ has \added{total} of $L$ Batch Normalization (BN) layers. \replaced{Each BN layer records the running mean and variance of the input during training (which includes both clean samples and adversarially poisoned samples)}{Each BN layer records the running mean and variance of the original input during training}, denoted as $\mathcal{B} = \{\hat{\mu}_i^l, \hat{\sigma}_i^{2 l} | i=1, \ldots, n\}$. \added{During retraining step,} when a batch of inputs $\mathcal{I}^l$ from the previous layer is provided, we can calculate their mean and variance, denoted as $\{\mu_i^l(\mathcal{I}^l), \sigma_i^{2 l}(\mathcal{I}^l) | i=1, \ldots, n\}$.
\replaced{Our objective is to guide the clean model to behave as if it is processing both clean and backdoored data by ensuring that the input to each layer matches the corresponding mean and variance statistics recorded in the BN layers of the backdoored model. By aligning these internal distributions, we effectively force the clean model to mimic the internal dynamics of the original model, thereby reconstructing the embedded backdoor function. }{The objective is to optimize a new model such that the input of each layer fits the statistics (mean and variance) in the BN layers stored in the backdoored model.} 
\added{To achieve this objective, we use two loss terms namely, clean loss $\mathcal{L}_{CE}$ and alignment loss $\mathcal{L}_{align}$, where $\mathcal{L}_{CE}$ is used as achieving benign task and $\mathcal{L}_{align}$ is used to achieving the backdoored task.}

\added{First, $\mathcal{L}_{\mathrm{ce}}$ is the cross-entropy loss between model output $f_{\tilde{\theta}}(\boldsymbol{x})$ and the true label $y$, i.e., $\mathcal{L}_{\mathrm{CE}} = - \sum_{i=1}^{C} y_i \log(f_{\tilde{\theta}}(\boldsymbol{x}))$.}
\replaced{We then define $\mathcal{L}_{align}$ as a layer-wise activation alignment objective~\cite{frantar2022spdy,lu2022domain,li2024purifying,li2024nearest}, which is widely used in knowledge distillation tasks.}{We leverage a layer-wise activation alignment objective~\cite{frantar2022spdy,lu2022domain,li2024purifying,li2024nearest} objective, which is widely used in knowledge distillation tasks.} 
Let $\boldsymbol{\theta}_L^n$ represent the weights of the $l$-th block that contains a BN layer and $l \in \{1, \ldots, n\}$ in the original backdoored model, and $\boldsymbol{\tilde{\theta}}_n^l$ denote the weights in the corresponding layer of the new model.
Using $\mathcal{I}^l$ to represent the batch of inputs for the $l$-th layer, the activation alignment objective can be formulated as:
{\small
\begin{equation}
\label{eqn:alignment}
\mathcal{L}_{align} = \sum_{l=1}^n || \mu(\boldsymbol{\theta}_o^l \mathcal{I}^l) -  \mu(\boldsymbol{\tilde{\theta}}^l \mathcal{I}^l) ||_2+|| \sigma^2(\boldsymbol{\theta}_o^l \mathcal{I}^l) -  \sigma^2(\boldsymbol{\tilde{\theta}}^l \mathcal{I}^l) ||_2,
\end{equation}
}
where $\mu$ and $\sigma$ are the batch-wise mean and variance estimates corresponding to the output of the $l^{\text{th}}$ layer and the $\|\cdot\|_2$ operator denotes $\ell_2$ norm calculations. 
The goal of this loss term is to minimize the L2 distance between the means and variances of the activations in the original backdoored model and the new model, layer by layer.
To this end, the overall objective for this step is represented as:
\begin{equation}
    \added{\mathcal{L}_{re} = \mathcal{L}_{\mathrm{ce}} + \alpha*\mathcal{L}_{align}},
    \label{eqn:align_loss}
\end{equation}
\deleted{Here, $\mathcal{L}_{\mathrm{ce}}$ is the cross-entropy loss between the model's output $f_{\tilde{\theta}}(\boldsymbol{x})$ and the true label $y$. $\mathcal{L}_{align}$ is the activation alignment objective defined previously, and where $\alpha$ is a hyperparameter that balances the two terms.}

\vspace{-2mm}
\replaced{When a newly initialized model $\tilde{\theta}$ is trained on this objective, this model will be optimized to achieve two objectives at the same time, which are learning the original benign task, i.e., classifying malware samples; and aligning the activations on the backdoored samples as the victim model.}{The overall goal is to jointly optimize the model's performance on the original task (minimizing $\mathcal{L}_{\mathrm{ce}}$) and align the activations with the backdoored model (minimizing $\mathcal{L}_{align}$).} 
Intuitively, the above objective learns the main task of the malware classifier from scratch while \replaced{self-implanting the backdoor inserted in $f_{\theta_0}$}{aligning the input of each layer to match the statistics in the corresponding BN layer stored in the original model $f_{\theta_0}$, to make this model self-implant the backdoor in $f_{\theta_0}$}. 
The hyper-parameter $\alpha$ term in Eq.~\ref{eqn:align_loss} is to bound the alignment loss added $\mathcal{L}_{align}$ and void overfitting\deleted{to the BN statistics}.

\noindent\textit{Important neuron mask generation. }
\added{To achieve dual objectives of backdoor purification, we need to focus on erasing behaviors caused by a set of backdoor neurons instead of the whole network. By scrutinizing the changes in neurons during the retraining phase above, we can find these neurons using the sparse nature of NNs. Indeed, } recent research has discovered the sparse nature of gradients in stochastic gradient descent (SGD), which means that during the training process, only a small number of coordinates are updated~\cite{ivkin2019communication,stich2018local}. This characteristic of SGD highlights that most updates are concentrated in a limited subset of parameters. 
In our work, we leverage the observation of the trace of gradients during training to identify important neurons for a training task. 
Empirically, the majority of the $\|\cdot\|_2$ norm of the training gradient is contained in a very small number of coordinates, which exploits the sparse nature of gradients in SGD~\cite{ivkin2019communication,zhang2022neurotoxin}. 
Specifically, for each layer in the neural network, we want to find the $top-K$ coordinates whose gradient has the highest magnitude:
$$
\mathcal{N}_{m} = \text{argmax}_k \|\nabla_\theta \mathcal{L}_{re}(\tilde{\theta})\|_2,
$$
where $\nabla_\theta \mathcal{L}(\tilde{\theta})$ is the gradient of the loss function $\mathcal{L}$ with respect to the model parameters $\theta$ at the current point $\theta'$, and $\text{argmax}_k$ returns the indices of the $k$ largest values. 
By computing the $\|\cdot\|_2$ norm of the gradient for each coordinate (i.e., each neuron or weight), we can identify the $top-K$ coordinates with the largest gradient magnitudes. 
These coordinates correspond to the neurons that have the highest impact on the loss function and are therefore considered the most important for the learning task with the objective of Eqn.~\ref{eqn:align_loss}. In other words, these neurons are the most important ones for aligning the activation of the new model with the backdoored one while achieving clean accuracy, i.e., achieving dual objectives of a backdoor attack in Def.~\ref{def:backdoor-obj}. \added{By suppressing neurons associated with the backdoor function, the neuron mask guides the erasing process to focus on the cause of the backdoor while ensuring that the remaining neurons still provide accurate predictions on clean data, maintaining the model's utility.}

\subsubsection{Activation-shift Model Fine-tuning}
To mitigate the backdoor in an input model, we first conduct model weight perturbation before starting the fine-tuning procedure. 
\deleted{Addressing backdoors is critical for ensuring the security and reliability of machine learning models, as they can lead to malicious exploitation of the model. }
\replaced{This step aims to perturb the model’s weights, establishing a new starting point that forces significant updates throughout the network toward the optimum based on the objective functions. Moreover, this step is considered as an ``initialization'' similar to fine-tuning methods to avoid bias and influence from the previous trained model.}{This step aims to 
dramatically alter the original model parameters, thereby forcing the fine-tuning process to update the backdoor-related neurons with large-magnitude changes.} This method is widely used in the literature to reduce the effect of adversarial attacks~\cite{xie2021crfl, pillutla2022robust, wu2021adversarial}, since perturbing the weight of the model also results in a change in the prediction of the model.
We rigorously add Gaussian noise to perform this step:
\begin{equation}
\theta_{0} = \theta_0 + \mathcal{N}(0, \sigma^2 I),
\label{eqn:smoothing}
\end{equation}
where $\theta_0$ are the model parameters of the backdoored model, and $\mathcal{N}(0, \sigma^2 I)$ is a Gaussian noise term with zero mean and covariance $\sigma^2 I$, and where $I$ is the identity matrix. 
Previous studies have shown that adding Gaussian noise can effectively disrupt adversarial attacks, making it a robust choice for this application~\cite{xie2021crfl, pillutla2022robust, wu2021adversarial}. 
Unlike the relatively large noise required for differential privacy, our goal is not privacy but rather the prevention of attacks. 
Therefore, we add a small amount of noise that is empirically sufficient to limit the success of attacks without significantly compromising the model's performance.
\replaced{This approach aims to mitigate the backdoor effect introduced by the adversary. After this step, the fine-tuned model $\theta_{0}$ will require substantial updates during fine-tuning to align with the original model. However, we observe that, despite using only clean data for fine-tuning, the model often converges back to the backdoored function, a phenomenon consistently seen across various fine-tuning methods.}{This approach not only mitigates backdoor threats but also maintains the integrity of the model, ensuring that it remains functional and accurate for legitimate tasks. After this step, we expect that the backdoored model $\theta_0$ will require significant updates during fine-tuning to converge to the original model.}
\replaced{To counter this, the applied perturbation disrupts the backdoor-related neurons, forcing the model to make additional adjustments to regain its performance. This disruption is crucial, as it lays the groundwork for the next step: reversing specific parameter updates in the backdoored neurons to fully neutralize the backdoor effect.}{
This expectation is based on the empirical observation that substantial changes to the model parameters disrupt the functionality of backdoor-related neurons, thereby necessitating considerable adjustments to restore the model's performance. 
This perturbation sets the stage for the next crucial step: reversing the updates of specific coordinates to further neutralize the backdoor effect.}

\input{algo/main-algo}
\replaced{We erase the backdoor function from model $f$ by reversing the gradient direction of the backdoor neurons by altering the signs of the corresponding updates while keeping the updates from the remaining coordinates. }{To move the model in a particular direction, we reverse the updates of specific coordinates by altering the signs of the updates. }
For every dimension belonging to $\mathcal{N}_m$, the learning rate is multiplied by $-1$, effectively maximizing the loss on that dimension instead. This process can be described mathematically as follows:
\begin{equation}
\boldsymbol{m}_{\theta, i}= \begin{cases}1, & i \notin \mathcal{N}_{m} \\
-1, & \text { otherwise }\end{cases}
\end{equation}
The model is then updated in each iteration using: 
\begin{equation}
\theta_{t+1}=\theta_{\mathrm{t}}-\eta \odot \boldsymbol{m} \odot \frac{\partial \mathcal{L}_{\mathrm{ce}}\left(f_{\tilde{\theta}}\left(\boldsymbol{x}\right), y\right)}{\partial \theta_t},
\end{equation}
where $\boldsymbol{m}_{\theta, i}$ is a masking vector that flips the sign of the gradient update for the important neurons, and $\eta$ is the learning rate. 
For dimensions where the neuron is important in aligning batch-norm statistics with the backdoored model, we move in the direction of the gradient, thereby attempting to maximize the loss. 
For other dimensions, we follow the negative gradient and attempt to minimize the loss as usual.
The intuition behind this step is to strategically influence the gradient updates. 
By projecting the gradient updates only onto the coordinates that are not critical for aligning the fine-tuned model's activation distribution with that of the backdoored model, we effectively disrupt the backdoor trigger while preserving the model's performance on the main task. This approach ensures that the neurons associated with the backdoor task are updated in an unlearning manner, thereby mitigating the threat without compromising the overall accuracy and functionality of the model. \added{However, keeping this masked gradient on every iteration will introduce degradation on the benign task due to the connection between all neurons. Therefore, \method{} uses an alternative optimization strategy (lines 15-24 Algo.~\ref{algo:main}), where }
The full algorithm is presented in Algo.~\ref{algo:main}.
\replaced{We further provide the proof of convergence followed~ \autoref{thm:convergence} for \method{} and leave all proofs of theoretical development in the Appendix.}{One possible concern is that because \method{}'s fine-tuning process flips the sign of the updates every iteration, it might oscillate and/or not converge as all. This is however not the case, as shown in \autoref{thm:convergence}. All proofs of theoretical development for this paper will be presented in the appendix in the interest of space.}

\begin{theorem}
\label{thm:convergence}
Let $\theta_0$ be the initial pretrained weights (i.e., line 13 in \autoref{algo:main}). If the fine-tuning learning rate \replaced{is satisfied}{satisfies}:
$$
\eta<\left\Vert\frac{\partial^2\mathcal{L}(w,x)}{\partial w^2}\Big|_{\theta_0}\right\Vert_2^{-1},
$$
\noindent\autoref{algo:main} will converge.
\end{theorem}

%% file: algo/main-algo.tex
\begin{algorithm}[]
\footnotesize
\caption{\method}
\label{algo:main}
\SetAlgoLined
\Input{Fine-tuning data $\mathcal{D}_{ft}$, initial backdoor model $\theta_0$, total iteration $T$, pre-finetune total iteration $T^{\prime}$, pre-finetune learning rate $\eta^{\prime}$, learning rate $\eta$.
}
\Output{The fine-tuned model $\hat{\theta}$ after $T$ fine-tuning iterations;}

\textcolor{gray}{$\slash \ast $ Neuron mask generation $\ast \slash$}\\
Initialize $\tilde{\theta}$;\\

\For {$i \in\{1 \ldots T'\}$} {
    \For{$batch (x, y) \in \mathcal{D}_{ft}$}{
        $\mathcal{L}_{align} (x, \theta_0)$ \quad $\triangleright$ calculate alignment loss using Eq.~\ref{eqn:alignment}; \\
        $\mathcal{L}_{re} = \mathcal{L}_{\mathrm{ce}}\left(f_{\tilde{\theta}}\left(\boldsymbol{x}\right), y\right) + \alpha*\mathcal{L}_{align}$; \\ 
        $\tilde{\theta} = \tilde{\theta} - \eta^{\prime} \cdot \frac{\partial\mathcal{L}_{re}}{\partial \tilde{\theta}}$;
    }
}
    $\mathcal{N}_{m} = \text{argmax}_k \|\nabla_\theta \mathcal{L}_{re}(\tilde{\theta})\|_2 $;\\
\textcolor{gray}{$\slash \ast $ Activation-shift fine-tuning $ \ast \slash$}

$\mathbf{m} := [-1, 1]^{|\tilde{\theta}|}$, where $m_i = -1$ if $i \in N_m$ else 1;\\
$\theta_{0} = \theta_0 + \mathcal{N}(0, \sigma^2 I)$;\\
\For {iteration $t$ in $[1, \ldots, T]$} {
    \For{batch $(\mathbf{x}, \mathbf{y})$ in $\mathcal{D}_{ft}$}{
    $\theta_{\mathrm{t}}=\theta_{\mathrm{t}-1}-\eta \odot \frac{\partial \mathcal{L}_{\mathrm{ce}}\left(f_{\tilde{\theta}}\left(\boldsymbol{x}\right), y\right)}{\partial \theta_t}$;\\
    }
    \If{$t \mod 2 = 1$}{
    $\theta_{\mathrm{t}}=\theta_{\mathrm{t}-1}-\eta \odot \boldsymbol{m} \odot \frac{\partial \mathcal{L}_{\mathrm{ce}}\left(f_{\tilde{\theta}}\left(\boldsymbol{x}\right), y\right)}{\partial \theta_t}$;\\
    }
}
\Return $\theta_T$
\end{algorithm}

%% file: secs_R2/experiments.tex
\section{Experiments}
\subsection{Experimental Setups}
\fixed{As argued above, a robust backdoor purification method should have the ability to effectively mitigate the impact of multiple backdoor attacks, maintain stability across varying attacker power and fine-tuning conditions, and perform efficiently in multiple settings and architectures.} Therefore, we study four research questions to evaluate the efficiency of \method{} in purifying backdoor attacks targeting malware classifiers as follows.

\begin{enumerate}
    \item \textbf{RQ1:} 
    How well does \method{} purify backdoor attacks compared to related fine-tuning methods?
    \item \replaced{\textbf{RQ2: }Is PBP effective against backdoor attacks carried out by attackers with varying levels of strength?}{\textbf{RQ2: }Can \method{} purify backdoor attacks given different attacker power?} %
    \item \textbf{RQ3: }Can \method{} purify backdoor attacks stably under different fine-tuning assumptions?
    \item \textbf{RQ4: }How is \method{}'s efficiency and sensitivity to its hyperparameters and model architectures?
\end{enumerate}

We first discuss our experimental settings, baselines, and metrics below.  Then, we answer each research question in turn.

\noindent\textbf{Backdoor attacks and settings. }
In this work, we focus on two state-of-the-art backdoor attack strategies designed for malware classifiers, which are \textit{Explanation-guided}~\cite{severi2021explanation} and \textit{Jigsaw Puzzle}~\cite{yang2023jigsaw}. 
Regarding the former attack, experiments are conducted mainly on the EMBER-v1~\cite{anderson2018ember} dataset. 
To study the latter attack, we conduct experiments on the Android malware dataset sampled from AndroZoo~\cite{allix2016androzoo}. 
The EMBER-v1 dataset consists of 2,351-dimensional feature vectors extracted from 1.1 million Portable Executable (PE) files for Microsoft Windows~\cite{severi2021explanation}. 
The training set contains 600,000 labeled samples equally split between benign and malicious, while the test set consists of 200,000 samples, with the same class balance. 
All the binaries categorized as malicious were reported as such by at least 40 antivirus engines on VirusTotal~\cite{severi2021explanation}.
For the Jigsaw Puzzle attack, we reuse a sampled dataset from the AndroZoo collection of Android applications, following the setting of the original paper~\cite{yang2023jigsaw}. 
The feature vectors for the Android apps were extracted using Drebin~\cite{arp2014drebin}. 
Each feature in the Drebin feature vector has a binary value, where ``1'' indicates that the app contains the specific feature (e.g., an API, a permission), and ``0'' indicates that it does not. The final dataset consists of 149,534 samples, including 134,759 benign samples and 14,775 malware samples. 
The dataset covers 400 malware families, with the number of samples per family ranging from 1 to 2,897, with an average size of 36.94 and a standard deviation of 223.38. \fixed{Both datasets are used for the binary classification task, and the backdoor task is classifying the malware samples given the presence of trigger as \replaced{``benign''.}{``benign.''}}

\input{tabs/ndss_r2/merge_ft_result}
\noindent\textbf{Training configurations. }
We consider several training configurations to ensure our approach will generalize across multiple indicative scenarios. 
For EMBER, we use the default training and testing sets of this dataset and reproduce the watermark for \textit{Explanation-guided} attacks of the original paper using the Combined strategy, which is a greedy algorithm to conditionally select new feature dimensions and their values such that those values are consistent with existing goodware-oriented points in the attacker’s dataset to generate a backdoor watermark. 
For AndroZoo, we follow the original work by Yang et al.~\cite{yang2023jigsaw} and split the dataset for training (67\%) and testing (33\%). 
We reproduce the backdoor attack on feature space with the ``Kuguo'' family as the targeted set, which is the largest family in this dataset (2,845 samples). 
Without further mentioning, for both datasets, we reserve 10\% from the training data for fine-tuning. \replaced{We leave the detailed settings for training and fine-tuning parameters, and the evaluation of backdoored models before fine-tuning in the Appendix~\ref{appendix:training_detail}.}{We detailed training parameters are left in Appendix~\ref{appendix:training_detail}.
The performance of trained models for both tasks is shown in Table~\ref{tab:model-acc}.}

\vspace{-2mm}
\noindent\textbf{Baselines.} 
We compare our method with five fine-tuning defenses used in backdoor purification tasks: Vanilla FT, Vanilla LP, FE-tuning, FT-init, and Feature-shift Fine-tuning. Since we use an MLP model architecture in all experiments, we consider the final layer as the classifier $\theta_w$ and all the previous layers as the feature extractor component $\theta_{\Phi}$. Specifically, 

\vspace{-2mm}
    \noindent --- Vanilla Fine-Tuning (vanilla FT): In this approach, we fine-tune the entire model, updating all parameters including both the feature extractor ($\theta_{\Phi}$) and the linear classifier ($\theta_w$). This means that the entire model architecture undergoes learning and adaptation to new data.\\
    --- Linear Probe (LP): Here, only the parameters of the linear classifier $\theta_w$ are fine-tuned while keeping the feature extractor's parameters ($\theta_{\Phi}$) unchanged and frozen. This method assesses how well the pre-trained features can linearly separate the data without altering the learned feature representations. \\
    --- Feature Extractor Tuning (FE-tuning)~\cite{qin2023revisiting}: For FE-tuning, the parameters in the model's ``head'' ($\theta_{w}$) are re-initialized and then frozen. The rest of the model, i.e., the feature extractor $\theta_{\Phi}$, is then fine-tuned. This approach is designed to update the core representation abilities of the model while keeping the decision layer fixed.\\
    --- Fine-Tuning with Initialization (FT-init)~\cite{min2024towards}: In FT-init, the linear head is randomly re-initialized, and the entire model architecture, including both the feature extractor and the linear head, is fine-tuned. This combines a fresh start for the linear classifier with an opportunity to further adapt the feature extractor to new tasks.\\
    --- Feature Shift Tuning (FST)~\cite{min2024towards}: In this fine-tuning method, the \replaced{authors}{author} \replaced{uses}{suggested using} feature shifts by actively deviating the classifier weights $\theta_{\Phi}$ from the originally compromised weights. 

\noindent\textbf{Evaluation metrics. }
\replaced{We use C-Acc, ASR, and DER to evaluate the purification efficacy. The objective of a defender is to maximize C-Acc and DER while minimizing ASR at the same time. These three metrics are defined in Sec.~\ref{sec:background}.}{We consider two evaluation metrics: (1) Clean Accuracy (C-Acc), the prediction accuracy of clean samples, and (2)  Attack Success Rate (ASR) the prediction accuracy of poisoned samples to the target class.
A lower ASR indicates a better defense performance as it means our approach successfully reduces the impact of the backdoor attack. In addition, we use DER in \textbf{RQ4}, a metric measuring how well a method balances between reducing ASR and maintaining C-Acc. These three metrics are defined in Sec.~\ref{sec:background}.}

\begin{figure}[t]
\centering
    \begin{subfigure}
    {0.45\columnwidth}
    \includegraphics[width=1.0\textwidth]{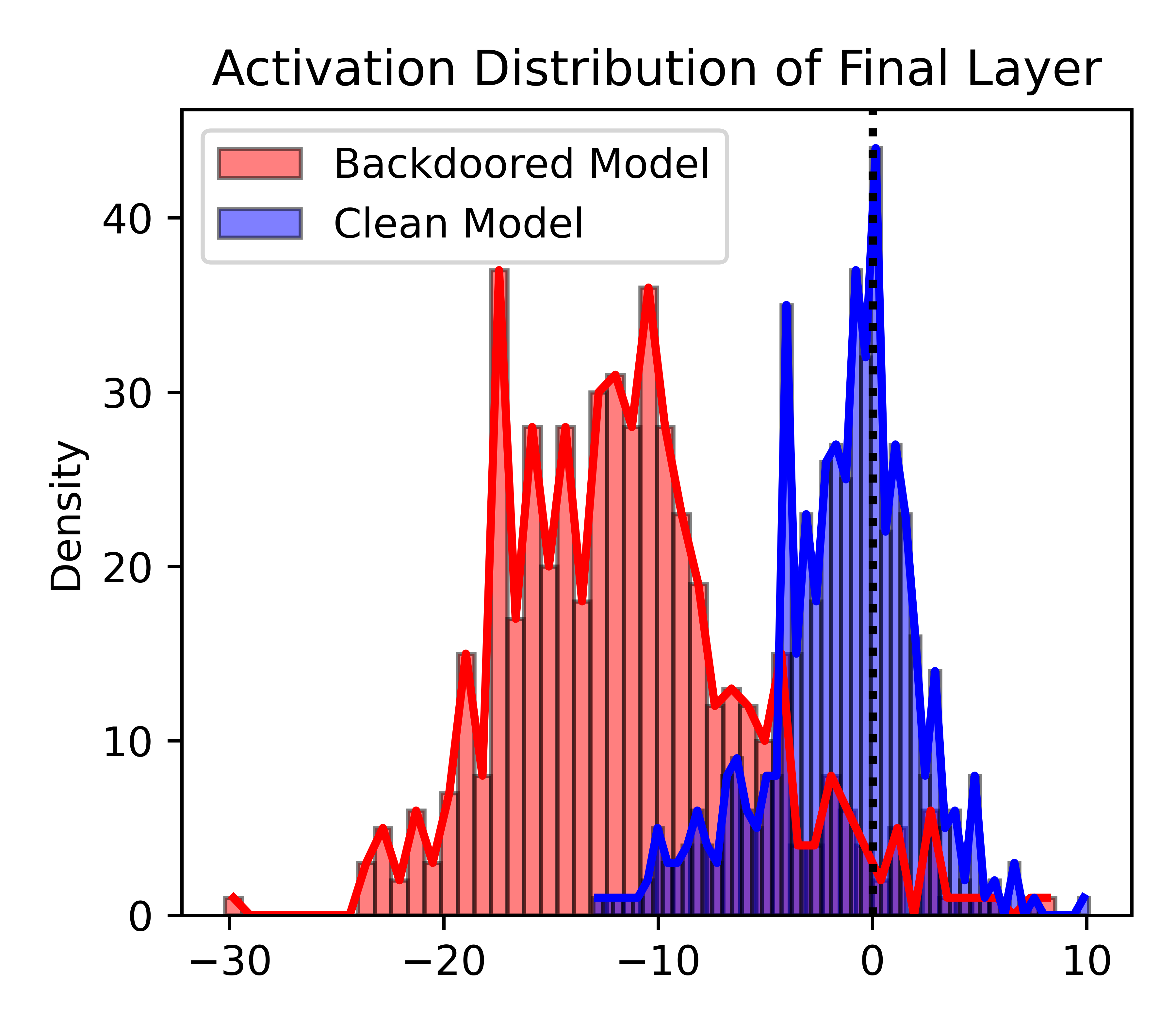}
    \caption{Clean and Backdoored Models}
    \end{subfigure}
    \begin{subfigure}
    {0.45\columnwidth}
    \includegraphics[width=1.0\textwidth]{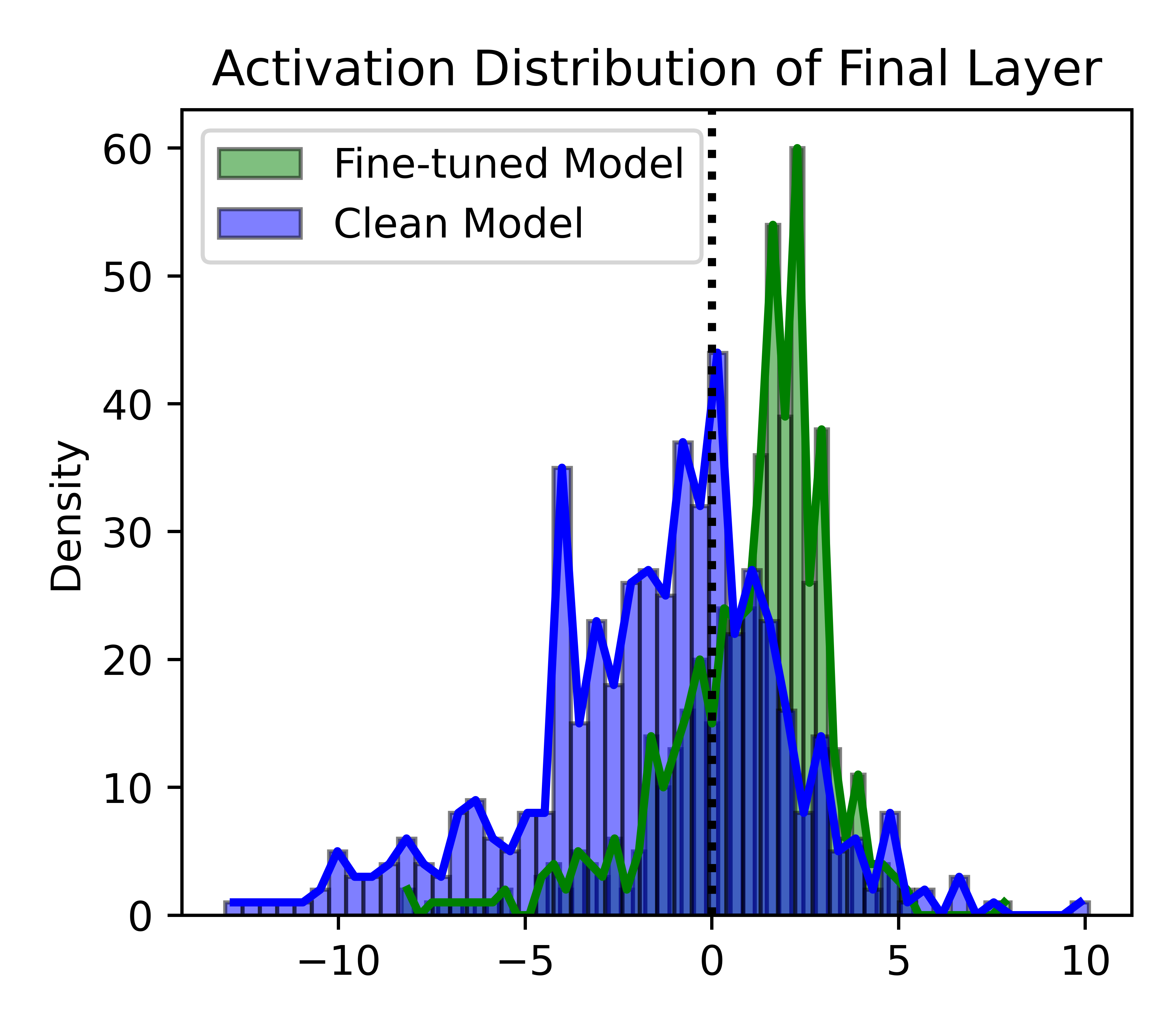}
    \caption{Clean and \method{}'s Fine-tuned Models}
    \end{subfigure}
    \caption{Final layer's activation of \textbf{non-family-targeted} backdoor attacks \added{on triggered samples}.}
\label{fig:ember-attack}
\vspace{-0.5cm}
\end{figure}
\begin{figure}[]
    \centering
    \begin{subfigure}
    {0.45\columnwidth}
    \includegraphics[width=1.0\textwidth]{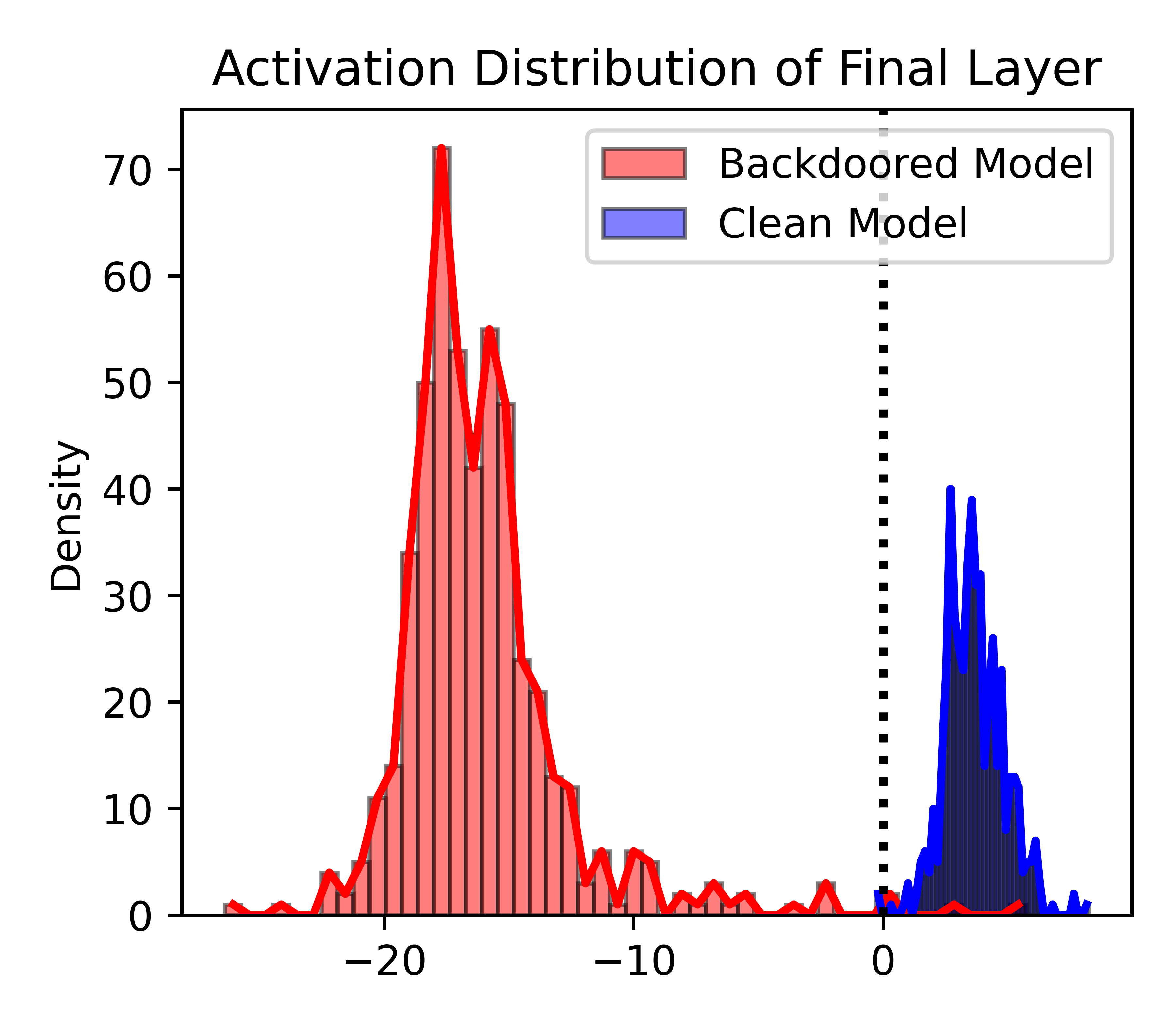}
    \caption{Backdoored Model}
    \end{subfigure}
    \begin{subfigure}
    {0.45\columnwidth}
    \includegraphics[width=1.0\textwidth]{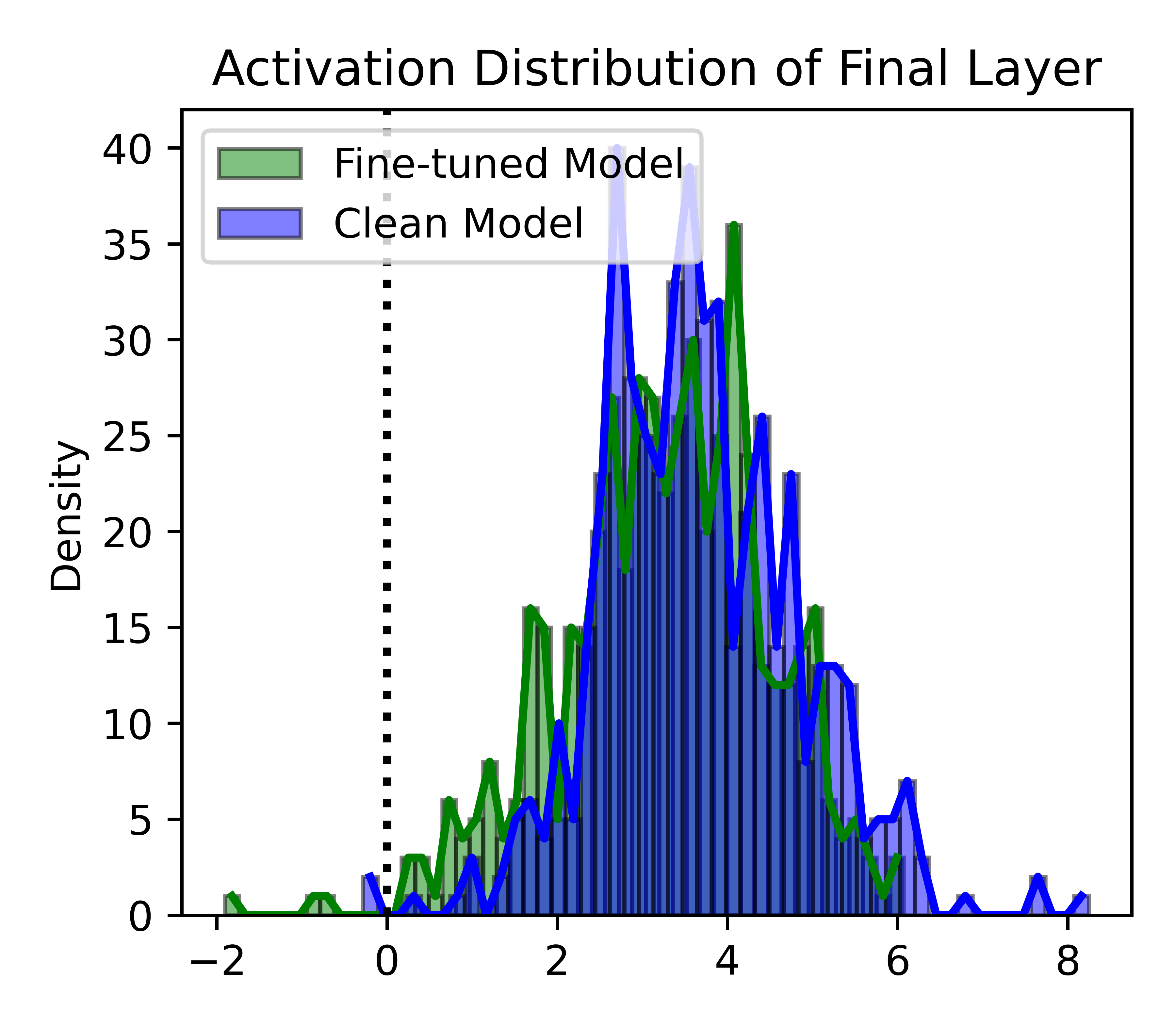}
    \caption{\method{}'s Fine-tuned Model}
    \end{subfigure}
    \caption{Final layer's activation of \textbf{family-targeted} backdoor attacks \added{on triggered samples}.}
    \vspace{-0.5cm}
    \label{fig:family-attack-act}
\end{figure}
\vspace{-2mm}
\subsection{Experimental Results}
\subsubsection{\textbf{RQ1: }Can \method{} purify the backdoor effectively on different backdoor attack strategies, and to what extent, compared to related fine-tuning methods?} 

We present the C-Acc and ASR of compared methods in 
\added{Table~\ref{tab:ember-apg-0.1}}
with a fine-tuning size of 10\% for Explanation-guided and Jigsaw attacks, respectively. 
The family-targeted backdoor attack (i.e., Jigsaw Puzzle) is more fragile during the fine-tuning phases, compared to a non-family backdoor attack. 
In \added{Table~\ref{tab:ember-apg-0.1}}, FT-init, FE-tuning, and FST can mitigate the backdoor effect on AndroZoo data up to 40\% in the best case. 
However, when addressing non-family targeted backdoor attacks launched on EMBER, all of the baselines fail to reduce the ASR. 
Our approach is the only one that almost perfectly mitigates the backdoor effect --- i.e., ASR decreases to nearly zero for family-targeted and to 15.44\% in the case of non-family-targeted backdoor attacks. 

\begin{figure*}[h]
    \centering
    \begin{subfigure}
    {0.16\textwidth}
    \includegraphics[width=1.0\textwidth]{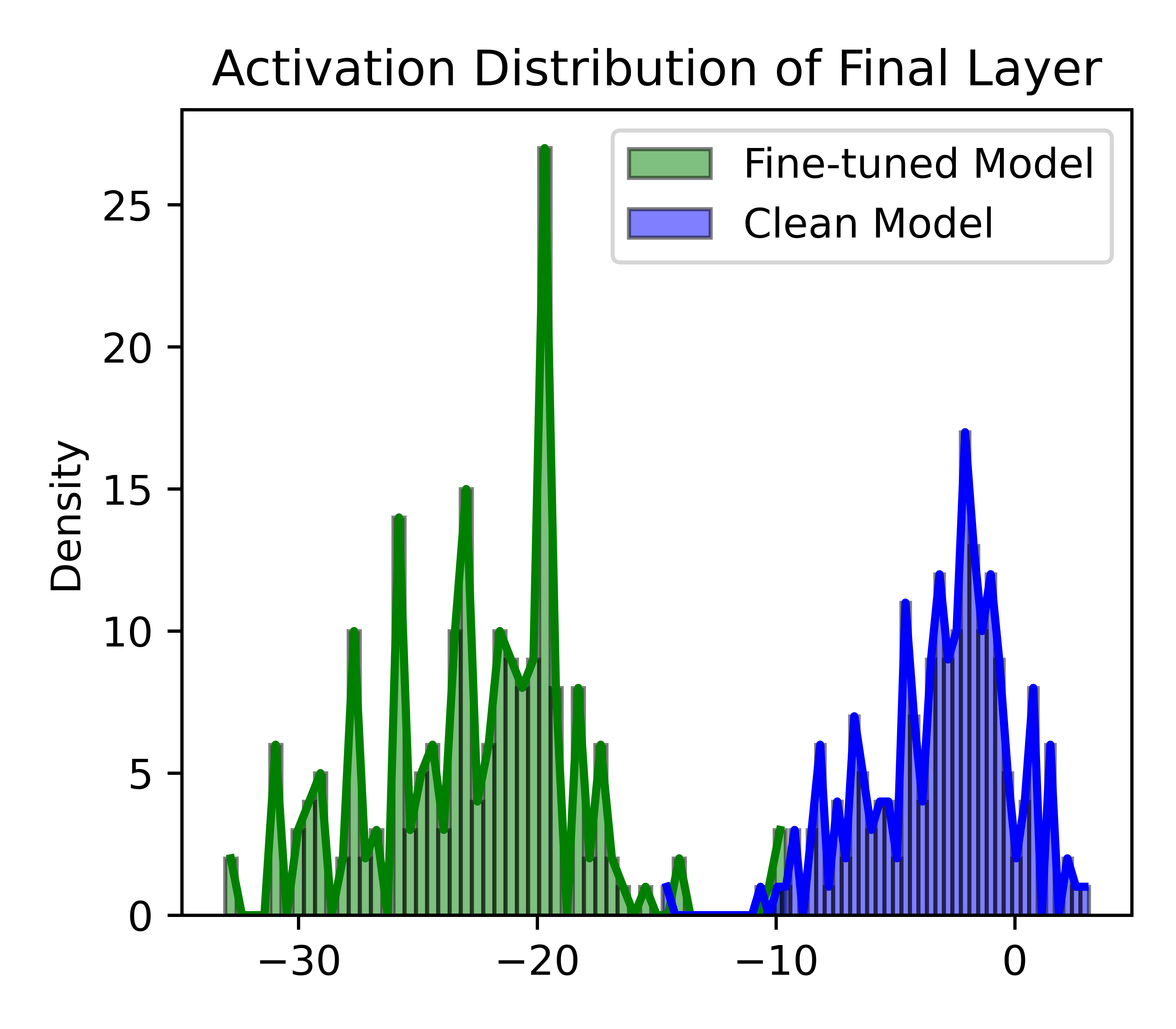}
    \caption{FT}
    \label{fig:illus-transfer}
    \end{subfigure}
    \begin{subfigure}
    {0.16\textwidth}
    \includegraphics[width=1.0\textwidth]{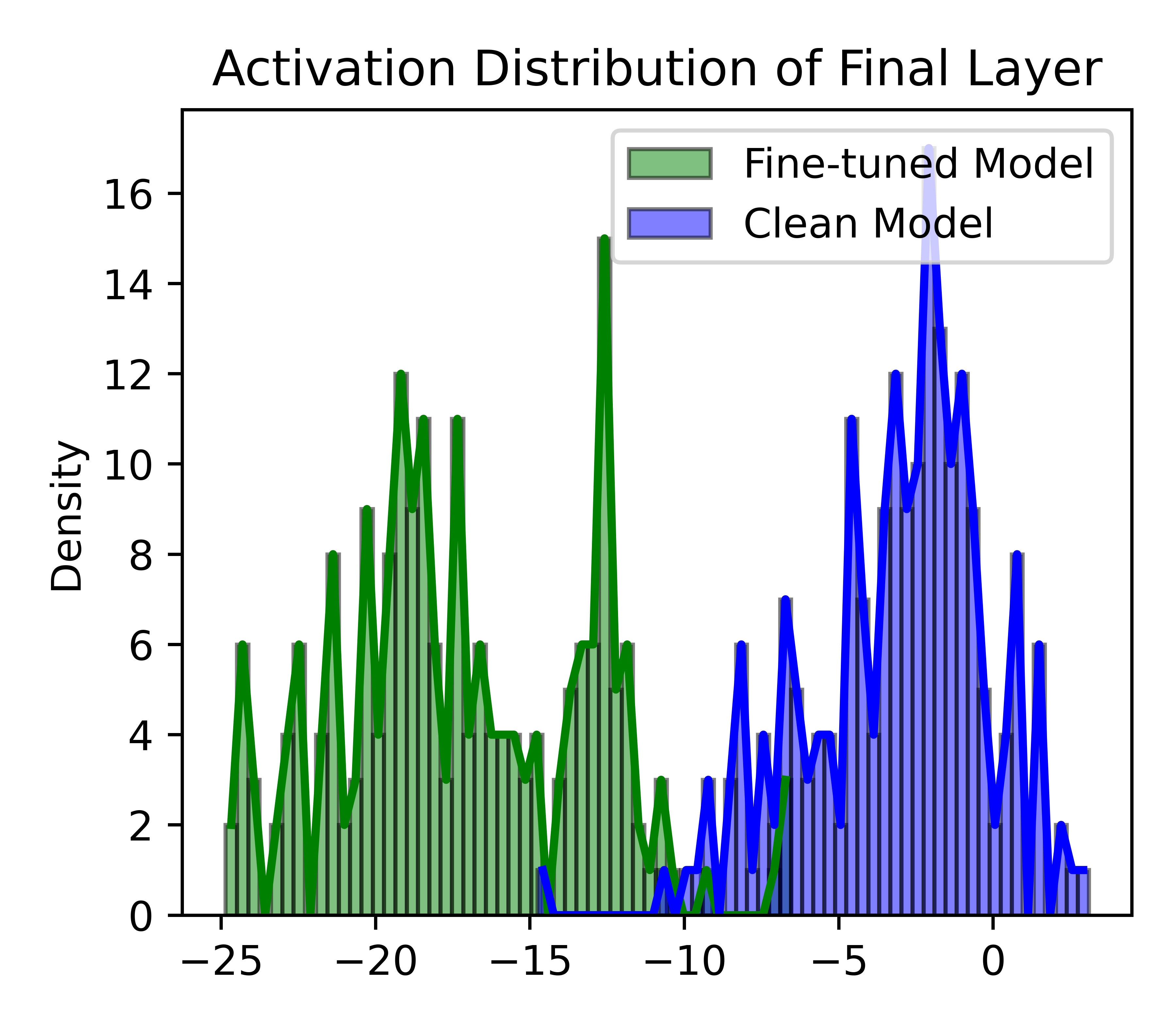}
    \caption{FT-Init}
    \label{fig:illus-transfer}
    \end{subfigure}
    \begin{subfigure}
    {0.16\textwidth}
    \includegraphics[width=1.0\textwidth]{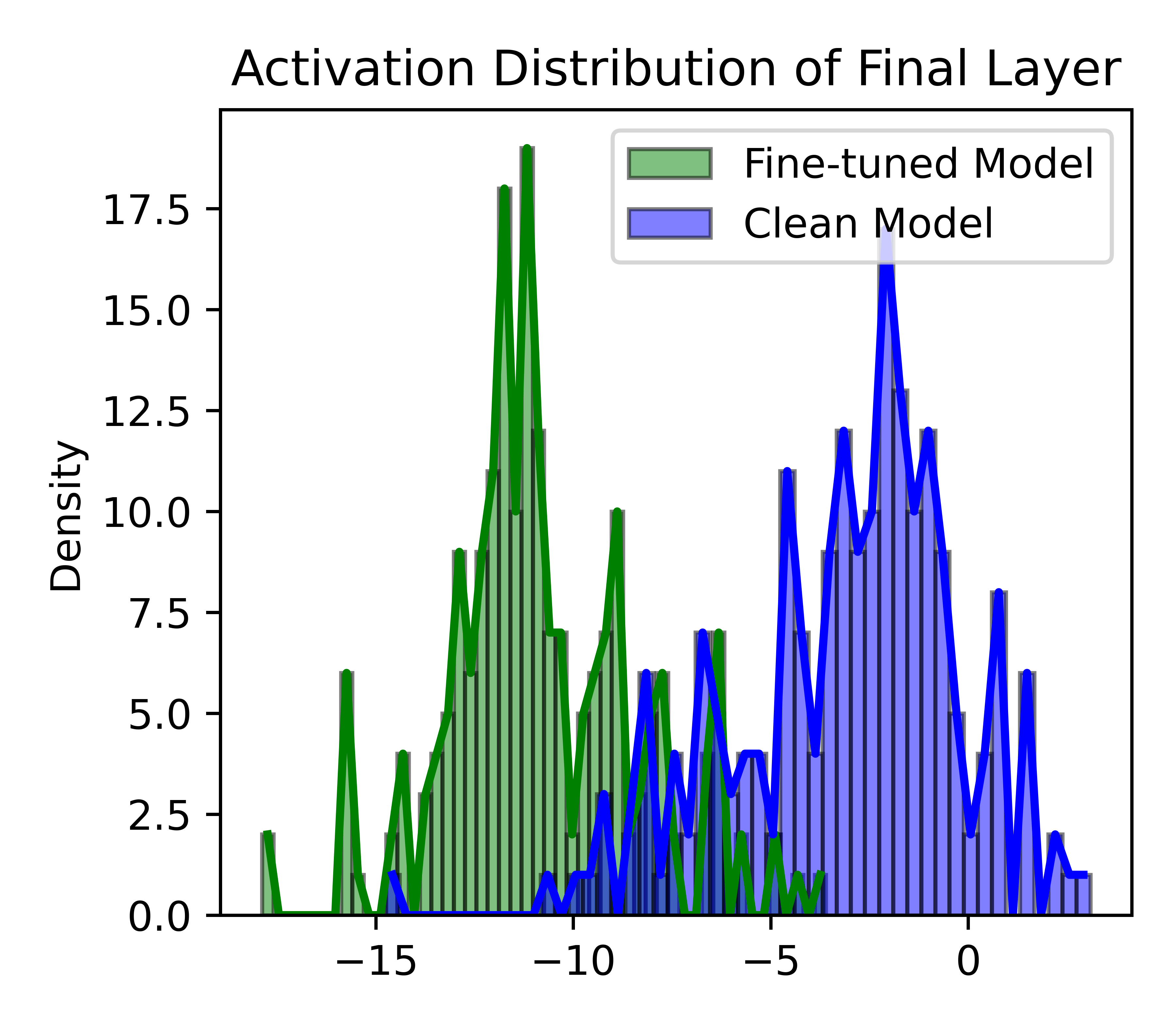}
    \caption{LP}
    \label{fig:illus-transfer}
    \end{subfigure}
         \begin{subfigure}
    {0.16\textwidth}
    \includegraphics[width=1.0\textwidth]{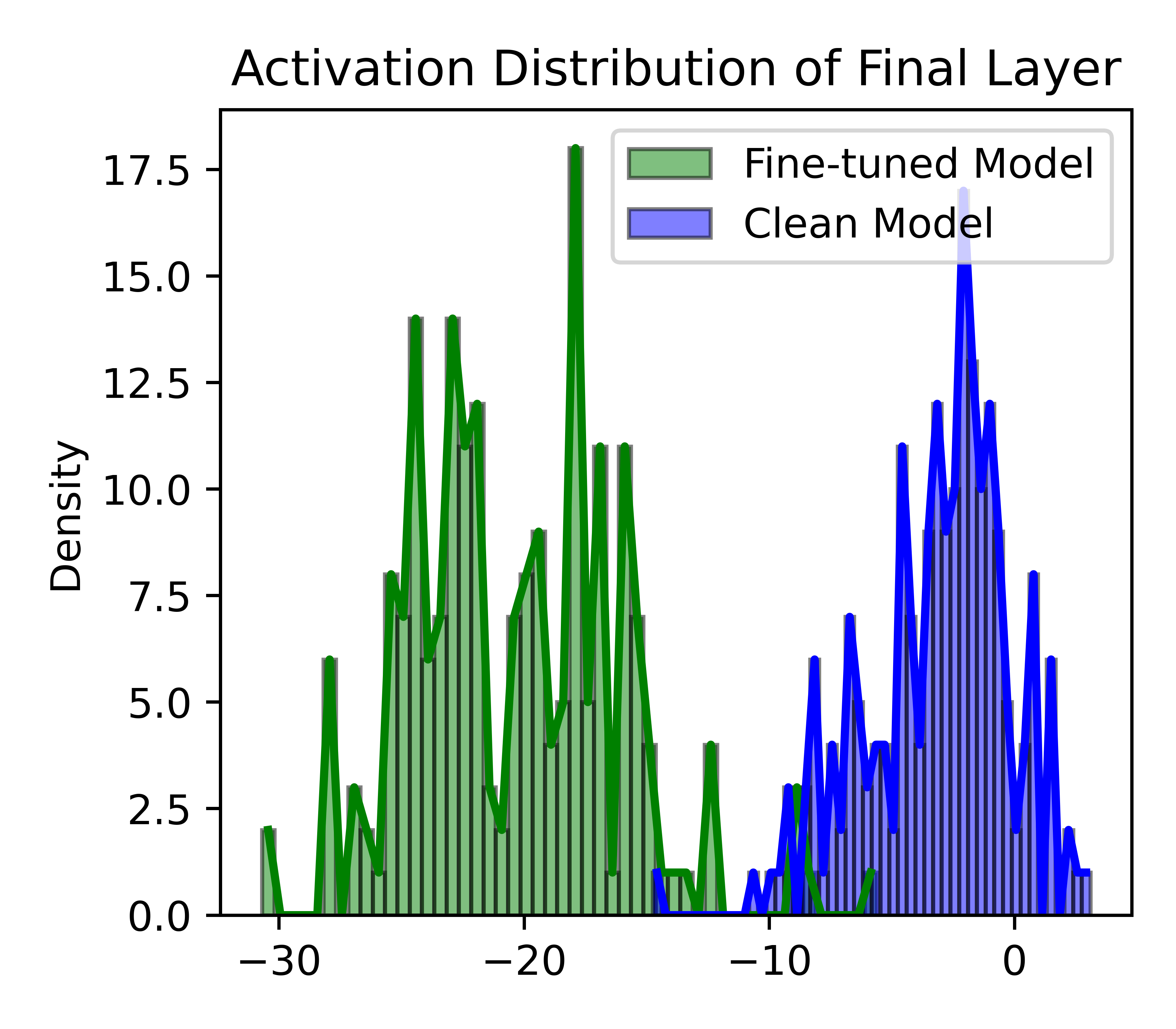}
    \caption{Fe-Tuning}
    \label{fig:illus-transfer}
    \end{subfigure}
     \begin{subfigure}
    {0.16\textwidth}
    \includegraphics[width=1.0\textwidth]{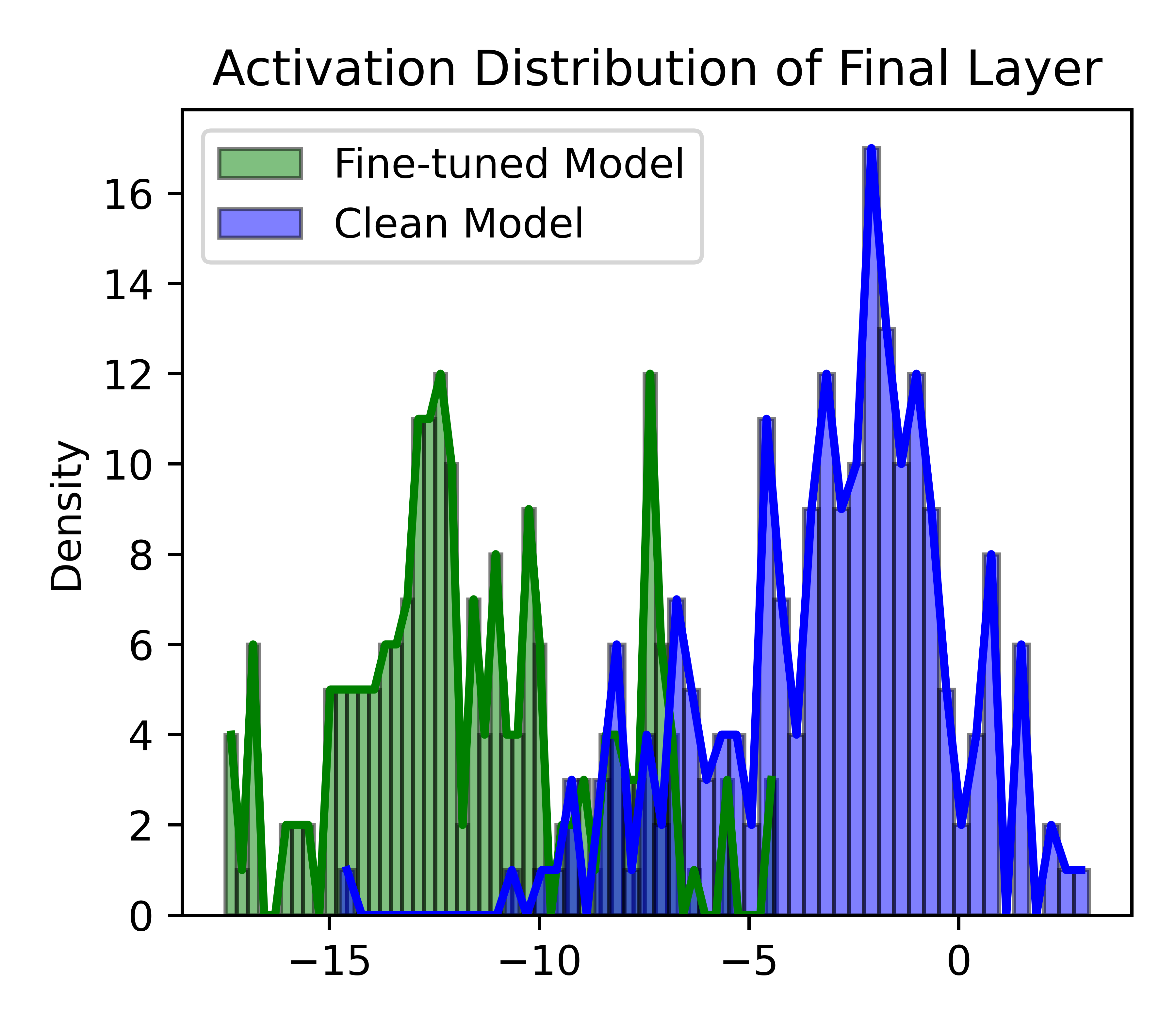}
    \caption{FST}
    \label{fig:illus-transfer}
    \end{subfigure}
     \begin{subfigure}
    {0.16\textwidth}
    \includegraphics[width=1.0\textwidth]{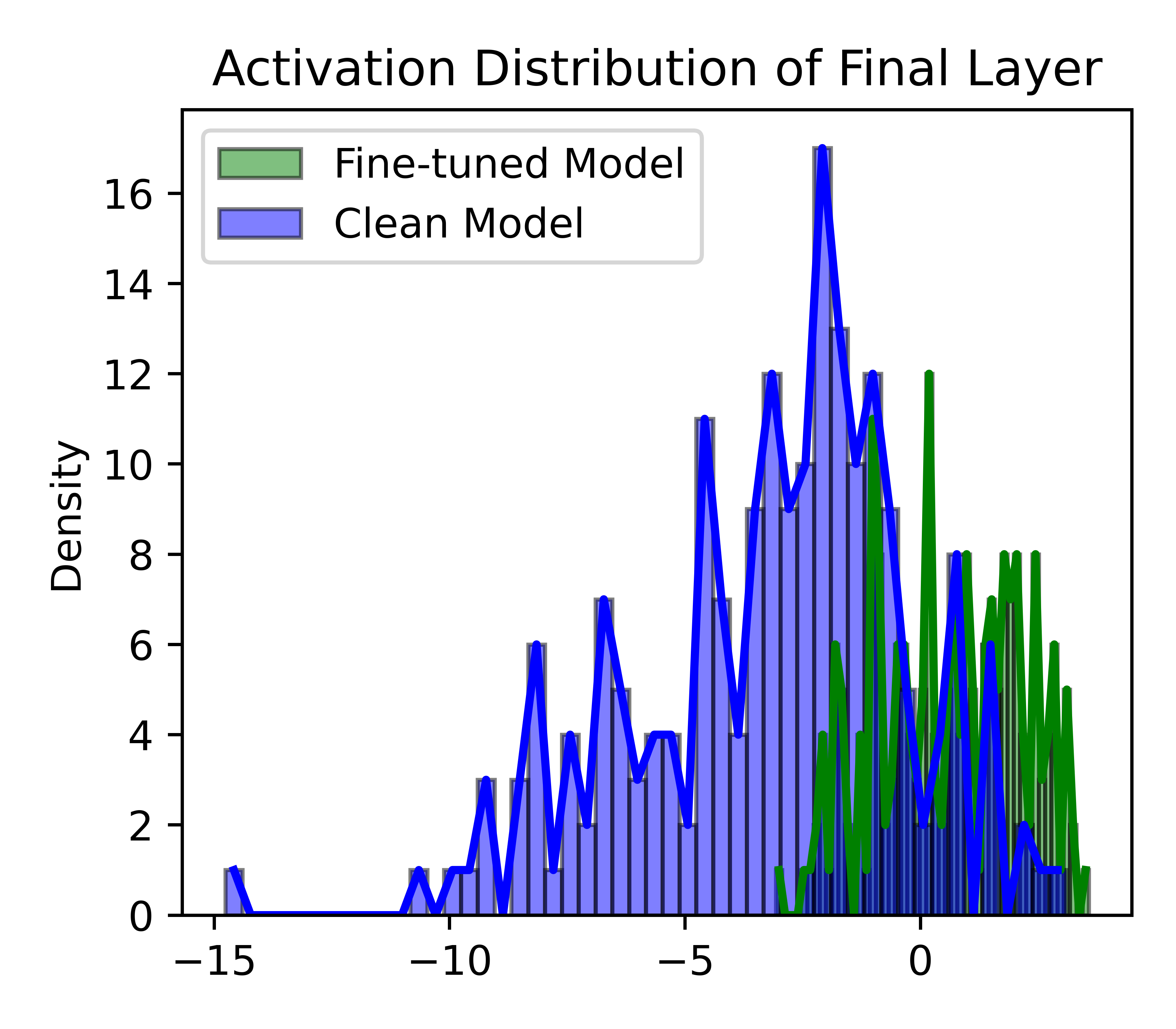}
    \caption{Ours}
    \label{fig:pbp-final-act}
    \end{subfigure}
    \caption{Comparison of model activation of different fine-tuning methods and a clean model on targeted malware samples on \textbf{EMBER} dataset.}
    \label{fig:decision-ember}
    \vspace{-0.5cm}
\end{figure*}
\begin{figure*}[h]
    \centering
    \begin{subfigure}
    {0.16\textwidth}
    \includegraphics[width=1.0\textwidth]{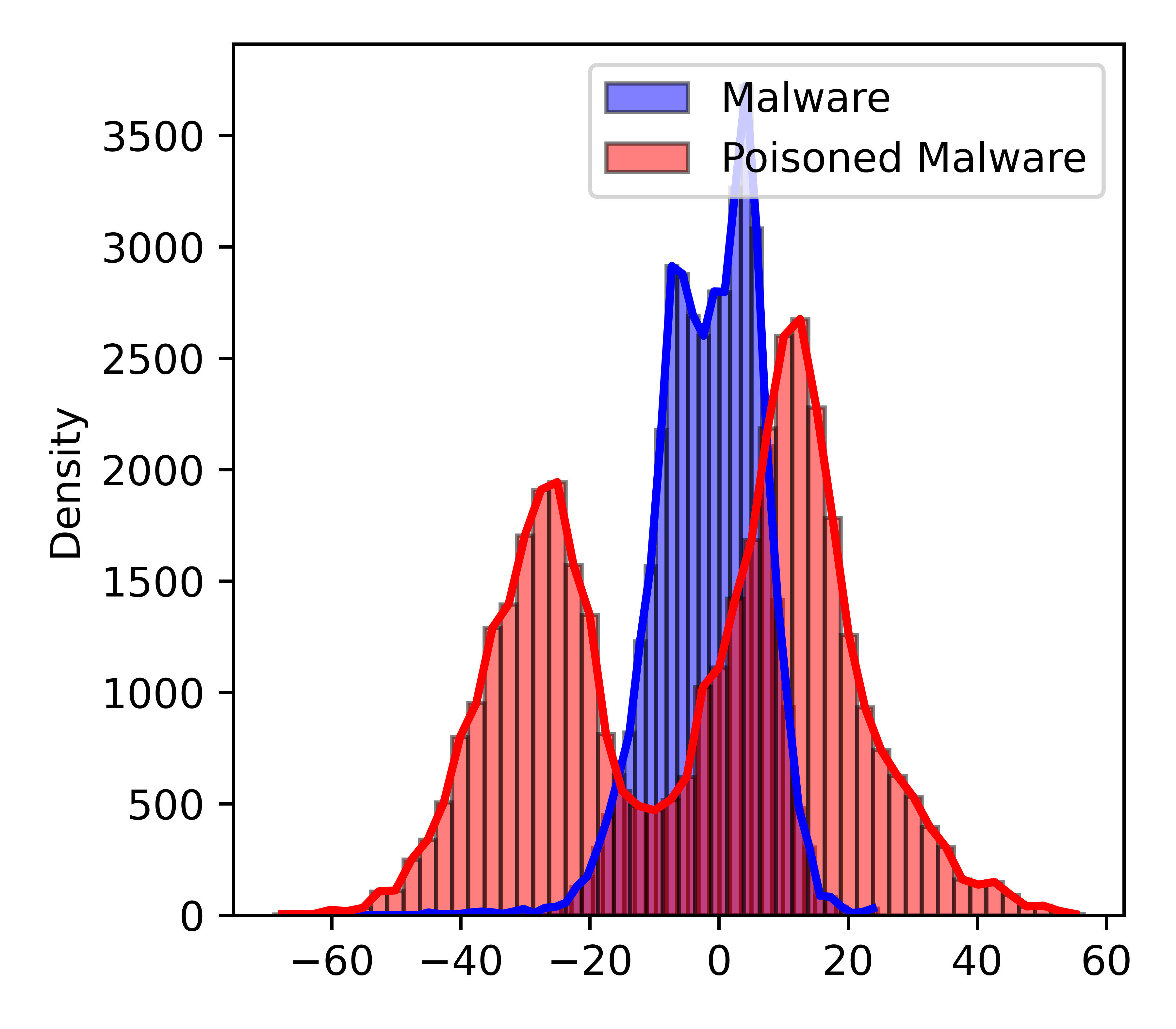}
    \caption{FT}
    \end{subfigure}
    \begin{subfigure}
    {0.16\textwidth}
    \includegraphics[width=1.0\textwidth]{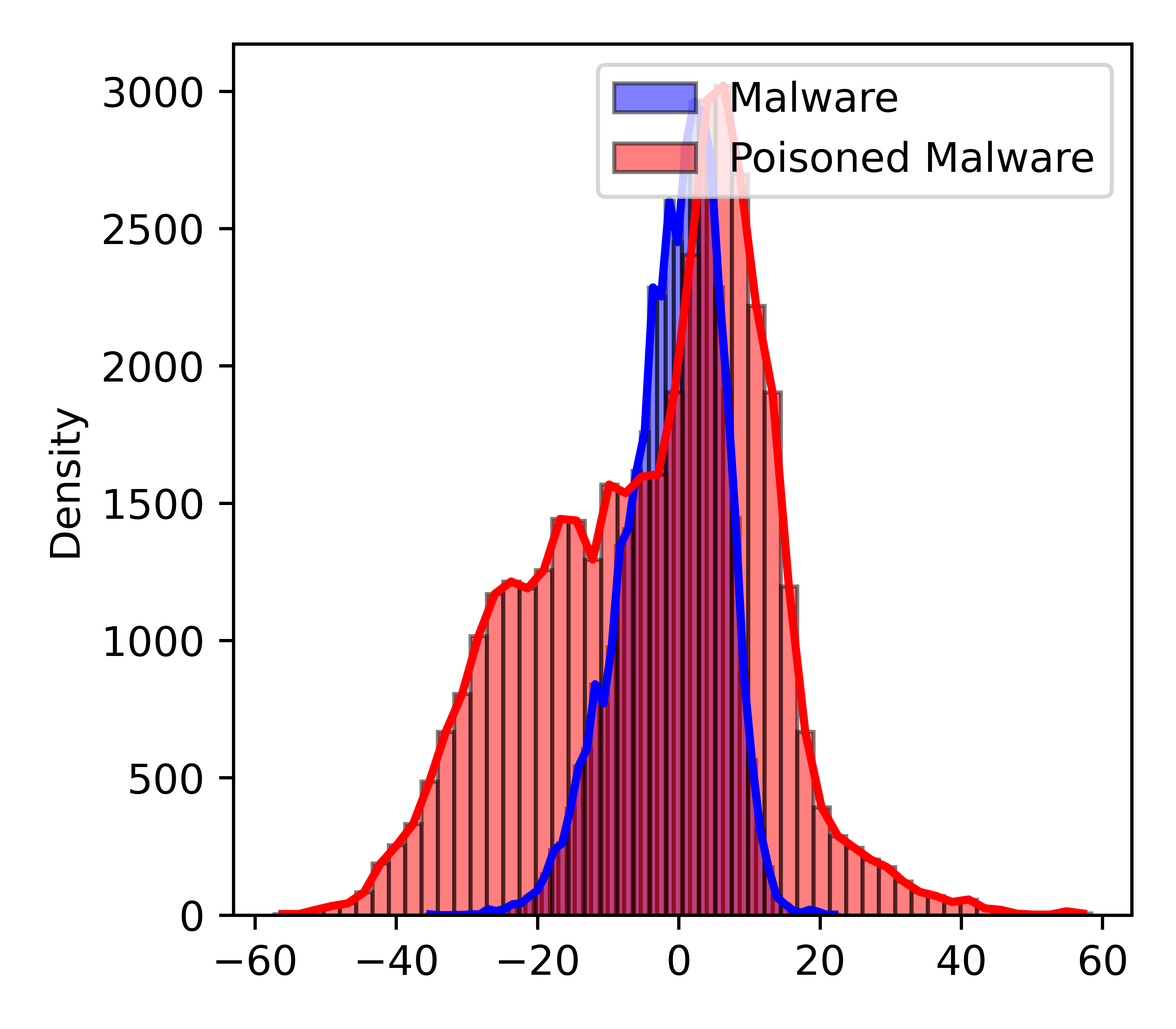}
    \caption{FT-Init}
    \end{subfigure}
    \begin{subfigure}
    {0.16\textwidth}
    \includegraphics[width=1.0\textwidth]{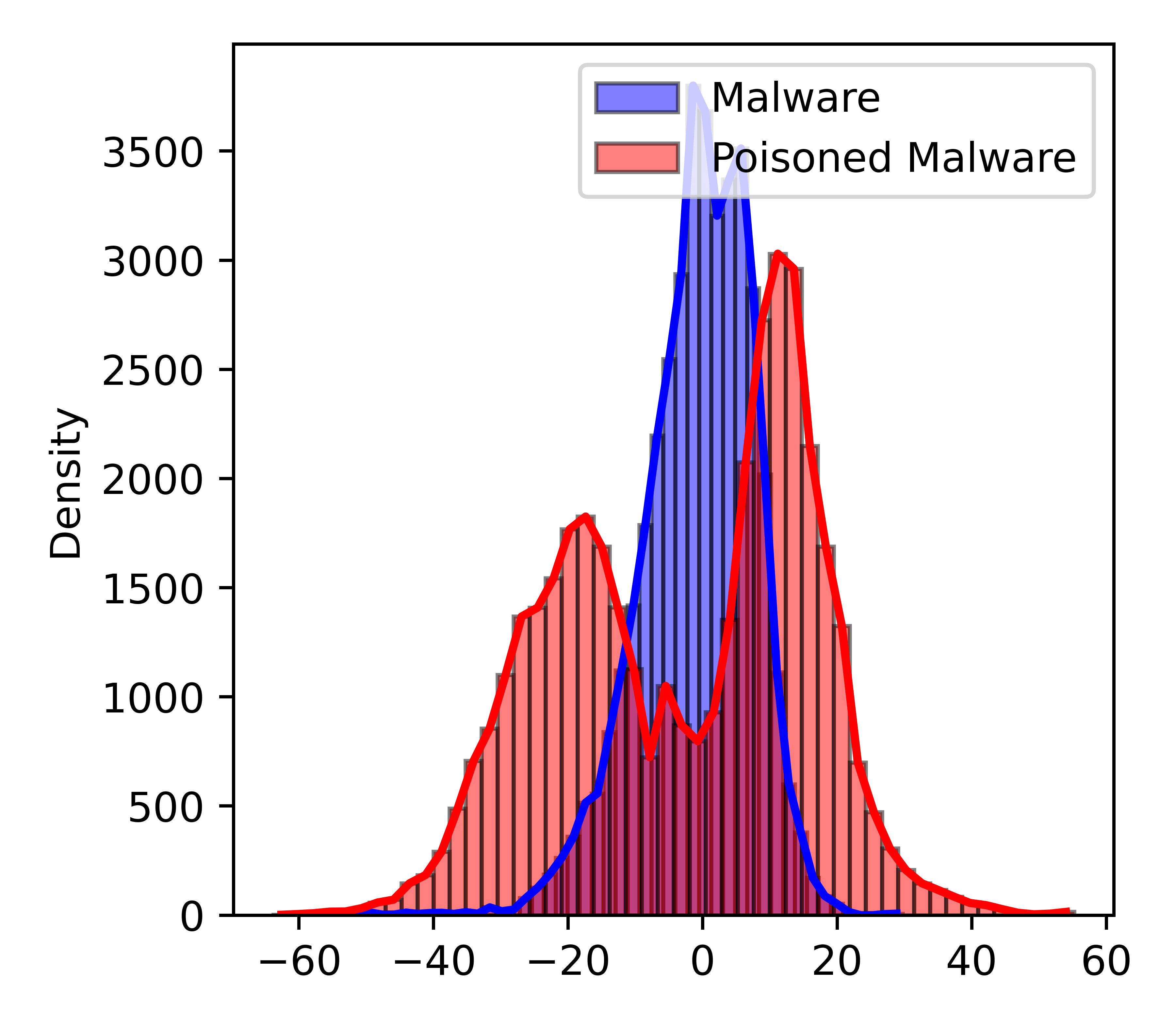}
    \caption{LP}
    \end{subfigure}
         \begin{subfigure}
    {0.16\textwidth}
    \includegraphics[width=1.0\textwidth]{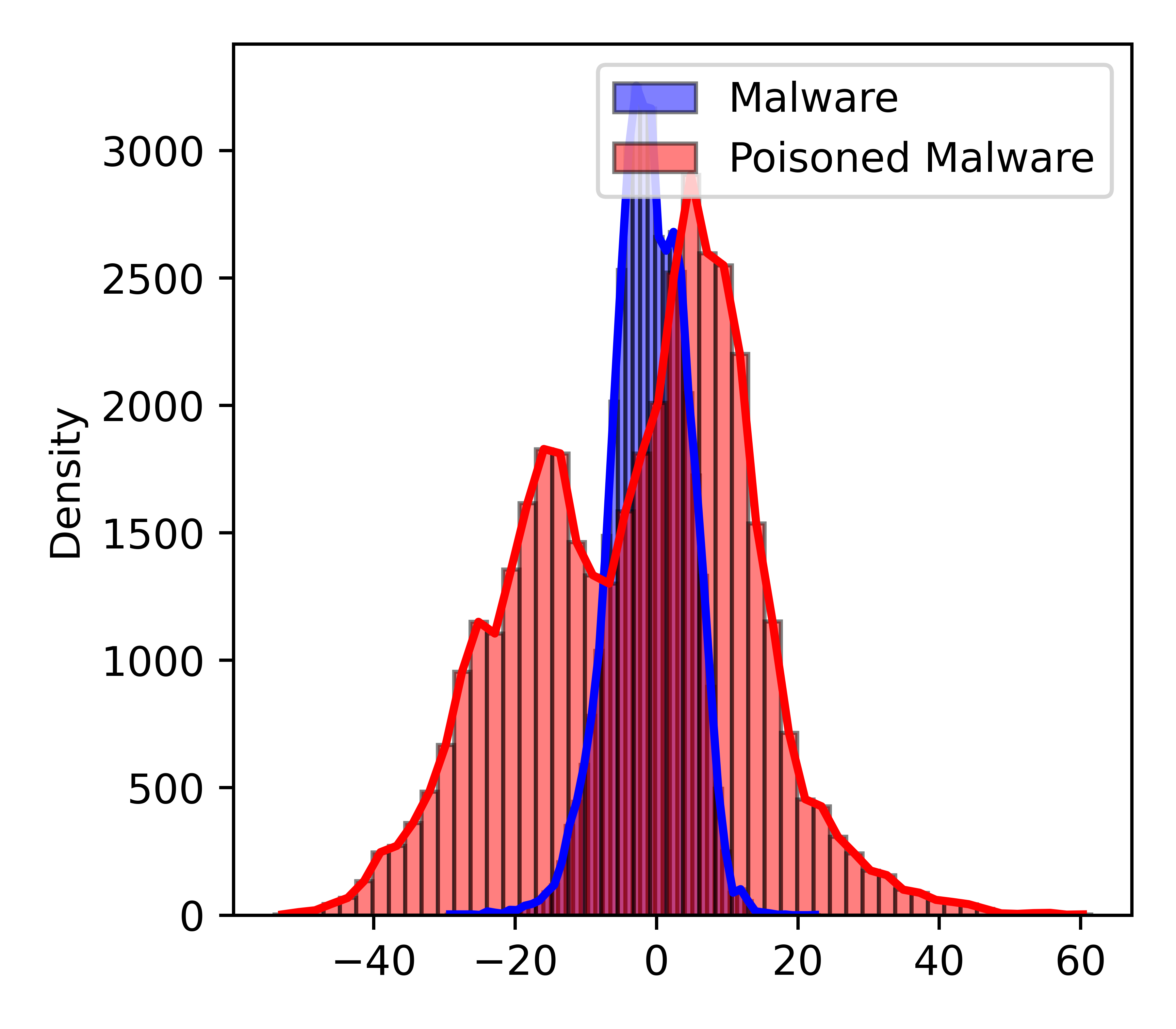}
    \caption{Fe-Tuning}
    \end{subfigure}
     \begin{subfigure}
    {0.16\textwidth}
    \includegraphics[width=1.0\textwidth]{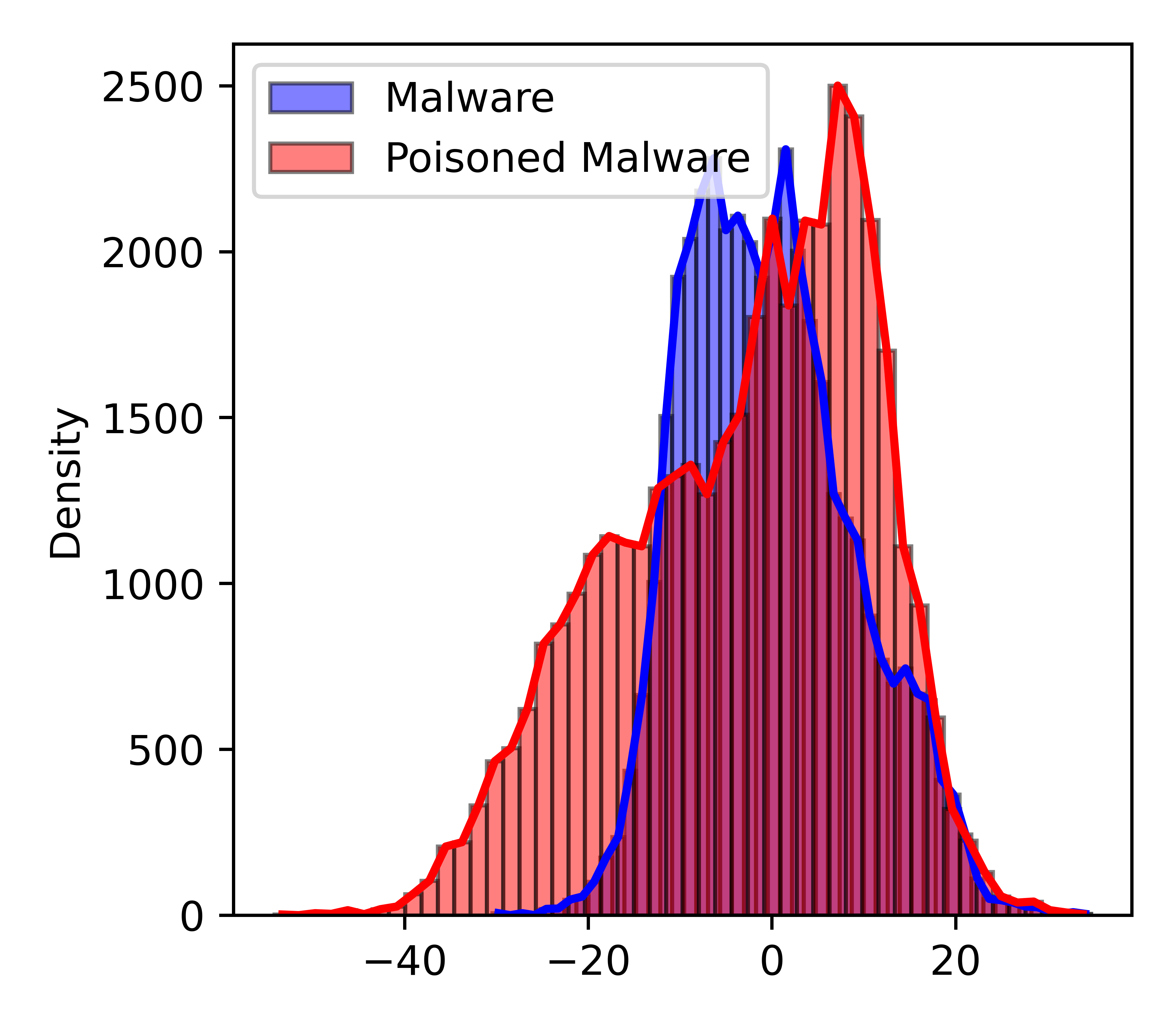}
    \caption{FST}
    \end{subfigure}
     \begin{subfigure}
    {0.16\textwidth}
    \includegraphics[width=1.0\textwidth]{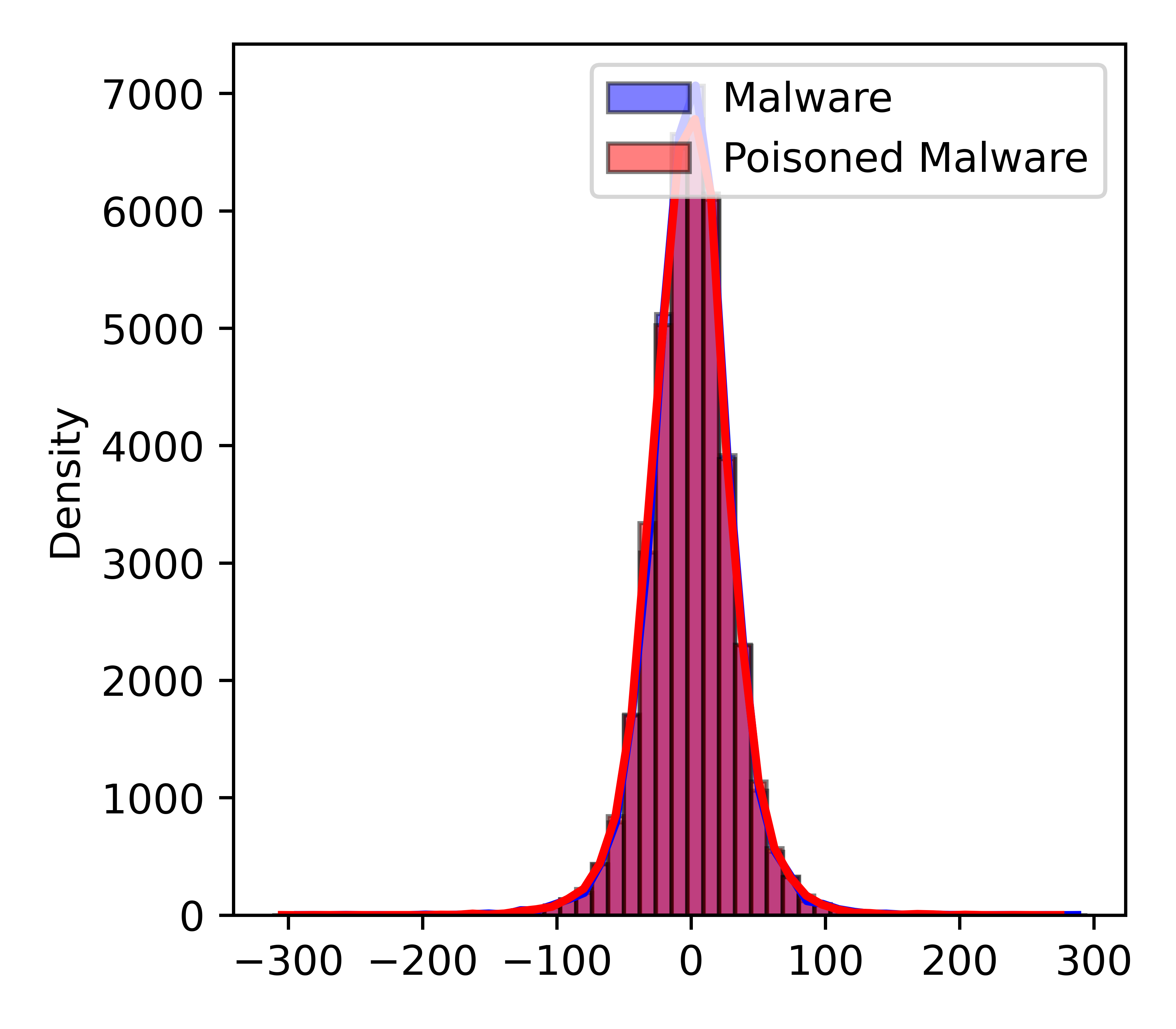}
    \caption{Ours}
    \end{subfigure}
    \caption{Model activation of different fine-tuning methods on targeted malware samples with and without the trigger on \textbf{EMBER} dataset on the final layer.}
    \label{fig:act-ember-finetunes}
    \vspace{-1cm}
\end{figure*}

\noindent\textbf{Non-family-targeted backdoor attack. }
In this attack, the adversary leverages SHAP values of a model trained on the same dataset to identify the most influential features in the feature space and then chooses values with the highest positive SHAP values to maximize the backdoor impact. 
\replaced{Therefore, even a clean model not trained on poisoned data can still exhibit a high ASR}{Therefore, even with a clean model that has not been trained on poisoned data, still has a substantially high ASR} --- up to 90\%. 
We show the model behavior of both the clean and backdoored models on the same set of poisoned malware samples in Fig.~\ref{fig:ember-attack}. 
Even with a clean model, \deleted{in the last layer of the model,}almost 90\% of the poisoned samples are activated below the decision threshold, then will be misclassified into ``benign.'' 
If the model is trained with poisoned samples combined with the learned watermark, the activation of the final layer is shifted drastically toward the ``benign'' decision.
After fine-tuning, our method \method{} shifts the model decision on these poisoned samples toward the ``malware'' decision, compared to the clean model without any trigger.

\noindent\textbf{Family-targeted backdoor attack.} In this attack, the adversary of the malware author only protects a specific target family $T$ rather than families as in the earlier attack.
In our experiment with this line of attacks, we consider ``kuogu'' as the targeted family.
\replaced{In contrast to non-family-targeted attacks, a clean model yields an ASR of zero, as the backdoor requires a trigger mask combined with benign data—something absent from the original training set.}{In contrast to the non-family-targeted attack, the ASR is zero for a clean model.
The reason is that this backdoor needs the presence of a combination of the trigger mask on the benign training data, and this mask may not appear on the training from the original.}
If the model is trained with poisoned samples combined with the learned watermark, the activation of the final layer is shifted drastically toward the ``benign'' decision.
\replaced{After fine-tuning, our \method{} shifts these decisions toward the clean model.}{After fine-tuning, our \method{} approach shifts the model decision on these poisoned samples toward that of the clean model.}

To further demonstrate the performance of different fine-tuning methods, we present the final layer activation of the model after fine-tuning using each method compared to the clean version on \added{Fig.~\ref{fig:decision-ember} and Fig.~\ref{fig:act-ember-finetunes}} \deleted{Fig.~\ref{fig:decision-ember} and Fig.~\ref{fig:decision-jigsaw}}. 
It is demonstrated that the activations of poisoned samples remain shifted toward ``benign'' even after fine-tuning, except for our method~\method{}. The rationale for this phenomenon aligns with our observation about activation distribution shift, i.e., other baselines \added{based on reinitialization and shifting model parameters} fail in deviating the activation distribution of backdoor neurons on triggered malware samples toward that of non-triggered malware samples (cf. in Fig.~\ref{fig:act-ember-finetunes} \deleted{and Fig.~\ref{fig:act-jigsaw-finetunes}}).
This is explained by the fact that these two attack strategies include knowledge of the adversary via a portion of the training data they control, and the trigger is optimized by observing the model learning process on these samples. 
Therefore, during the fine-tuning process, if the distribution of the data exhibits properties previously known by the adversary,
the trigger can still cause the victim model to malfunction. %
Empirically, even if only the final layer of the victim model is initialized, the fine-tuning process cannot make it forget the backdoor properties. 
This is presented via the activation of the backdoor neurons if the targeted sample combined with the trigger deviates significantly compared to the original malware samples.
Our method, \method{}, is the only fine-tuning method that can help fix the activation distribution of the model and help it activate indiscriminately given the malware and a corresponding triggered malware sample.
\replaced{In conclusion}{To this end}, \method{} demonstrates superior performance in purifying backdoor attacks for both family-targeted and non-family-targeted backdoor attacks.

\begin{figure}[h!]
    \centering
    \vspace{5mm}
    \begin{subfigure}{0.45\linewidth}
        \includegraphics[width=1.0\linewidth]{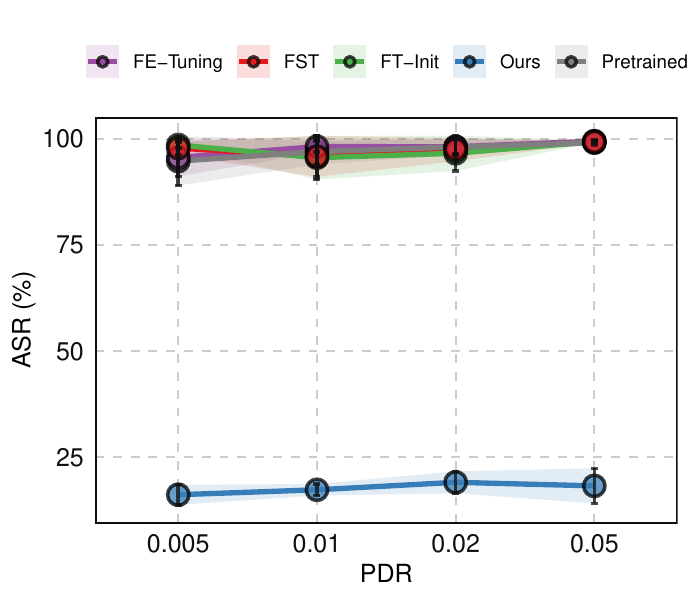}
        \caption{EMBER}
    \end{subfigure}
    \begin{subfigure}{0.45\linewidth}
        \includegraphics[width=1.0\linewidth]{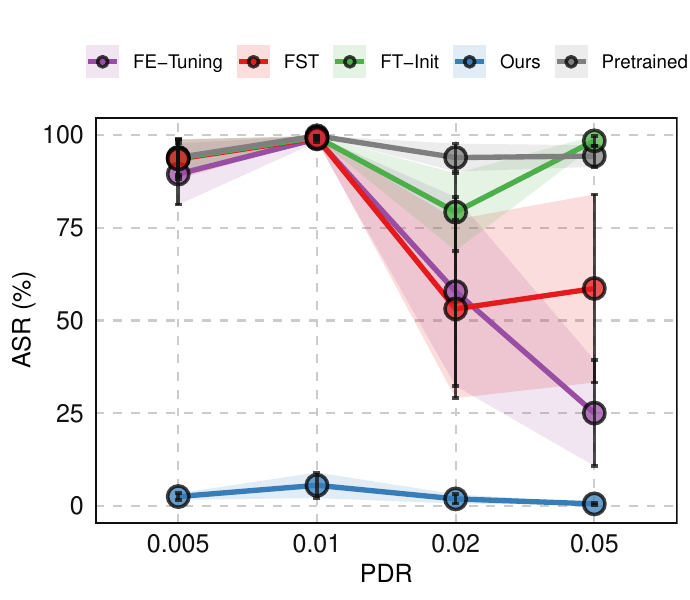}
            \caption{AndroZoo}
    \end{subfigure}
    \caption{\added{Attack Success Rates of Fine-tuning methods under different poisoning rates with fine-tuning size of 10\%.}}
    \label{fig:poison-rate-asr}
    \vspace{-1cm}
\end{figure}
\begin{figure*}[h!]
    \centering
    \begin{subfigure}{0.99\textwidth}
        \includegraphics[width=1.0\textwidth]{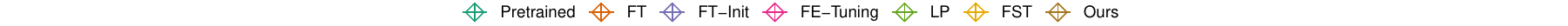}
    \end{subfigure}
    \begin{subfigure}{0.24\textwidth}
        \includegraphics[width=1.0\textwidth]{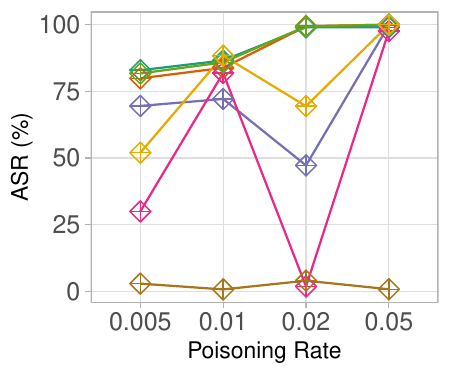}
        \caption{FT size = 1\%}
        \label{fig:ft-size-0.01-jigsaw}
    \end{subfigure}
    \begin{subfigure}{0.24\textwidth}
        \includegraphics[width=1.0\textwidth]{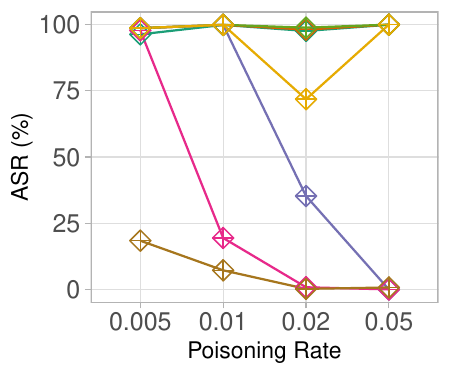}
        \caption{FT size = 2\%}
    \end{subfigure}
    \begin{subfigure}{0.24\textwidth}
        \includegraphics[width=1.0\textwidth]{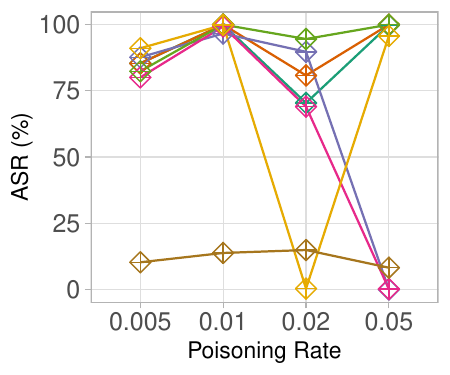}
        \caption{FT size = 5\%}
        \label{fig:ft-size-0.05-jigsaw}
    \end{subfigure}
    \begin{subfigure}{0.24\textwidth}
        \includegraphics[width=1.0\textwidth]{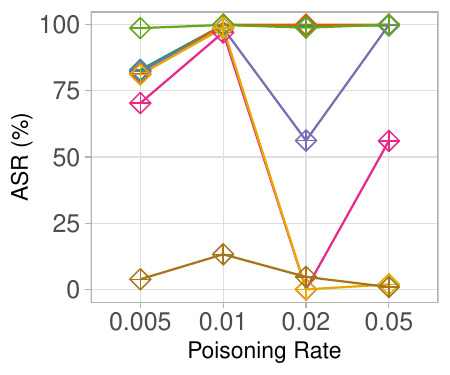}
        \caption{FT size = 10\%}
        \label{fig:ft-size-0.1-jigsaw}
    \end{subfigure}
    \caption{Comparison of different methods under small to large fine-tuning data size \added{with different PDRs}.}
    \label{fig:change-ft-size-jigsaw}
    \vspace{-0.5cm}
\end{figure*}
\newcommand{\reusedfootnote}{\footnotemark[1]}
\subsubsection{\textbf{RQ2:} \replaced{Is PBP effective against backdoor attacks carried out by attackers with varying levels of strength?}{Can \method{} purify backdoor attacks given different attacker power?}} 
\replaced{To answer this question, we evaluate the performance of \method{} and selected baselines across varying poisoning data rates (PDR) — the proportion of training data manipulated by the adversary. PDR reflects both the strength of backdoor attacks and the adversary’s power.}{To answer this question, we evaluate the performance of our \method{} approach with selected baselines for different poisoning data rates (PDR) --- i.e., by varying the proportion of training data controlled by the adversary. 
The poisoning data rate is an important factor in the strength and durability of backdoor attacks as well as the power of the adversary.} 
\added{
In realistic scenarios, attackers often have limited access to training data and aim to succeed with low PDR ($\leq$ 1\%). To assess defense efficacy, we vary the PDR across 
$[0.5\%, 1\%, 2\%, 5\%]$ and perform multiple runs to ensure the stability of each method.}
We plot ASRs of the compared fine-tuning methods in Fig.~\ref{fig:poison-rate-asr}. For better visualization, we plot the performance of the top-4 methods. 
\deleted{The non-family-targeted attack on EMBER is resilient under all fine-tuning baselines except for our \method{} approach.} 
\replaced{All baselines fail to reduce the ASR of the backdoor with the EMBER dataset or achieve unnoticeable reduction across multiple runs. Meanwhile, \method{} can achieve the lowest ASR across all PDRs considered and low deviation.}{The efficiency of our method under different poisoning rates does not deviate substantially.} 
Concerning the family-targeted attack on AndroZoo, FE-Tuning and FST can purify the backdoor when the poisoning rate increases to 2\%--5\%. 
However, these methods cannot purify the backdoor when the poisoning rate is \replaced{low}{small}, i.e., 0.5\% on both \deleted{backdoor} attacks. \added{To this end, \method{} is the only fine-tuning strategy that can mitigate the backdoor effect across various attacker-power settings with the lowest variation, where the PDR increases from 0.5\% to 5\%.}

\subsubsection{\textbf{RQ3:} Can \method{} purify the backdoor stably under different fine-tuning assumptions?}
We evaluate our approach against state-of-the-art fine-tuning methods under different \replaced{ratios of fine-tuning data to training data}{ratios of fine-tuning data}, increasing from 1\% to 10\%, using the family-targeted attack.\footnote{We elide discussion of the non-family-targeted attack since we previously demonstrated existing approaches are not robust against these attacks
}.
Fig.~\ref{fig:change-ft-size-jigsaw} shows that \method{} performs well even when given a small portion of fine-tuning data, \added{1\% -- 983 samples}.
With the family-targeted attack on AndroZoo, other baselines underperform when the poisoning rate is small, and are thus not effective against non-family-targeted backdoor attacks.
\deleted{On the other hand, \method{} shows stability and effectiveness in mitigating the ASR in these cases}.
The FST method can perform better when the fine-tuning data size is greater than 5\%.
Specifically, when the fine-tuning size is 10\% (Fig.~\ref{fig:ft-size-0.1-jigsaw}), FST can mitigate the ASR low to almost zero when the poisoning rate is large, i.e., 2-5\%.
However, given a fine-tuning size small around 1\% (Fig.~\ref{fig:ft-size-0.01-jigsaw}), AST does not reduce ASR, i.e., the ASR still maintains almost 100\% in the worst case.
Other baselines such as FE-Tuning and FT-Init, can mitigate the attack success rates in the case the poisoning rate is small---i.e., in the best case, they achieve 0\% ASR, but their performance is unstable across different settings.
The reason is that these methods depend on the initialization procedure before fine-tuning---i.e., some layers are initialized using Gaussian noise and can be frozen during fine-tuning.
\method{} is the only one that exhibits \added{stability and outperformed effectiveness in mitigating the ASR across these settings.} \deleted{stability under all settings and outperforms other baselines across in most of the settings.}

\begin{figure}[h!]
    \centering
    \begin{subfigure}{0.45\linewidth}
        \includegraphics[width=1.0\linewidth]{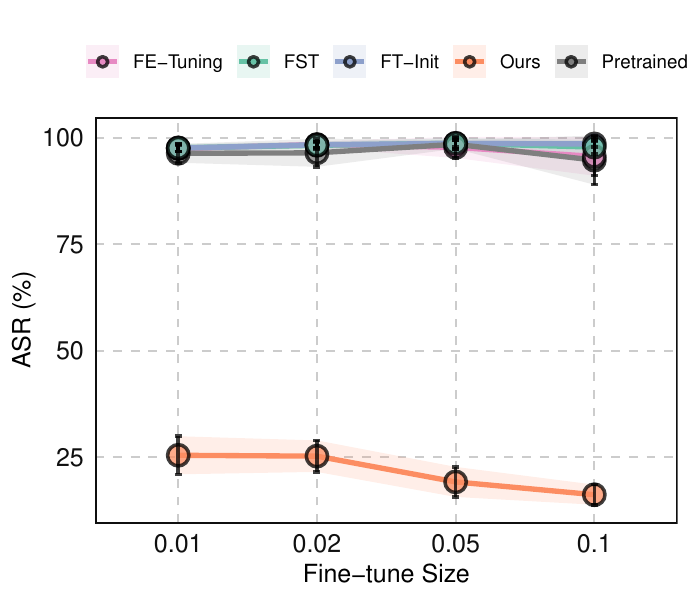}
        \caption{EMBER}
        \label{fig:ember-pdr-0.005}
    \end{subfigure}
    \begin{subfigure}{0.45\linewidth}
        \includegraphics[width=1.0\linewidth]{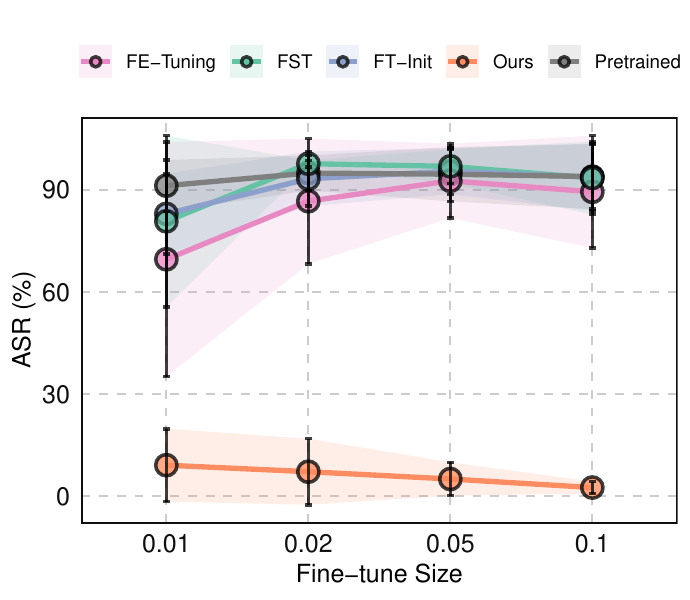}
            \caption{AndroZoo}
            \label{fig:jigsaw-pdr-0.005}
    \end{subfigure}
    \vspace{-0.25cm}
    \caption{\added{Attack Success Rates of Fine-tuning methods under different fine-tuning sizes with PDR of 0.5\%.}}
    \label{fig:jigsaw-ember-pdr-0.005}
    \vspace{-0.5cm}
\end{figure}

\added{Based on the analysis from Yang et al.~\cite{yang2023jigsaw}, the effectiveness of the backdoor inserted can be affected by other families presenting in the training set. After analyzing their feature distributions, there exist common features across different malware families. Such similarity could be caused by many reasons, e.g., code reuse among
different malware authors, shared libraries, or the reuse of specific attack techniques~\cite{calleja2018malsource}. This insight helps explain the variability in fine-tuning performance. As the size and composition of the fine-tuning dataset change, so does the distribution of malware families, introducing shifts in the feature space. When the features in the fine-tuning set resemble those of the original target families, the backdoor model is more likely to generalize well, maintaining high attack success rates. In such cases, benign samples containing features that overlap with the adversary’s crafted triggers can act as subtle reminders of the backdoor pattern, helping the model retain the malicious behavior.
We further rigorously evaluate the performance under multiple runs corresponding to each fine-tuning size to simulate different malware family distributions in the fine-tuning set and present the results in Fig.~\ref{fig:jigsaw-ember-pdr-0.005}. For better visualization, we only plot the performance of the top-4 methods.}
\added{First, Fig.~\ref{fig:ember-pdr-0.005} confirms that the dynamic of the families presenting in the fine-tuning dataset only holds with the family-based attack, i.e., JIGSAW on AndroZoo. Specifically, the variation on multiple runs with EMBER (Fig.~\ref{fig:ember-pdr-0.005}) is much lower than the corresponding number of AndroZoo (Fig.~\ref{fig:jigsaw-pdr-0.005}). }
\added{Our observations indicate that larger fine-tuning datasets with more diverse malware families are more effective in helping \method{} dilute the impact of backdoors. As shown in Fig.~\ref{fig:jigsaw-pdr-0.005}, increasing the fine-tuning size reduces the variation in \method{}'s purification performance, leading to more consistent results. However, with larger fine-tuning sizes, baseline methods that rely on re-initialization such as FE-Tuning struggle to prevent the model from converging back to the original backdoored state, demonstrating their limited efficacy in maintaining robustness.}
\added{The results demonstrate that \method{} is the most effective and stable approach across different fine-tuning sizes, with its overall performance improving as the fine-tuning size increases.}

\begin{figure}[h!]
    \centering
    \includegraphics[width=0.98\linewidth]{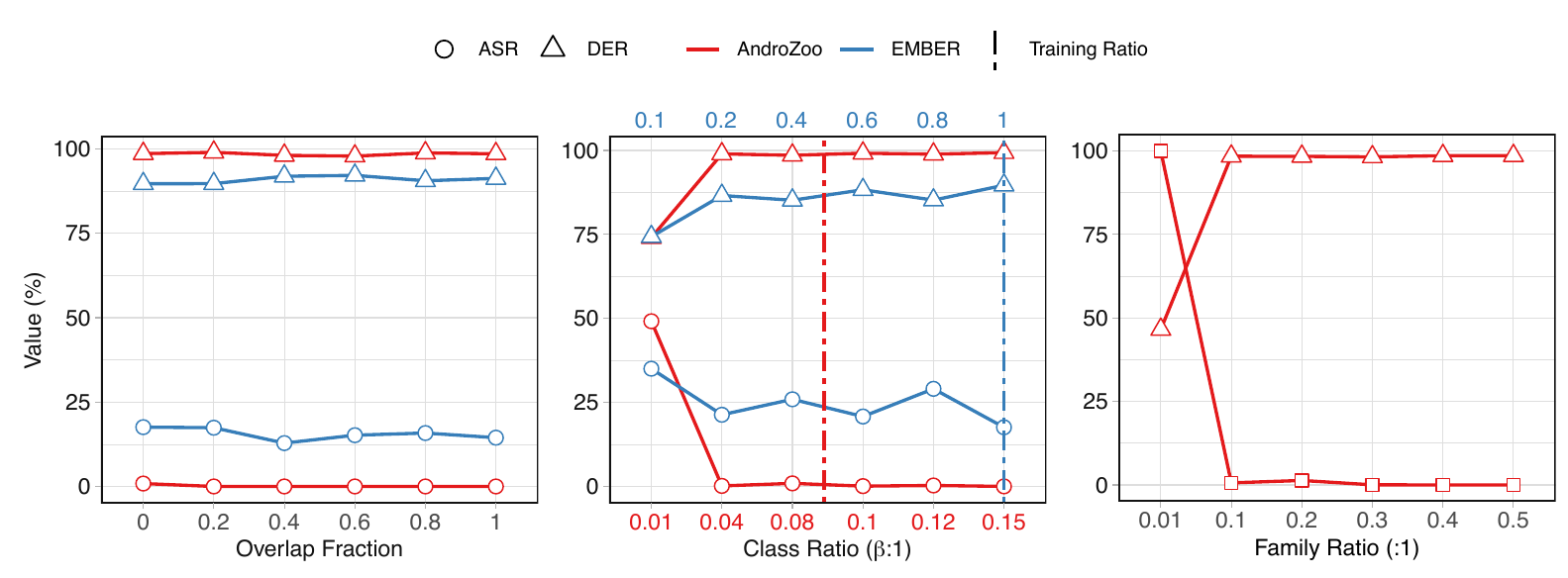}
    \caption{\added{\method{}'s performance under different fine-tuning dataset conditions.}}
    \label{fig:ftdata-study}
    \vspace{-0.8cm}
\end{figure}

\vspace{-1mm}
\noindent\textbf{\added{Discussion on the construction of the fine-tuning dataset.}} 
\added{Since the fine-tuning dataset plays an important role in the performance of the purification methods, we further analyze different factors for constructing a fine-tuning dataset including (i) overlapping fraction with training data, (ii) class ratio and (iii) number of malware families in the fine-tuning dataset. The summarized results are plotted in Fig.~\ref{fig:ftdata-study}.}
\added{We rigorously vary these three factors from extreme to most favorable cases to strengthen the practicability of the defender assumptions. First, \method{} can achieve robust purification efficacy even when the fine-tuning dataset is constructed by reusing a portion of available training data, showing that the fine-tuning data is not necessarily non-overlapping with the training data. Second, \method{} does not require the exact original class ratio for successful purification. Indeed, with AndroZoo, our method can erase the backdoor under an extreme case where the negative per positive class ratio is $0.04:1$. Third, the fine-tuning process does not require the presence of all malware families, where \method{} is effective from the family ratio of $0.1:1$, compared to the original malware families in the training set, i.e., approximately 30 families in our setting. More detailed results and analysis are left in the Appendix.}


\subsubsection{\textbf{RQ4:} How is \method{}'s efficiency and sensitivity to its hyperparameters and model architectures?} 
\begin{figure}
    \centering
    \begin{subfigure}{0.45\linewidth}
        \includegraphics[width=1.0\linewidth]{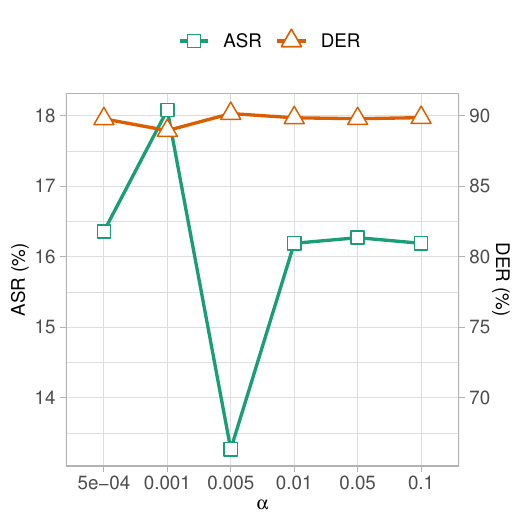}
        \caption{Loss alignment control parameter $\alpha$}
        \label{fig:change-alpha}
    \end{subfigure}
    \begin{subfigure}{0.45\linewidth}
        \includegraphics[width=1.0\linewidth]{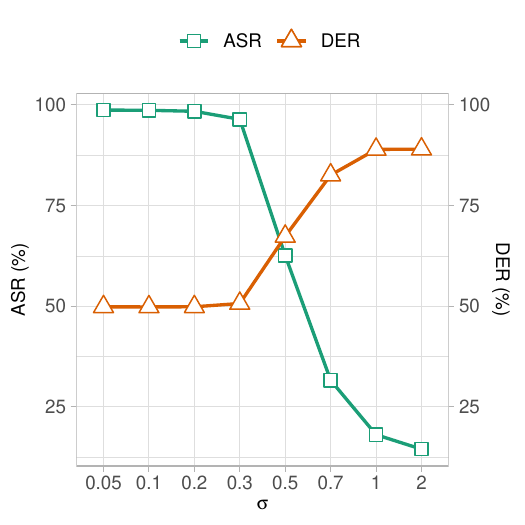}
        \caption{Parameter smoothing control parameter $\sigma$}
        \label{fig:change-sigma}
    \end{subfigure}
    \caption{Performance of model with different settings for hyperparameters.}
    \label{fig:enter-label}
    \vspace{-0.75cm}
\end{figure}

\added{To answer this question, we evaluate the performance of \method{} under varied settings for its hyperparameters, the depth of the model, and network architecture.}
We first evaluate the performance of \method{} under different settings of two important hyperparameters: $\alpha$ (Eqn.~\ref{eqn:align_loss}), which is the factor balancing activation alignment in the first step, and $\sigma$ (Eqn.~\ref{eqn:smoothing}), which controls the magnitude of the isotropic Gaussian noise added to the original model $\theta_0$ in the second phase.
The results are shown in Fig.~\ref{fig:change-alpha} and Fig.~\ref{fig:change-sigma}. 
All the experiments are conducted with the non-family-targeted attack on EMBER with a fine-tuning size of 10\%, and the most stealthy attack with 0.5\% poisoning rate.

In \added{the} first experiment, we select the value for $\alpha$ from $[5e-04, 0.001, 0.005, 0.01, 0.05, 0.1]$, and measure ASR and DER to evaluate the effectiveness of each setting. 
From Fig.~\ref{fig:change-alpha}, the selection of $\alpha = 0.005$ is the most effective since it brings the lowest ASR and the highest DER. 
We observe that higher $\alpha$ causes the model to align with the activation distribution of the backdoored model faster, and we suggest setting this value from the range of $0.001$ to $0.01$ so that the training accuracy increases during the first phase of \method{}. 
In the second experiment, we select the value for $\sigma$ from small to large $[0.05, 0.1, 0.2, 0.3, 0.5, 0.7, 1, 2]$.
In Fig.~\ref{fig:change-sigma}, the higher the value of $\sigma$ is, the lower the ASR a fine-tuned model achieves. 
However, as $\sigma$ increases, the clean accuracy decreases, because the fine-tuned model deviates a larger distance compared to the original model.  
To balance this trade-off, we suggest that this value should be tuned in a range of $0.5$ to $1.0$. 
\replaced{Selecting too small a $\sigma$ reduces backdoor mitigation effectiveness, as the model remains too close to the backdoored state, with minimal updates per round rendering activation shifts ineffective.}{Selecting too small a value for $\sigma$ can lead to a low effect in mitigating the backdoor since this model is already close to the backdoored model, and the update in each round is also small, the activation shift will not be effective in this case.} \\
\input{tabs/different-models}
In addition, we demonstrate the stability of \method{}
 under different model architectures, we vary the size of the MLP models increasingly from 20M to 120M. The results in Table~\ref{tab:diff-models} showcases that \method{} performs effectively with different model settings. Specifically, \method{} can successfully reduce ASR by a gap of 90\% when the model is largest, i.e., 120M. 
\added{In the Appendix, we further investigate performance of \method{} with additional model architectures including CNN~\cite{lecun1998gradient}, ResNet-18~\cite{he2016deep} and VGG~\cite{simonyan2014very}. The rationale that \method{} can work with different network architectures is as follows. First, our method finds the backdoor neuron mask $\mathcal{N}_m$ in a layer-wise manner following Eqn. 4. Thus, a model with more layers would not impact this approach to identifying. Second, our approach addresses not only the backdoored neurons but also the relationships with benign neurons via alternative optimization, which ensures the attack remains localized and does not disrupt the entire network. Therefore, our approach can also handle neural architectures where inter-layer relationships differ.
Third, our intuition is based on the sparsity of gradients of backdoor neurons, which matches other Computer Vision architectures like ResNet, LSTM, GPT-2, and Transformer~\cite{zhang2022neurotoxin,li2023reconstructive,li2024purifying}.}
\added{Our conclusion is that \method{} demonstrates stability across different parameter settings and performs consistently across both simple and complex model architectures.}

\vspace{-2mm}
\subsection{Discussion and Limitations}
\label{sec:discussion}
Through extensive evaluation, we demonstrate the effectiveness and stability of \method{} across various poisoning rates, surpassing existing state-of-the-art strategies. 
\added{In this section, we further discuss the practicability and impact of our approach, insights from different backdoor strategies, and the clean accuracy trade-off.}
\deleted{However, our tuning methods assume that the defender possesses a clean tuning set, which may not always be feasible.} 
\deleted{Additionally, while our approach achieves highest backdoor mitigation effectiveness and performs stably under various settings, it may slightly compromise clean accuracy. 
This necessitates careful attention to protecting learned pretraining features from being compromised during the fine-tuning procedure.}
\deleted{Furthermore, model degradation and fine-tuning data may contain poisoned data, which must be addressed to maintain overall model integrity. The study of how poisoned data can affect backdoor purification methods can be beneficial for the community and should be addressed in future works.}

\noindent\textbf{\added{Practicability of post-defense solution.}}
\added{We want to argue that collecting such a fine-tuning dataset is feasible with multiple solutions. First of all, our method can work with a small dataset size, i.e., 923 samples, and it does not require the presence of all families during the training. To collect those, the defender can rely on open-source repositories and public malware archives such as AndroZoo~\cite{allix2016androzoo}, Malware Bazaar~\cite{malwarebazaar}, VirusShare~\cite{bruzzese2024building}, and MalShare~\cite{malshare}. These repositories allow easy access to malware datasets and eliminate the need to collect raw data independently. In addition, the fine-tuning process only requires one sample for the rare families, and the training data can be reused if it is available. These factors make the collection of fine-tuning datasets more feasible. Recent researches show that synthetic data and augmented data can be generated by mutating existing samples (e.g., adding junk code, or altering headers) to mimic how malware evolves~\cite{smtith2021malware,sharma2024migan,catak2021data}, which can help to seed the dataset. }
\added{This feasibility enhances the impact of post-defense backdoor purification across critical scenarios. Organizations/defenders often acquire pre-trained or public backbone models for malware detection through purchases from third-party vendors or open-source repositories. 
However, vendor-provided models can be compromised or backdoored, potentially leading to catastrophic failures, such as banking systems overlooking malware that siphons customer data and healthcare institutions misclassifying ransomware, exposing patient records to attackers.
The purification is necessary when an institution/defender uses the malware classifier model and analysts notice that certain samples related to identified malware variants are consistently being labeled as safe. 
Instead of retraining, which is resource-intensive, defenders can fine-tune these models using small, curated datasets. This approach ensures reasonable deployment of a trustworthy, customized detection system, neutralizing backdoors while adapting to evolving threats with minimal cost and effort.}
\added{Our method currently assumes the defender can control the trustworthiness of the fine-tuning data and conduct additional investigating such as third-party labeling to ensure the fine-tuning data is clean. The study of how poisoned data can affect backdoor purification methods can be beneficial for the community and should be addressed in future works.}

\noindent\textbf{Family-targeted backdoor attack is more fragile during fine-tuning.}
In the original work of Yang et al.~\cite{yang2023jigsaw}, the Jigsaw Puzzle attack was stealthier than non-family-targeted backdoor attack in bypassing mainstream defense such as MNTD~\cite{xu2021detecting}. 
Specifically, the detection AUCs\footnote{AUC is a measure of the Area Under Curve, which captures how well a classifier can distinguish classes.  0.5 implies the classifier is no better than random guessing.  Scores near 1.0 indicate more perfect discriminatory power.} %
are below 0.557 (slightly better than random guessing) given Jigsaw Puzzle attacks, while the corresponding number for the Explanation-guided attack is up to 0.919. 
However, during the fine-tuning process, the Jigsaw Puzzle is more fragile compared to the Explanation-guided, as fine-tuning defenses such as FE-Tuning, FT-Init, and FST can mitigate the ASR near zero in many scenarios. 
Our \method{} approach can erase the backdoor effect in all considered scenarios with smaller compensation for clean accuracy. 
However, the lowest ASR that \method{} can reach with the Explanation-guided attack is around 15\%, and all considered baselines fail in mitigating this attack during the fine-tuning process even when the fine-tuning size is large. 
\input{tabs/continue_train}

\noindent\textbf{\replaced{Addressing accuracy degradation during fine-tuning. }{Addressing accuracy degradation during training. }}
\replaced{While our approach achieves state-of-the-art backdoor mitigation and remains stable across diverse settings, it may slightly reduce clean accuracy—--a trade-off common in backdoor purification methods. }{A key threat to our approach is the trade-off between its effectiveness in mitigating backdoor attacks versus the degradation in clean accuracy.}
\added{The reduction in clean accuracy results from the purified model misclassifying some benign samples as malware, leading to a higher false positive rate. As shown in Fig.~\ref{fig:pbp-final-act}, the decision boundary of the fine-tuned model expands beyond that of the clean model. This occurs because PBP assigns smaller activation values to backdoored malware samples, attempting to separate them more distinctly from benign ones.
This behavior contrasts with the backdoor insertion process, which seeks to bring backdoored samples closer to those of the targeted class, disguising the malware effectively. In reversing this effect, \method{} creates a more distinct boundary between benign and malicious samples. However, this expanded boundary can cause overfitting on the fine-tuning dataset, introducing degradation in overall accuracy.
Despite the reduction in clean accuracy, the primary objective of backdoor mitigation is to ensure robust security. This involves making a trade-off: sacrificing a degree of performance to achieve higher security, which aligns with common practices across general ML applications. As shown by recent research~\cite{yi2024badacts,min2024towards}, this security-performance trade-off remains a fundamental and active area of research.}
To better address this issue, we conduct an additional experiment to continue training the fine-tuned model $\theta_T$ produced by our post-training backdoor purification in Algo.~\ref{algo:main}. %
Specifically, we continue to train this model using only the cross entropy loss $\mathcal{L}_{CE}$ on the fine-tuning data $\mathcal{D}_{ft}$ for 20 more epochs. 
Table~\ref{tab:continue-train} demonstrates the 
C-Acc improvement was achieved by continuing the training process. In the optimal scenario, the continued training process can reduce the trade-off in clean accuracy by up to 1.42\%, 
when the poisoning rate is small (i.e., 0.5\%). 
Empirically, the accuracy improvement does not vary much beyond 20 epochs. 
Thus, the trade-off created by our method can be addressed by continued training, but cannot be completely addressed due to the stealthiness of the inserted backdoor attacks, leaving an open research direction for the future. 
\added{Incorporating reverse engineering techniques can be a potential solution to enhance purification efficiency by accurately identifying backdoored neurons. This targeted approach ensures the fidelity of pre-trained features during fine-tuning, preserving essential representations while effectively neutralizing potential threats. }
Overall, \method{} is the first post-training method that can mitigate both family-targeted and non-family-targeted backdoor attacks in malware classifiers. 

%% file: tabs/ndss_r2/merge_ft_result.tex
\begin{table*}[h!]
\centering
\caption{\added{Performance of Fine-tuning Methods under Explain-Guided Backdoor Attacks on EMBER and JIGSAW Backdoor Attacks on AndroZoo. The best numbers are highlighted in \best{bold-underline}, the second-best numbers are in \second{underline}.}}
\label{tab:ember-apg-0.1}

\resizebox{1.0\textwidth}{!}{
\begin{tabular}{@{}l|l|cc|cc|cc|cc|cc|cc|cc@{}}
\toprule
\midrule
\multirow{2}{*}{Dataset}  & \multirow{2}{*}{\begin{tabular}[c]{@{}l@{}}Poisoning\\ Rate\end{tabular}} & \multicolumn{2}{c}{Pre-trained} & \multicolumn{2}{c}{FT} & \multicolumn{2}{c}{FT-init} & \multicolumn{2}{c}{FE-tuning} & \multicolumn{2}{c}{LP} & \multicolumn{2}{c}{FST} & \multicolumn{2}{c}{Ours} \\ \cmidrule(lr){3-4} \cmidrule(lr){5-6} \cmidrule(lr){7-8} \cmidrule(lr){9-10} \cmidrule(lr){11-12} \cmidrule(lr){13-14} \cmidrule(lr){15-16}
                          &                                                                           & C-Acc          & ASR            & C-Acc      & ASR       & C-Acc        & ASR          & C-Acc         & ASR           & C-Acc      & ASR       & C-Acc      & ASR        & C-Acc       & ASR        \\ \midrule
\multirow{4}{*}{EMBER}    & 0.005                                       & 99.01          & 99.23         
& 99.10      & \second{99.50}     
& 99.07        & 99.27        
& 99.11         & 99.50         
& 99.11      & 99.52     
& 99.07      & 99.61      
& 96.57       & \best{17.83}      \\
  & 0.01                                                                & 98.94          & 98.79          
  & 99.06      & 99.54     
  & 99.04        & 99.41        
  & 99.03         & \second{99.16}         
  & 99.08      & 99.39     
  & 99.04      & 99.59      
  & 96.52       & \best{15.44}      \\
& 0.02                                                                  

& 98.98      & 99.43          
& 99.08      & 99.69     
& 99.01      & \second{99.52}        
& 99.06      & 99.63         
& 99.10      & 99.61     
& 99.04      & 99.66      
& 96.57       & \best{17.83}      \\
& 0.05                                                                  & 98.99          & 99.43          
& 99.08      & 99.87     
& 99.06        & 99.91        
& 99.07         & 99.82         
& 99.03      & 99.83     
& 99.90      & \second{99.76}      
& 96.41       & \best{17.58}      \\ \cmidrule(lr){1-16} 
\multirow{4}{*}{AndroZoo} & 0.005                                                                     
& 98.53                     
& 82.91                   
& 98.63                     
& 81.53                   
& 98.62                     
& 82.36                   
& 98.55                     
& \second{70.38}                   
& 98.57                     
& 98.69                   
& 98.66                     
& 81.12                   
& 96.76                     
& \best{3.83}                    \\
  & 0.01                                                                      & 98.56                     
& 99.90                   
& 98.67                     
& 100.0                   
& 98.67                     
& 98.62                   
& 98.60                     
& \second{97.07}                   
& 98.58                     
& 99.90                   
& 98.68                     
& 98.76                   
& 96.88                     
& \best{13.26}                   \\
                          & 0.02                                                                      & 98.58                     
& 99.45                   
& 98.45                     
& 100                     
& 98.53                     
& 56.23                   
& 98.55                     
& \second{0.03}                    
& 98.57                     
& 98.86                   
& 98.55                     
& \best{0.01}                       
& 96.64                     
& 4.73                    \\
                          & 0.05                                                                      & 98.59                    
& 99.72                   
& 98.58                     
& 100.0                   
& 98.62                     
& 99.90                   
& 98.57                     
& 56.09                   
& 98.53                     
& 100.0                   
& 98.63                    
& \second{1.90}                    
& 96.86                     
& \best{0.89} \\        \midrule\bottomrule
\end{tabular}}
\vspace{-2mm}
\end{table*}

%% file: tabs/different-models.tex
\begin{table}[t!]
\centering
\caption{Performance of \method{} under different model settings.}
\label{tab:diff-models}
\scriptsize
\resizebox{0.8\linewidth}{!}{
\begin{tabular}{@{}llcccc@{}}
\toprule
\multirow{2}{*}{Model Architecture} & \multirow{2}{*}{\#Params} & \multicolumn{2}{c}{Clean} & \multicolumn{2}{l}{Fine-tuned} \\ \cmidrule(lr){3-4}  \cmidrule(lr){5-6}
                                    &                           & C-Acc          & ASR            & C-Acc             & ASR              \\ \cmidrule(lr){1-6}
2000/1000/500                       & 22M                       & 98.63          & 98.72          & 95.91             & 14.64            \\
4000/2000/1000                      & 50M                       & 98.53          & 82.91          & 96.76             & 3.83             \\
4000/4000/2000                      & 64M                       & 98.57          & 96.10          & 97.71             & 8.87             \\
8000/4000/2000                      & 120M                      & 98.60          & 95.34          & 97.50             & 5.70             \\ \bottomrule
\end{tabular}
}
\vspace{-0.5cm}
\end{table}

%% file: tabs/continue_train.tex
\begin{table}[t!]
\vspace{-0.5cm}
\centering
\caption{\added{Results of continuing training to improve C-Acc on AndroZoo dataset with varied PDR.}}
\label{tab:continue-train}
\scriptsize
\resizebox{0.98\linewidth}{!}{
\begin{tabular}{@{}lcccccccc@{}}
\toprule
\multirow{2}{*}{Models} & \multicolumn{2}{c}{0.005} & \multicolumn{2}{c}{0.01} & \multicolumn{2}{c}{0.02} & \multicolumn{2}{c}{0.05} \\ \cmidrule(lr){2-3} \cmidrule(lr){4-5} \cmidrule(lr){6-7} \cmidrule(lr){8-9}
                        & C-Acc       & ASR         & C-Acc       & ASR        & C-Acc       & ASR        & C-Acc       & ASR        \\ \midrule
Clean                   & 98.52       & 0.01        & 98.52       & 0.01       & 98.52       & 0.01       & 98.52       & 0.01       \\
Backdoored              & 98.53       & 82.91       & 98.48       & 99.90      & 98.58       & 99.45      & 98.59       & 99.72      \\
Fine-tuned              & 96.76       & 3.83        & 96.88       & 13.26      & 96.64       & 4.73       & 96.86       & 0.89       \\
Continued               & 98.18       & 5.73        & 97.65       & 8.25       & 97.47       & 0.17       & 97.55       & 0.03       \\ \bottomrule
\end{tabular}
}
\vspace{-0.5cm}
\end{table}

%% file: secs_R2/related.tex
\section{Related Works}
In this section, we discuss the current body of work on backdoor attacks and defenses for malware classification, specifically backdoor purification during fine-tuning.

\smbf{Backdoor attacks for malware classification.}
Even though backdoor attacks have been extensively studied in the image domain, most of them cannot be applied to malware classifiers due to two main reasons: 
(1) incompatible domain (e.g., style transfer does not apply even in the case where the binary is represented as an image), and 
(2) realizing the trigger embedding from the feature space to the problem space is very difficult, as feature extraction is not bijective~\cite{Lucas2019MalwareMB,Ebrahimi2020BinaryBE}. 
Overcoming these concerns, \citet{Li20223357} devised a feasible backdoor attack by appropriately using evolutionary algorithms to generate realizable triggers.
However, the full-access assumption in their work is usually too strong for the case of malware classification, as multiple trusted third-party AV vendors usually perform labeling. 
\citet{severi2021explanation} relaxed this requirement by introducing a clean-label backdoor attack.  
Similarly, \citet{yang2023jigsaw} proposed a selective backdoor that improves stealthiness and effectiveness, following the intuition that a malware author would prioritize protecting their own malware family instead of all malware in general.

\smbf{Backdoor countermeasures for malware classifications.}
Similar to the image space setting, the most popular defense mechanism against backdoor attacks for malware classification is adversarial training~\cite{wang2020mdea}, in which the model is trained and/or fine-tuned to correctly predict on adversarially crafted samples. 
However, generating new adversarial examples for the model at every epoch is very computationally intensive, and can take much longer than traditional training \cite{tran2022mpa}.
Another approach leverages various heuristics to remove adversarial samples from the training dataset. 
This way, any manipulations introduced by adversaries can be undone before the samples are sent to the malware detector. 
However, these empirical defenses usually only work for very few adversarial attack methods, and are thus attack-specific~\cite{ling2023adversarial}.
For the situation where the end user only has limited additional labeled clean data and/or the required resources for retraining becomes infeasible, previous works have adapted image-space backdoor detection mechanisms~\cite{gao2019strip,xu2021detecting} to malware~\cite{d2023lookin}. 
However, the ultimate results were not compelling, sometimes performing only as well as random guessing.

\smbf{Backdoor purification during fine-tuning.}
Fine-tuning has been proven to work well as a post-training defense mechanism against backdoor attacks~\cite{liu2018fine,wu2022backdoorbench}, are model-agnostic, and can be combined with existing training methods and/or orthogonal approaches to robustness~\cite{zhu2023enhancing}. 
While fine-tuning-based methods are most popular with large pre-trained models~\cite{wei2021finetuned,radford2021learning,zhang2022fine} and will perform better than test-time defenses~\cite{sha2022finetuningneedmitigatebackdoor}, they still require a considerable amount of finetuning data to effectively remove the embedded backdoor. 
\replaced{Recent methods, such as Teacher-Student (T-S) purification, have shown promise by leveraging neuron pruning or similar techniques to mitigate backdoors during model refinement\cite{li2023reconstructive}. However, these methods face limitations in models without residual-block components or in non-image tasks\cite{li2021neural}. To the best of our knowledge, no prior work has addressed the purification of backdoors through fine-tuning in the context of malware classification. This gap highlights the potential of using fine-tuning and purification techniques—such as PBP (Post-Backdoor Purification)—in malware models, which operate under constraints different from traditional image-based tasks.}
{To the best of our knowledge, there has been no prior work on purifying backdoors during fine-tuning for malware classification.}

%% file: secs_R2/artifact_appendix.tex
\appendices

\section{Artifact Appendix}
\label{sec:require}
The artifact appendix is meant to be a self-contained document presenting a roadmap for setting up and evaluating your artifact~\footnote{AEC evaluated a
prior version of the artifact and the modifications do not affect the claims and experiments in our submitted artifacts.}. It should provide the following elements:
\begin{enumerate}
\item a list of the hardware, software, and configuration \textbf{requirements} for running the artifact;
\item a clear description of how, and in what respects, the artifact \textbf{supports} the research presented in the paper;
\item a guide for how others can \textbf{execute and validate} the artifact for its functional and usability aspects;
\item the \textbf{major claims} made by your paper and a clear description of how to obtain data for each claim through your supplied artifact.
\end{enumerate}

\subsection{Description \& Requirements}
This section lists all the information necessary to recreate the experimental setup we used to run our artifact.

\subsubsection{How to access} 
We make our code and dataset publicly available at 
\url{https://github.com/judydnguyen/pbp-backdoor-purification-official}. The artifact materials for this paper are permanently available at DOI: \url{https://doi.org/10.5281/zenodo.14253945}. 

\subsubsection{Hardware dependencies} an NVIDIA A6000 GPU is encouraged but not required. Our artifacts can be run on a commodity desktop machine with an x86-64 CPU with storage of at least 22GB for data only. To ensure that all artifacts run correctly, it is recommended to use a machine with at least 16 cores and 48 GB of RAM. In the original experiments, we set the number of workers to 54, but for a computer with a smaller number of cores, the number of workers could be reduced, so we set the default as 16. Our code can be run using a CPU if a GPU is not available. However, the using of CPU may not completely ensure to achieve numbers we reported.

\subsubsection{Software dependencies} 
the required packages are listed on \texttt{environment.yml}. We encourage installing and managing these packages using \texttt{conda}. Our artifacts have been tested on Ubuntu 22.04.

\subsubsection{Benchmarks} 
\label{subsec:data}
In our experiments, we used two datasets.
1. Ember-v1 dataset, accessed at \url{https://github.com/elastic/ember}.
2. AndroZoo dataset, accessed at \url{https://androzoo.uni.lu/}.
We included the implementation for all the baselines used in our works in our code repository. We provide a cloud link to share both processed datasets \href{https://doi.org/10.5281/zenodo.14253945}{here}. After downloading and decompressing file \texttt{NDSS603-data-ckpt.zip}, the structure of this folder is as follows:
\begin{verbatim}
NDSS603-data-ckpt/
    └── apg/
    ├── ember/
    └── ckpts/
\end{verbatim}
Here, dataset folders \texttt{apg} and \texttt{ember} are ready to use and need to be moved to \texttt{datasets/} folder. The clean checkpoint version of the EMBER dataset is saved at \texttt{ckpts} and needs to be moved to \texttt{ models/ember/torch/embernn.}

\subsection{Artifact Installation \& Configuration}
We first require that our repository at \url{https://github.com/judydnguyen/pbp-backdoor-purification-official} is downloaded local folder, e.g., via \texttt{git clone} or \texttt{.zip} download. Then, we require downloading datasets and a checkpoint mentioned in Section~\ref{subsec:data}.

We conduct all the experiments using PyTorch version 2.1.0 and run experiments on a computer with an Intel Xeon Gold 6330N CPU and an NVIDIA A6000 GPU. In our original environment, we use \texttt{Anaconda} at ~\url{https://anaconda.org/} to manage the environment dependencies efficiently, ensuring the reproducibility of our experiments. Specifically, we create a virtual environment with \texttt{conda} version \texttt{23.7.3} and install all necessary packages, including PyTorch, NumPy, and Scikit-learn, as detailed in the provided \texttt{environment.yml} file. 
The new environment can be created using:
\begin{verbatim}
conda config --set channel_priority flexible &&
conda env create --file=environment.yml &&
conda activate pbp-code
\end{verbatim}
\subsection{Experiment Workflow}
\textbf{We already set up the environments and data required for our experiments in the Amazon Cloud instance. Therefore, the previous steps can be skipped.}
Our artifacts contain two independent experiments. The first experiment consists of (i) training and (ii) purifying a backdoor model with the EMBER dataset.
The second experiment consists of (i) training and (ii) purifying a backdoor model with the AndroZoo dataset.
The proposed workflow runs the two experiments sequentially. Our repository contains scripts and a bash file that can be used to automate all experiments.

\subsection{Major Claims}

\begin{itemize}
    \item (C1): \method{} achieves better backdoor purification performance compared to baselines in terms of the lowest Backdoor Accuracy (BA) it can produce on the EMBER dataset. This is proven by the experiment (E1) whose results are illustrated/reported in [TABLE II: Performance of Fine-tuning Methods under Explain-Guided Backdoor Attacks on EMBER].
    \item (C2): \method{} achieves better backdoor purification performance compared to baselines in terms of the lowest Backdoor Accuracy (BA) it can produce on the AndroZoo dataset. This is proven by the experiment (E2) whose results are illustrated/reported in [TABLE II: Performance of Fine-tuning Methods under Jigsaw Puzzle Backdoor Attacks on AndroZoo].
\end{itemize}

\subsection{Evaluation}
This section includes all the operational steps and experiments that must be performed to evaluate our artifacts and validate our results. In total, all experiments require between 1 and 5 human minutes and around 1 compute hour. We assume that the machine is configured correctly with the required dependencies, as described in Section~\ref{sec:require}. The same instructions, along with additional details, are provided in the top-level \texttt{README.md} file of our repository.

\subsubsection{Experiment (E1) - Claim (C1)}
[Backdoor Purification With EMBER-v1] [5 human-minutes + 20 compute-minutes]: evaluating the efficacy of different purification methods. \textbf{Noted} that without using of GPU, this experiment may take up to 1 hour for a CPU with 64 cores.

\textit{[Preparation]} In this step, the Ember-v1 data should be ensured to be downloaded and extracted correctly. 
The feature extraction steps are detailed in the original publication, which can be accessed via the following \href{https://github.com/elastic/ember}{GitHub link}. The data should be structured as follows:

\begin{verbatim}
datasets/
└── ember/
    └── np
        ├── watermarked_X_test_32_feats.npy
        └── wm_config_32_feats.npy
    ├── y_train.dat
    ├── X_train.dat
    ├── y_train.dat
    ├── X_test.dat
    └── y_test.dat

\end{verbatim}

\textit{[Execution]}
First, we need to train the backdoored model and then apply purification for this model. In the implementation of Explanation-guided backdoor attacks with EMBER dataset, a clean model is required and can be downloaded at \href{https://cumberland.isis.vanderbilt.edu/NDSS-F24-603.zip}{this link}.
After downloading, put this model in the following folder: \texttt{models/ember/torch/embernn}. To train a backdoored model, we use the following command:
\begin{verbatim}
./train_backdoor_ember.sh
\end{verbatim}
After successfully training a backdoor model, we evaluate different fine-tuning methods by the following command:
\begin{verbatim}
./experiment1_finetune_backdoor_ember.sh
\end{verbatim}

\textit{[Results]}
Upon completion, the results are printed to standard output a summary of the performance of \method{} and other baselines in terms of Backdoor Accuracy (BA) and Clean Accuracy (C-Acc). In our case, the experiment can be considered successful if the BA of our method outperforms considered baselines across different scenarios, i.e., the following line is output.
\texttt{Verified outperforms of PBP: [True, True, True, True, True]}

\subsubsection{Experiment (E2) - Claim (C2)}
[Backdoor Purification With AndroZoo] [5 human-minutes + 1 compute-hour]: provide a short explanation of the experiment and expected results.

\textit{[Preparation]} In this step, the data should be ensured to be downloaded and processed correctly. The data should be structured as follows:

\begin{verbatim}
apg/
    ├── X_train_extracted.dat
    ├── X_test_extracted.dat
    ├── apg-X.dat
    ├── apg-y.dat
    ├── apg-meta.dat
    └── apg_sha_family.csv
\end{verbatim}

\textit{[Execution]}
First, we need to train the backdoored model and then apply purification for this model.
The commands for these are listed in \texttt{experiment1\_finetune\_backdoor\_jigsaw.sh}.

\textit{[Results]}
Upon completion, the results are printed to standard output a summary of the performance of \method{} and other baselines in terms of Backdoor Accuracy (BA) and Clean Accuracy (C-Acc). In our case, the experiment can be considered successful if the BA of our method outperforms considered baselines across different scenarios, i.e., the following line is output. The BA should fall in the range [0-10\%] for the AndroZoo dataset.
\texttt{Verified outperforms of PBP: [True, True, True, True, True]}

\subsection{Notes}
To facilitate the evaluation, we have set up the environment and prepared data beforehand at our provided cloud server. By using this setup, we can skip all the preparation steps and start directly with the actual reproduction. For example,
\begin{verbatim}
    ssh ubuntu@$IP
    cd pbp-code/PBP-BackdoorPurification
    conda activate pbp-code
    ./train_backdoor_ember.sh
    ./experiment1_finetune_backdoor_ember.sh
\end{verbatim}

%% file: secs_R2/appendix.tex
\section{Technical Appendix}
This document serves as an extended exploration of our research, providing an overview of our methods and results. Appendix~\ref{appendix:training_detail} provides a comprehensive discussion of the training process, including datasets, model structures, and configurations used to reproduce the reported results. Next, we provide the proof for our theorems in Appendix~\ref{appendix:theorem}. Additionally, we present supplementary results not included in the main paper in Appendix~\ref{appendix:additional-result}. 
\added{Following that, we conduct an extensive analysis of the effect of the fine-tuning dataset on the \method{}'s performance and the rationale of intuition from our method in Appendix~\ref{appendix:analysis}. We demonstrate the potential efficacy of our method in the Computer Vision (CV) domain with multiple backdoor attacks and model architectures in Appendix~\ref{appendix:cv-domain}.}
\vspace{-2mm}
\subsection{Training configurations}
\label{appendix:training_detail}
\vspace{-2mm}
\input{tabs/training_params}
Table ~\ref{tab:main-params} presents the training and fine-tuning configurations. With both datasets, we use MLP binary classifier models with three hidden layers as \replaced{following}{followed} previous works on malware classifiers~\cite{xu2018deeprefiner,yang2023jigsaw,li2021robust}. We conduct all the experiments using PyTorch version 2.1.0~\cite{pytorch} and run experiments on a computer with an Intel Xeon Gold 6330N CPU and an NVIDIA A6000 GPU.

\subsection{Proof of convergence}
\label{appendix:theorem}

\begin{repeatthm}{thm:convergence}
Let $w_0$ be the initial pretrained weights (i.e. line 13 in \autoref{algo:main}). If the fine-tuning learning rate satisfies:
$$
\eta<\left\Vert\frac{\partial^2\mathcal{L}(w,x)}{\partial w^2}\Big|_{w_0}\right\Vert_2^{-1},
$$
\noindent\autoref{algo:main} will converge.
\end{repeatthm}
\begin{proof}
Starting with some weight set $w_0$, we finetune our backdoored model with learning rate $\eta$. Denoting $x$ as our finetuning dataset, our method can be formulated as follows:

\begin{equation}
\label{eq:taylor2}
w_{i+1} - w_{i} = (-1)^{i} \eta\frac{\partial\mathcal{L}(w, x)}{\partial w}\Big|_{w_{i}}.
\end{equation}

We apply first-order Taylor approximation around $w=w_0$ to get the following:

\begin{alignat*}{3}
    \frac{\partial\mathcal{L}(w,x)}{\partial w}\Big|_{w_{i}}
&=  \frac{\partial\mathcal{L}(w,x)}{\partial w}\Big|_{w_0 + (w_i-w_0)} \\
&\approx    \frac{\partial\mathcal{L}(w,x)}{\partial w}\Big|_{w_0} + (w_i-w_0)\frac{\partial^2\mathcal{L}(w,x)}{\partial w^2}\Big|_{w_0}.
\end{alignat*}

Replacing the derivative with its approximation in Eq. \ref{eq:taylor2}:
\begin{alignat*}{3}
w_{i+1} - w_{i} &= (-1)^{i} \eta\frac{\partial\mathcal{L}(w, x)}{\partial w}\Big|_{w_{i}}\\
&= (-1)^{i} \eta\Bigg[\frac{\partial\mathcal{L}(w,x)}{\partial w}\Big|_{w_0} \\
&\qquad+ (w_{i}-w_0)\frac{\partial^2\mathcal{L}(w,x)}{\partial w^2}\Big|_{w_0}\Bigg] \\
&= (-1)^{i} \eta[\bar{w} + (w_{i}-w_0)\bar{W}]
\end{alignat*}
\\
where
$\bar{w}=\frac{\partial\mathcal{L}(w,x)}{\partial w}\Big|_{w_0}$ and
$\bar{W}=\frac{\partial^2\mathcal{L}(w,x)}{\partial w^2}\Big|_{w_0}$.
Given our alternating optimization scheme flip the gradient sign at every iteration, we can expand the $(-1)^i$ coefficients so that constant terms cancel themselves:
\begin{alignat*}{3}
&w_{2k+2} - w_{2k+1} +  w_{2k+1} - w_{2k} \\
=\;& \sum_{i=2k}^{2k+1}(-1)^{i} \eta[\bar{w} + (w_{i}-w_0)\bar{W}] \\
=\;& -\eta(w_{2k+1} - w_{2k})\bar{W}
\end{alignat*}
which gives
\begin{equation}
\label{eq:even_diff}
w_{2k+2} - w_{2k+1} = (w_{2k+1} - w_{2k})(-I-\eta\bar{W}).
\end{equation}

Similarly, the odd terms can be calculated:
\begin{alignat*}{3}
&w_{2k+3} - w_{2k+2} +  w_{2k+2} - w_{2k+1} \\
=\;& \sum_{i=2k+1}^{2k+2}(-1)^{i} \eta[\bar{w} + (w_{i}-w_0)\bar{W}] \\
=\;& \eta(w_{2k+2} - w_{2k+1})\bar{W},
\end{alignat*}
which leads to
\begin{equation}
\label{eq:odd_diff}
w_{2k+3} - w_{2k+2} = (w_{2k+2} - w_{2k+1})(-I+\eta\bar{W}).
\end{equation}

Substituting \autoref{eq:even_diff} with \autoref{eq:odd_diff} repeatedly, we have this recurrence relation:
\begin{alignat}{3}
\label{eq:even_root}
w_{2k+3} - w_{2k+2}
&= (w_{2k+1} - w_{2k})(-I-\eta\bar{W})(-I+\eta\bar{W}) \nonumber \\
&= (w_{2k+1} - w_{2k})(I-\eta^2\bar{W}^2) \nonumber \\
&= (w_1 - w_0)(I-\eta^2\bar{W}^2)^{k+1} \nonumber \\
&= \eta\bar{w}(I-\eta^2\bar{W}^2)^{k+1}.
\end{alignat}

The matrix power is legal because $\bar{W}$ is a Hessian matrix, which makes it symmetric, and as a result $(I-\eta^2\bar{W}^2)$ is also symmetric. The even-odd index pair difference can also be derived in the same vein:
\begin{alignat}{3}
\label{eq:odd_root}
w_{2k+2} - w_{2k+1}
&= (w_{2k+1} - w_{2k})(-I-\eta\bar{W}) \nonumber\\
&= (w_{1} - w_{0})(-I-\eta\bar{W})(I-\eta^2\bar{W}^2)^k \nonumber\\
&= \eta\bar{w}(-I-\eta\bar{W})(I-\eta^2\bar{W}^2)^{k}.
\end{alignat}

It is straightforward to see that if a symmetric matrix's largest eigenvalue is smaller than 1, its exponential will converge to the zero matrix. We also have the property of a matrix's $2$-norm from its operator norm:

$$
\Vert Ax\Vert_2 \le \Vert A\Vert_2 \Vert x\Vert_2.
$$

Therefore for \autoref{eq:even_root}, if we set $\eta<\Vert\bar{W}\Vert_2^{-1}$, we will get the limit:

$$
\lim_{k\rightarrow\infty}\Vert w_{2k+3}-w_{2k+2}\Vert_2 = 0.
$$

For \autoref{eq:odd_root}, it gets slightly more complicated with an additional factor. However, when we realize that:

$$
\Vert -I-\eta\bar{W}\Vert_2 = \Vert I+\eta\bar{W}\Vert_2\le 2,
$$

\noindent the exponential convergence proof then follows identically. Combining these two limits, we can conclude that $\{w_{i+1}-w_i\}$ as a sequence will converge to 0.
\end{proof}

\subsection{Additional Results}
\label{appendix:additional-result}
We provide the detailed results for experimental results in this section. Table~\ref{tab:ember-apg-0.1} in the main manuscript presents the result for the fine-tuning size of 0.1 for the AndroZoo dataset. 
\input{tabs/model_acc}
\added{The performance of trained models for both tasks is shown in Table~\ref{tab:model-acc}}
Regarding results with different fine-tuning sizes results in Fig.~\ref{fig:change-ft-size-jigsaw}.
We give the detailed results in Tables
\ref{tab:jigsaw-full-0.01}~\ref{tab:jigsaw-full-0.02} and ~\ref{tab:jigsaw-full-0.05}. The results demonstrate that \method{} outperforms selected baselines consistently on different settings, i.e., achieving the lowest attack success rates and balancing between main accuracy and attack success rate. However, we can observe the trade-off between accuracy degradation when the fine-tuning data is limited.  
Specifically, it drops from 96\% to 94\% when the fine-tuning size shrinks. The potential solution for addressing this trade-off is previously discussed in Section~\ref{sec:discussion}.

\input{tabs/ft_jigsaw_0.01}
\input{tabs/ft_jigsaw_0.02}
\input{tabs/ft_jigsaw_0.05}

\begin{figure}[h]
    \centering
    \begin{subfigure}
    {0.30\linewidth}
    \includegraphics[width=1.0\textwidth]{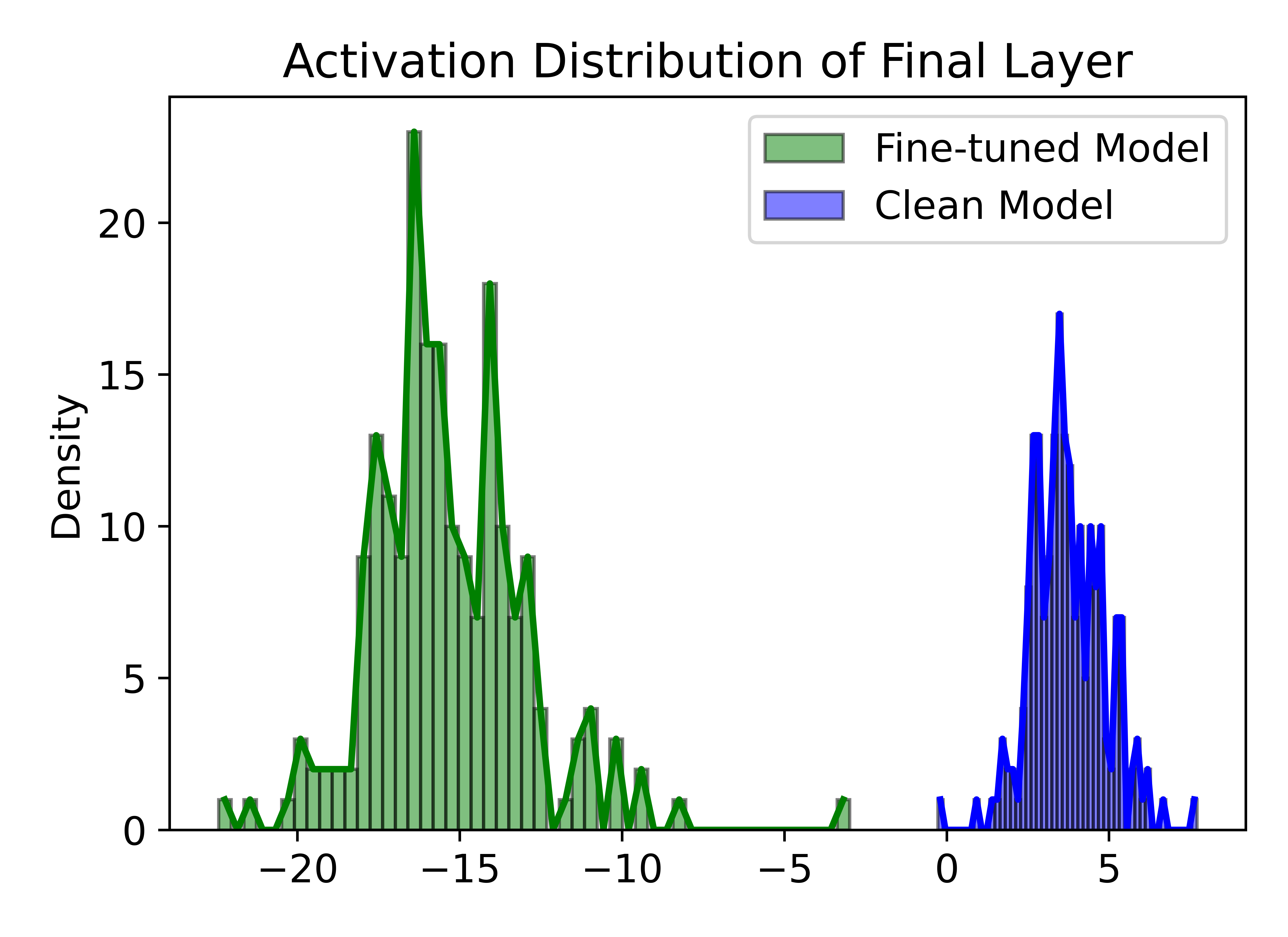}
    \caption{FT}
    \end{subfigure}
    \begin{subfigure}
    {0.30\linewidth}
    \includegraphics[width=1.0\textwidth]{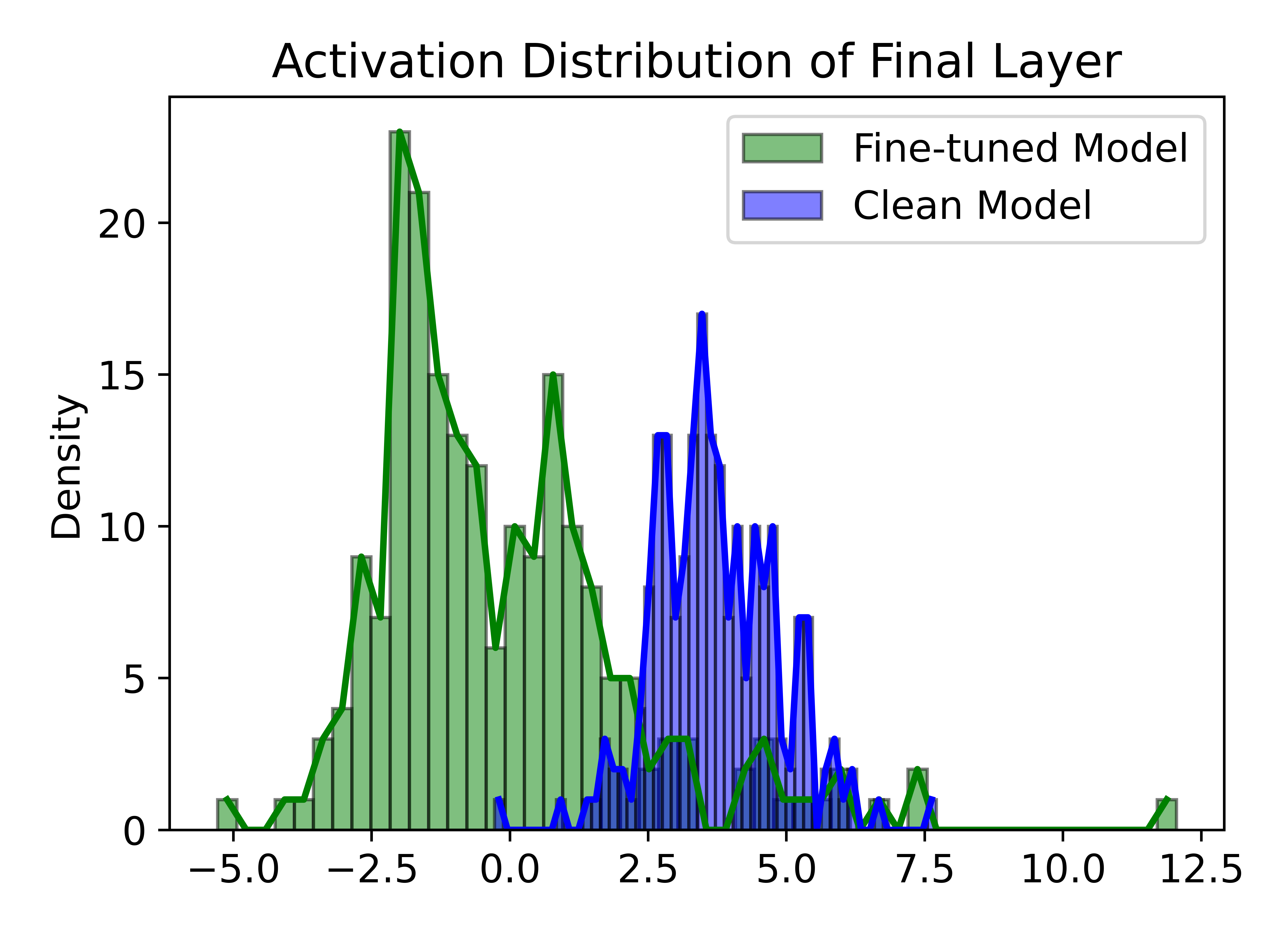}
    \caption{FT-Init}
    \end{subfigure}
    \begin{subfigure}
    {0.30\linewidth}
    \includegraphics[width=1.0\textwidth]{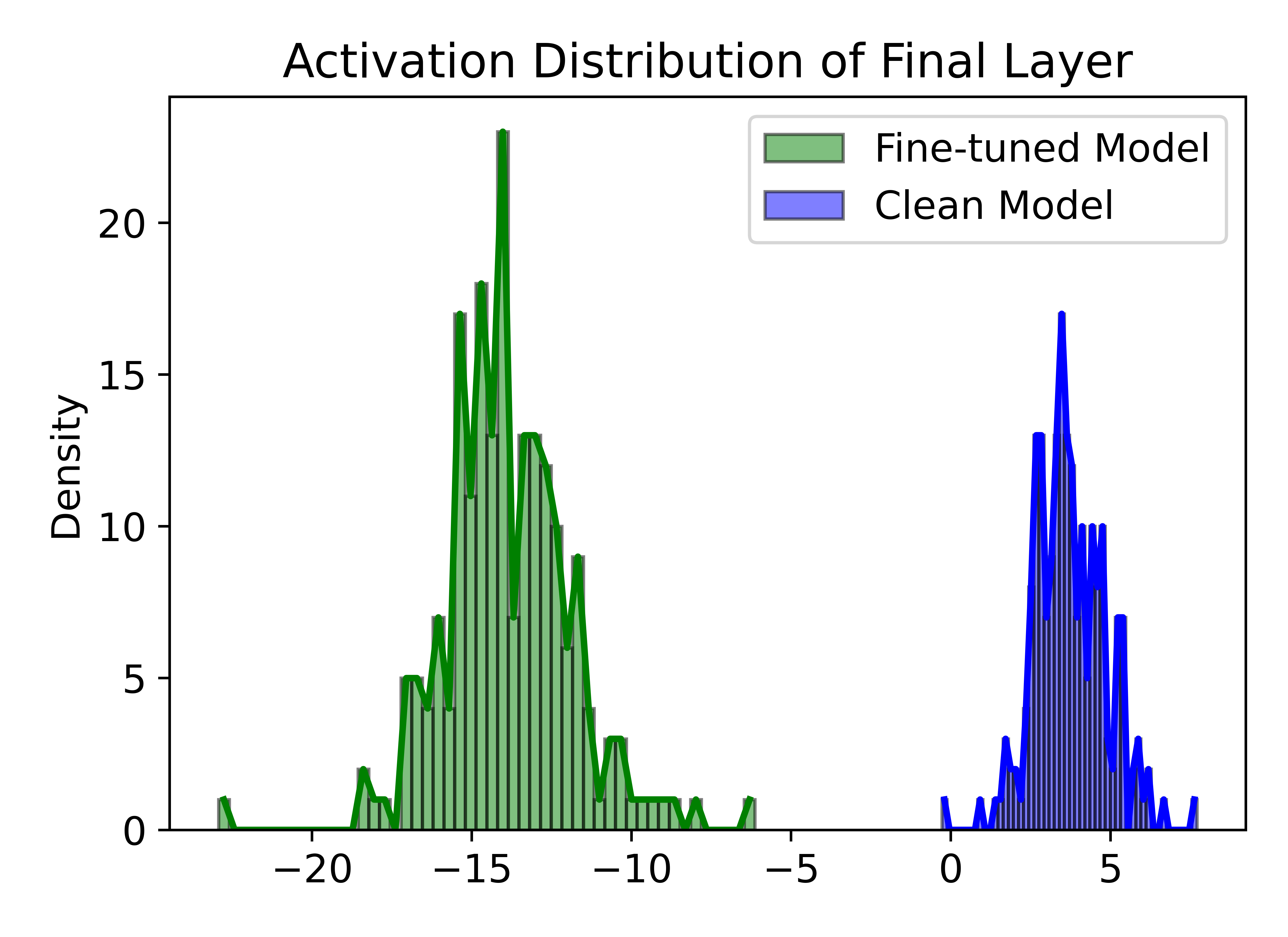}
    \caption{LP}
    \end{subfigure}
         \begin{subfigure}
    {0.30\linewidth}
    \includegraphics[width=1.0\textwidth]{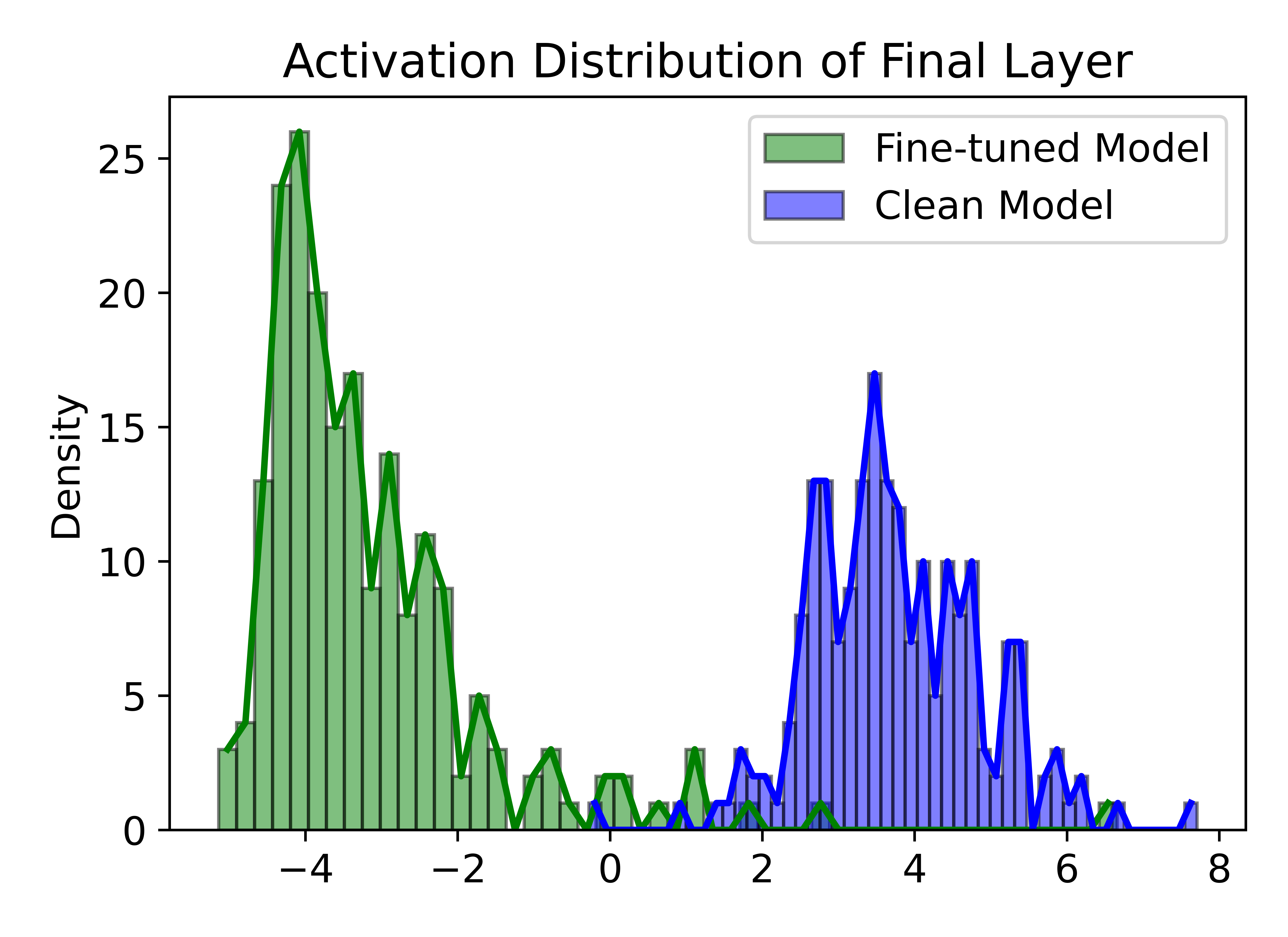}
    \caption{Fe-Tuning}
    \end{subfigure}
     \begin{subfigure}
    {0.30\linewidth}
    \includegraphics[width=1.0\textwidth]{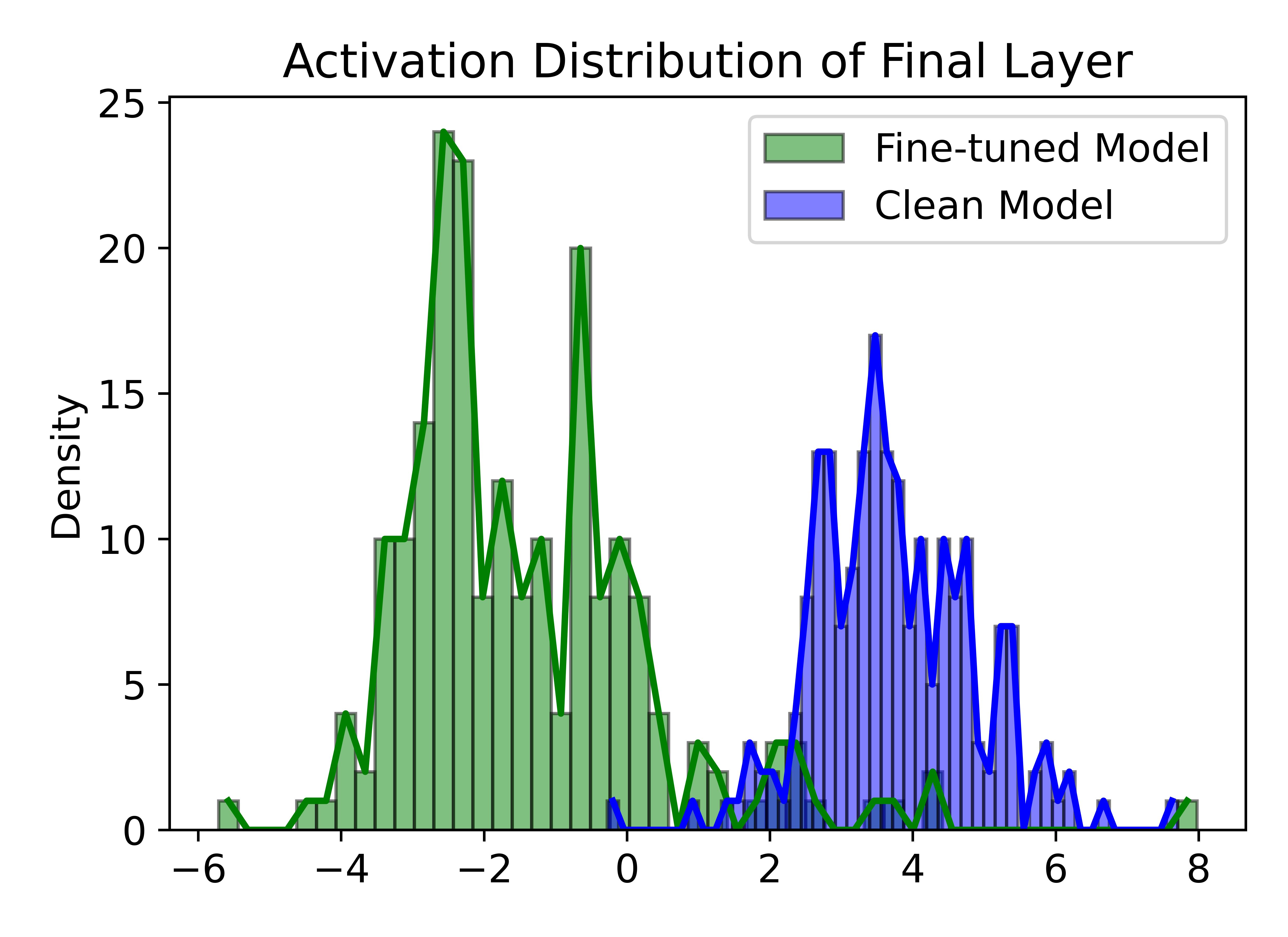}
    \caption{FST}
    \end{subfigure}
     \begin{subfigure}
    {0.30\linewidth}
    \includegraphics[width=1.0\textwidth]{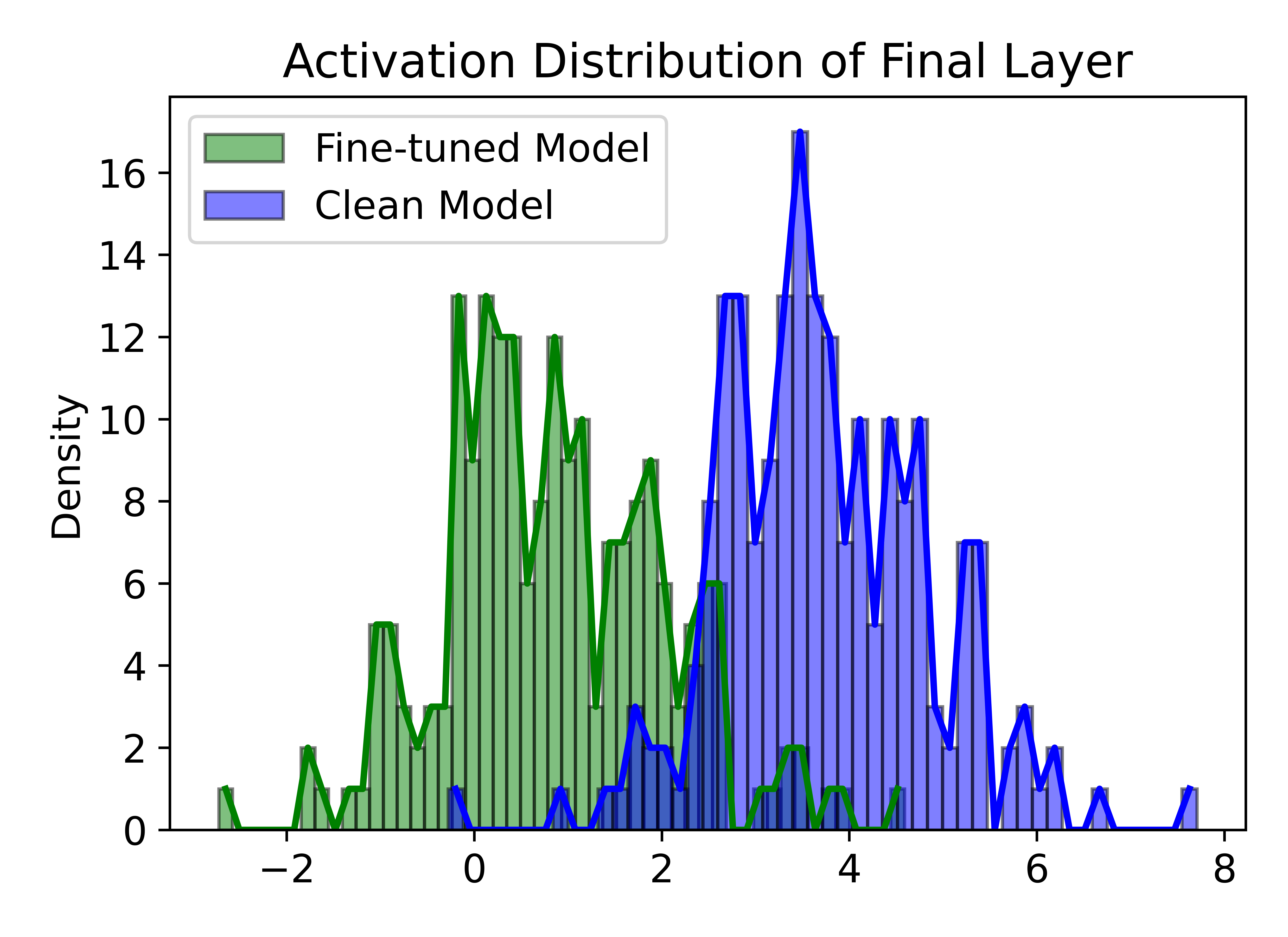}
    \caption{Ours}
    \end{subfigure}
    \caption{Comparison of model activation of different fine-tuning methods and a clean model on targeted malware samples on \textbf{AndroZoo} dataset.}
    \vspace{-2mm}
    \label{fig:decision-jigsaw}
\end{figure}
\begin{figure}[h]
    \centering
    \begin{subfigure}
    {0.30\linewidth}
    \includegraphics[width=1.0\textwidth]{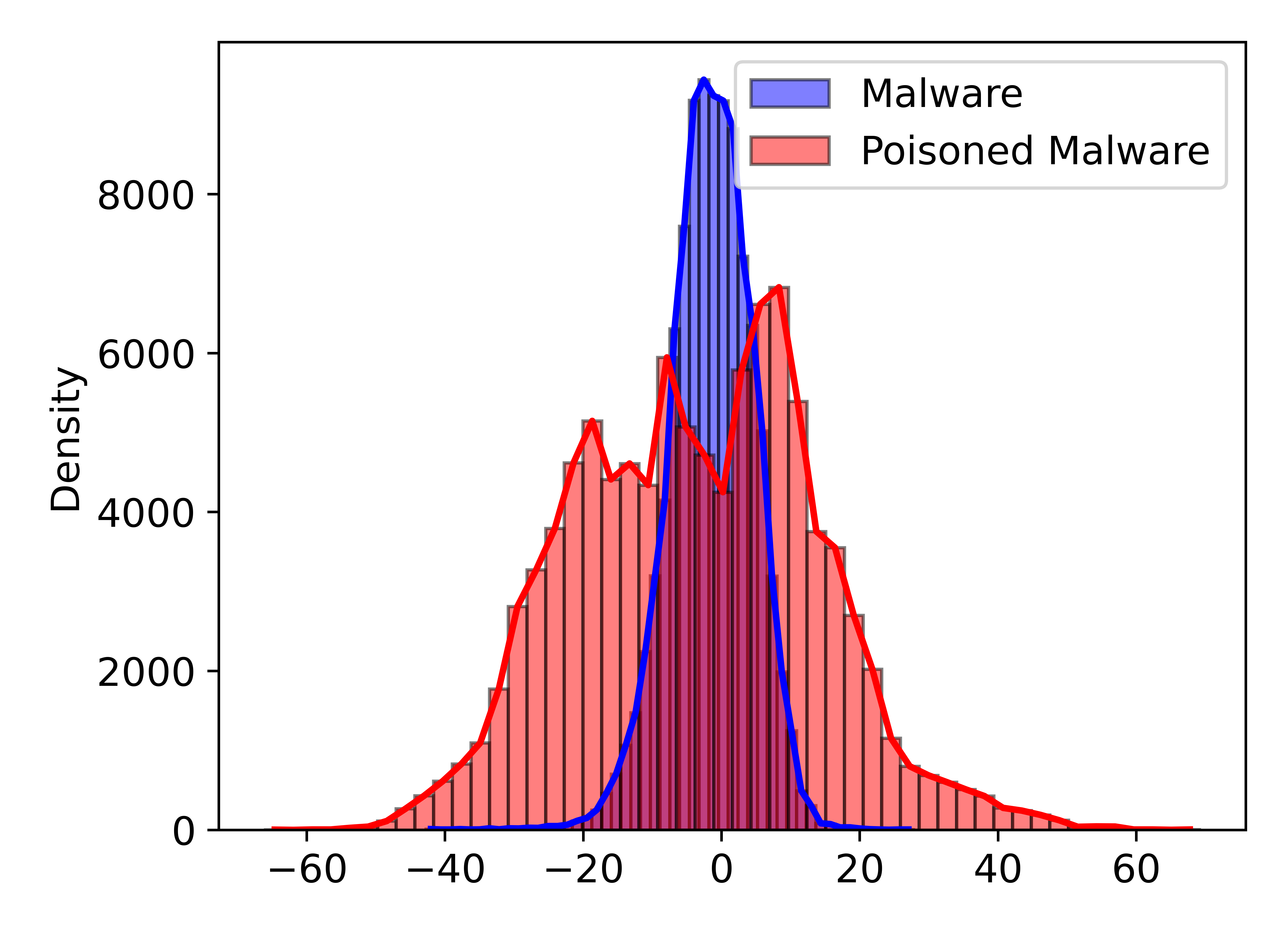}
    \caption{FT}
    \label{fig:illus-transfer}
    \end{subfigure}
    \begin{subfigure}
    {0.30\linewidth}
    \includegraphics[width=1.0\textwidth]{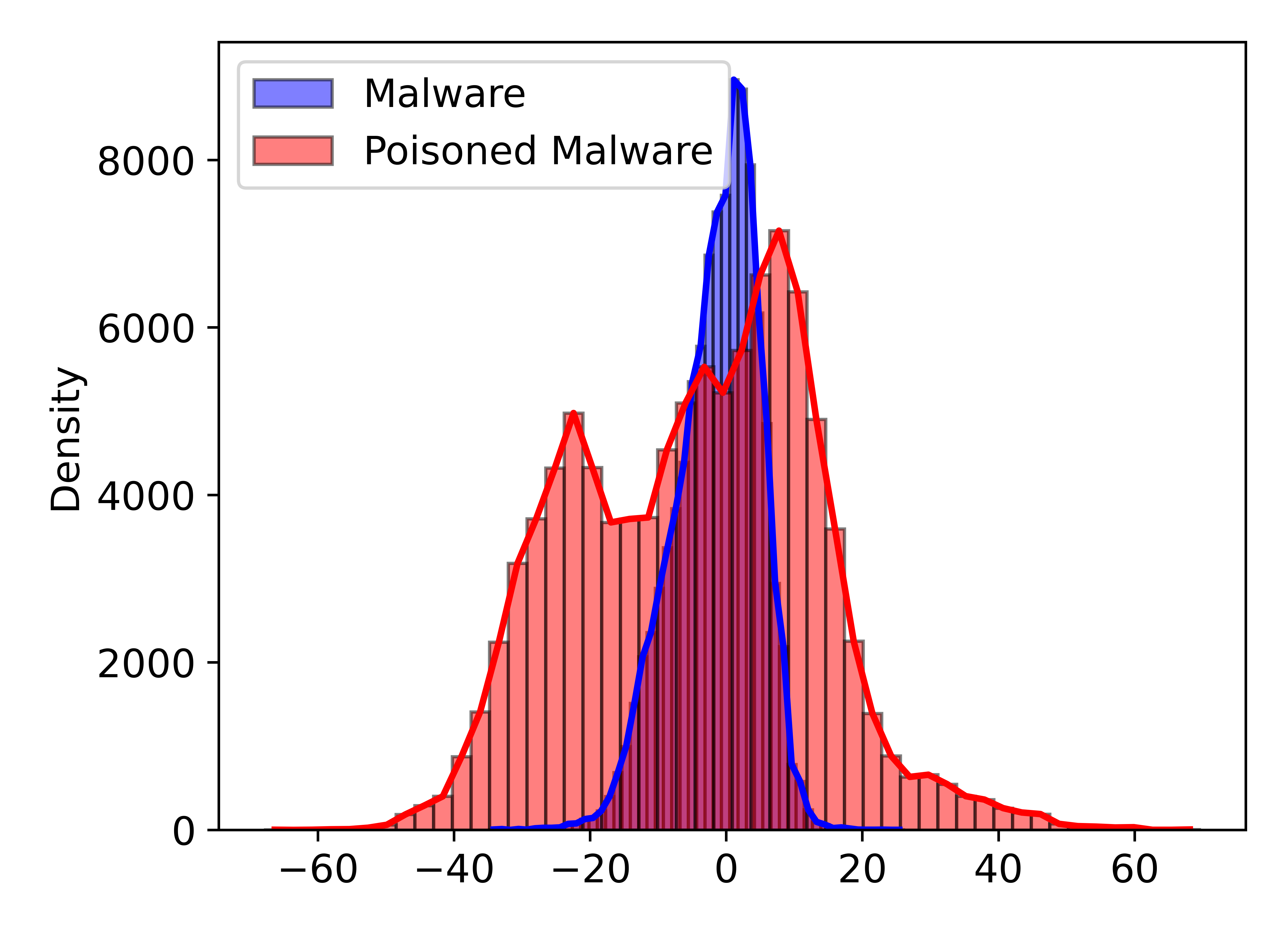}
    \caption{FT-Init}
    \label{fig:illus-transfer}
    \end{subfigure}
    \begin{subfigure}
    {0.30\linewidth}
    \includegraphics[width=1.0\textwidth]{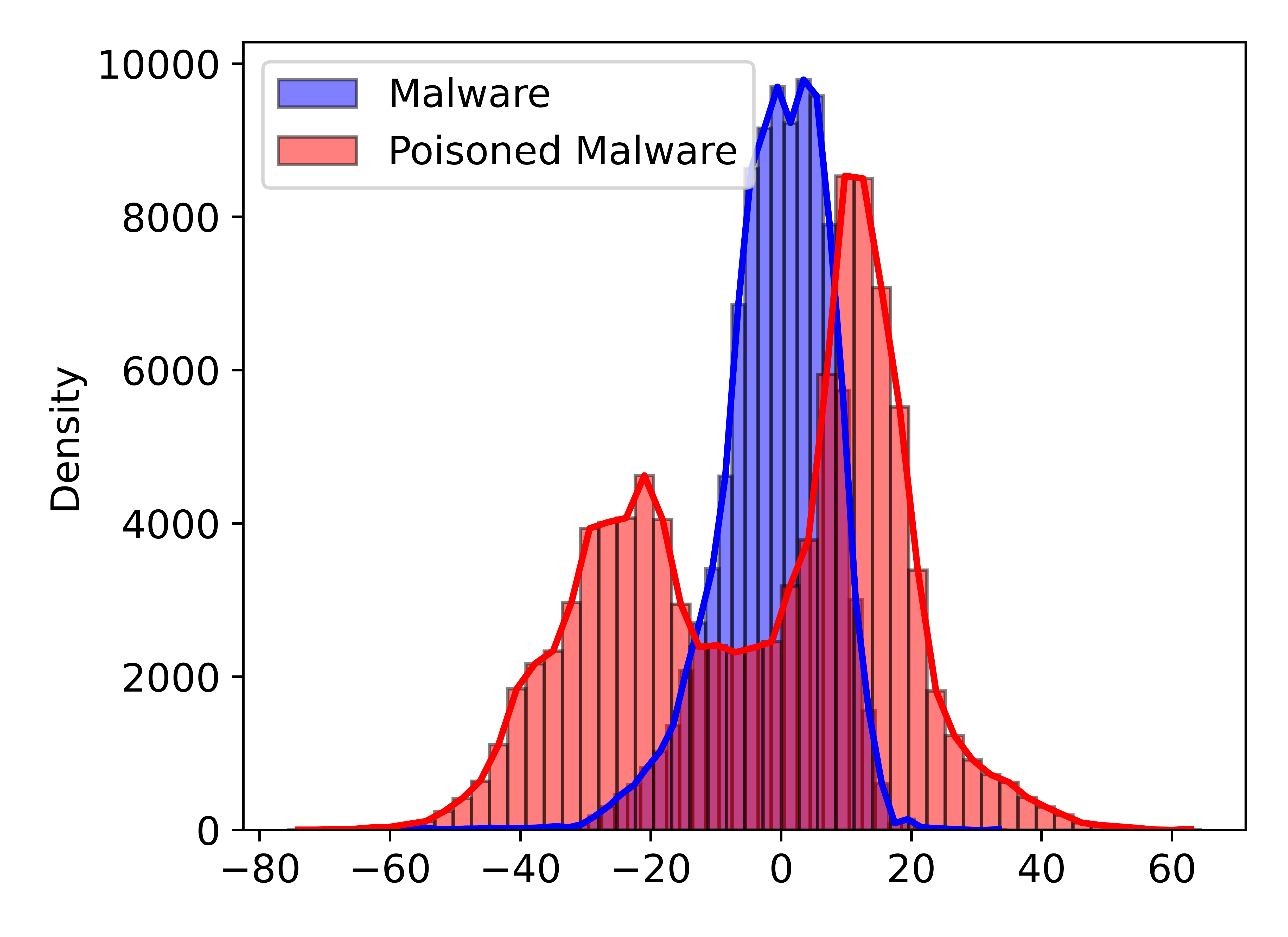}
    \caption{LP}
    \label{fig:illus-transfer}
    \end{subfigure}
         \begin{subfigure}
    {0.30\linewidth}
    \includegraphics[width=1.0\textwidth]{figs/ft_mode_mode_fe-tuning_sub_on_neurons.png}
    \caption{Fe-Tuning}
    \label{fig:illus-transfer}
    \end{subfigure}
     \begin{subfigure}
    {0.30\linewidth}
    \includegraphics[width=1.0\textwidth]{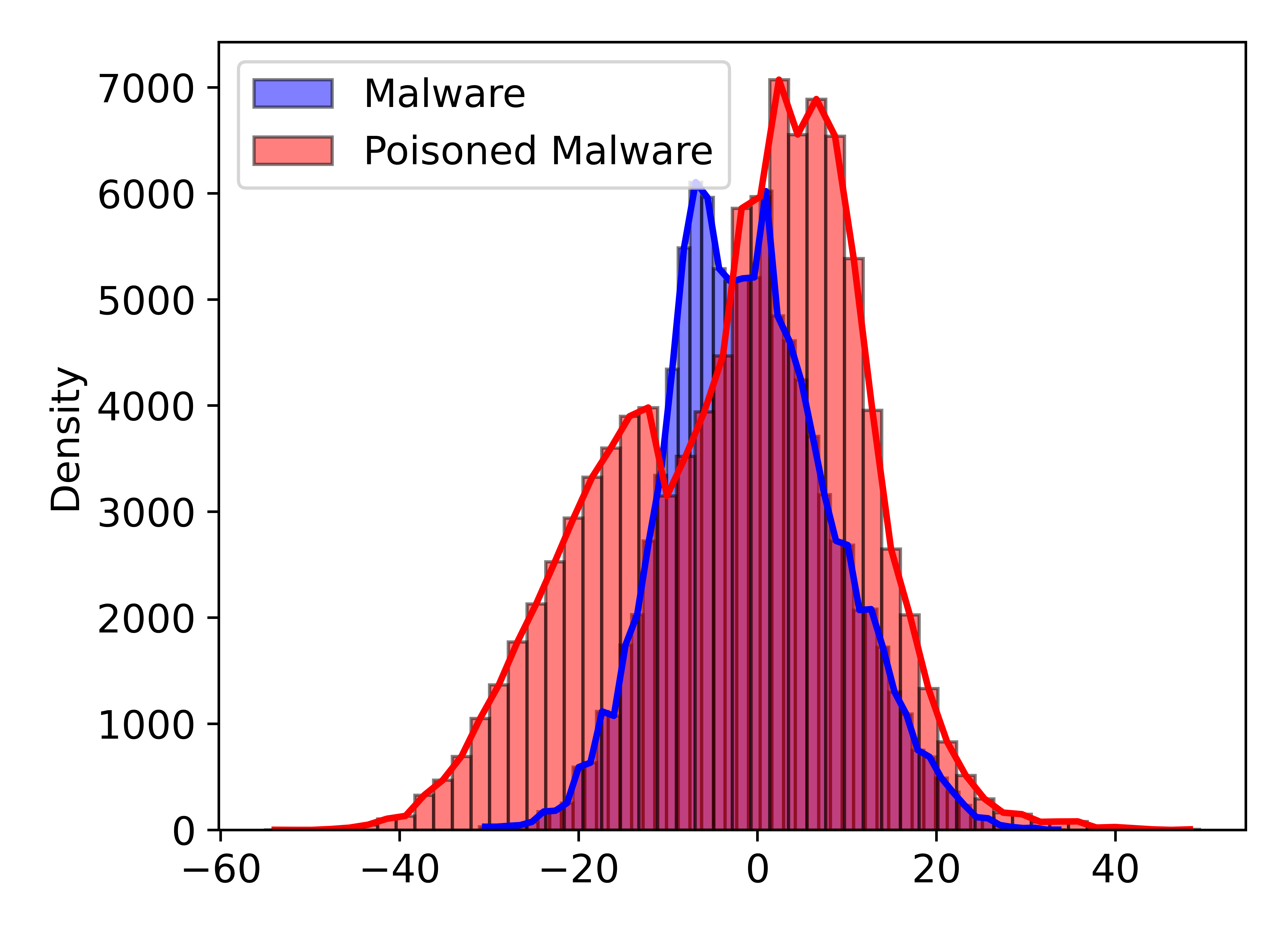}
    \caption{FST}
    \label{fig:illus-transfer}
    \end{subfigure}
     \begin{subfigure}
    {0.30\linewidth}
    \includegraphics[width=1.0\textwidth]{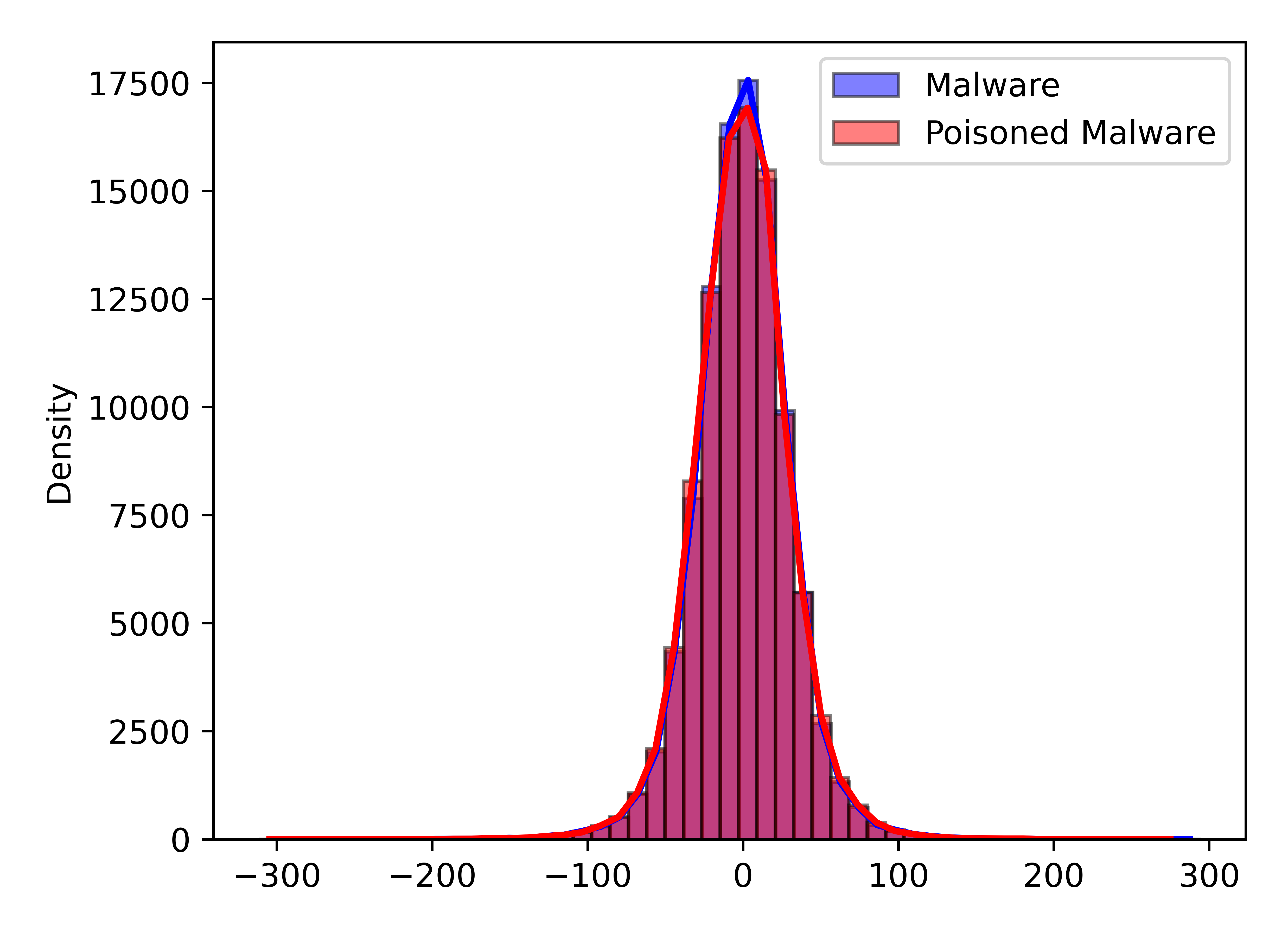}
    \caption{Ours}
    \label{fig:illus-transfer}
    \end{subfigure}
    \caption{Model activation of different fine-tuning methods on targeted malware samples with and without the trigger on \textbf{AndroZoo}
dataset on the final layer.}
 \label{fig:act-jigsaw-finetunes}
 \vspace{-8mm}
\end{figure}
\subsection{\added{Additional Analysis}}
\label{appendix:analysis}
\added{In this subsection, we further analyze the effect of the fine-tuning dataset on the backdoor purification efficacy. We consider the following factors: overlapping fraction, negative per positive class ratio, and number of malware families.}

\subsubsection{\added{Effect of overlapping fraction fine-tuning dataset}} 
\added{In this experiment, we study \method{} under different ways of constructing fine-tuning datasets, leading to three possible scenarios: (i) the training dataset is completely unavailable during fine-tuning, resulting in an overlapping fraction of 0 with the training data; (ii) partial overlap between the original training dataset and the fine-tuning dataset, where a certain positive portion of the samples is reused to guide fine-tuning; and (iii) complete overlap, where the entire fine-tuning data comes from the original training data.}

\added{We iteratively replace a fraction of the fine-tuning dataset with the corresponding number of samples from the training set, such that the fine-tuning dataset size is constant. The results of these scenarios are summarized in Table~\ref{tab:overlap-apg}.
We report three metrics for each scenario: Clean Accuracy (C-Acc), Attack Success Rate (ASR), and Defense Effectiveness Rating (DER). 
As the fine-tuning and original training datasets overlap increases, the results highlight that the performance on clean data (BA) remains relatively unchanged across most overlapping ratios. Interestingly, even with significant overlap (0.8 or 1.0), the ASR metric does not exceed 0.03, and the defense effectiveness rating is always greater than 98\%. The same observation is experienced with the EMBER dataset, and reusing training data can help boost defense effectiveness up to 3\%. The analysis reveals two key insights. 
First, to construct the fine-tuning dataset, we can reuse a part of the training data, as long as we can ensure that it does not affect negatively the performance on the benign task. Second, in most cases, using the original portion of the training data helps to achieve better performance of the backdoor purification task, because it can simulate the learning process and the relationship between benign and backdoored neurons during the mask generation process.}
\added{This finding suggests a practical approach to constructing fine-tuning datasets, allowing defenders to combine original training data with new fine-tuning samples for optimal results or addressing the difficulty in collecting malware samples.}

\input{tabs/ndss_r2/overlapping}

\subsubsection{\added{Effect of class ratio in the fine-tuning dataset}} 
\added{We analyze the performance of \method{} with different class ratios with both datasets \textbf{AndroZoo} and \textbf{EMBER}. Specifically, we modify the fine-tuning dataset to create different class ratios between positive and negative samples. First, we analyze the existing class distribution in the fine-tuning dataset and calculate the number of positive and negative samples required to meet the target ratio, i.e., $
\#\text{Positive\_Samples} = \frac{\text{class\_ratio}}{1 + \text{class\_ratio}} \times \text{total\_samples}$. If the fine-tuning dataset lacks sufficient samples, additional samples are randomly selected from the original training set, and a corresponding number of the other class is removed to ensure the fine-tuned data size is constant.}
\added{
The original class distributions of the two datasets differ significantly, influencing the achievable class ratios during training. In AndroZoo, the class ratio in the training data is approximately $0.089:1$, reflecting a strong imbalance. Therefore, we vary the ratio from lower to higher around the original training ratio to explore feasible ranges. In contrast, the EMBER dataset has a balanced ratio of $1:1$. For EMBER, we decrease the original ratio, as it is impractical for the number of malware samples to exceed the number of goodware samples in real-world scenarios.
This setting results in the following ranges: $[0.01, 0.04, 0.08, 0.10, 0.12, 0.15]$ for AndroZoo, centered around its original ratio, and $[0.10, 0.20, 0.40, 0.60, 0.80, 1.0]$ for EMBER, reflecting a controlled reduction from the balanced distribution.}
\added{
The experimental results are summarized in Table~\ref{tab:class-ratio}. Each row shows the performance of \method{} with different class ratios for both datasets.  AndroZoo, with its inherent imbalance, exhibits higher variance in MA and BA across ratios. For the extreme case, i.e., the negative samples account for less than 1\% of the fine-tuning dataset, the purification performance degrades by half. In the favorable case (i.e., the negative samples account for less than 10\% - 15\% of the fine-tuning dataset), the defender can collect more positive samples, resulting in a higher class ratio than the training ratio, which can indeed improve the performance of \method{} to 99.33\% defense effectiveness. On the other hand, with the EMBER dataset, since the original training was conducted on a balanced dataset, reducing the negative class ratio can reduce the fine-tuned model's performance.} 
\added{In conclusion, constructing a fine-tuned dataset with the close class ratio as in the training can help improve the performance of \method{}.}

\input{tabs/ndss_r2/class_ratio}
\subsubsection{Effect of family ratio in the fine-tuning dataset} 
\begin{figure}[h!]
\begin{minipage}{.42\linewidth}
\centering\includegraphics[width=0.8\linewidth]{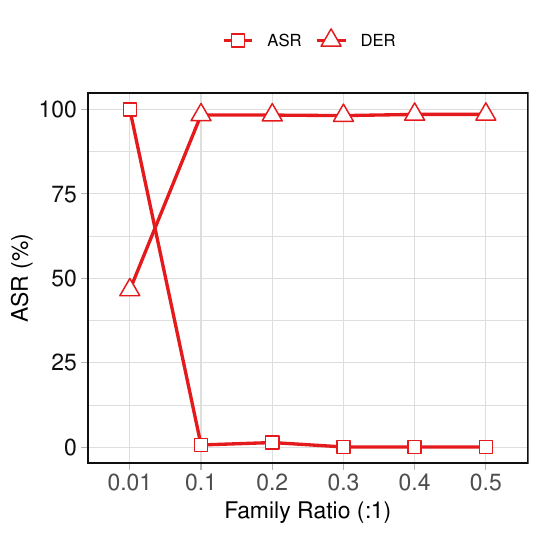}
    \caption{\added{ASR and DER of \method{} under varied family ratio with AndroZoo dataset.}}
\end{minipage}\hfill
\begin{minipage}{.52\linewidth}
    \input{tabs/ndss_r2/family_ratio}
\end{minipage}
\vspace{-7mm}
\end{figure}
\added{In this experiment, we study the effect of the number of families appearing in the fine-tuning datasets. This experiment is only applicable to the AndroZoo dataset since in the version EMBER dataset following implementation of~\cite{severi2021explanation}, the family information is not used and not available. AndroZoo has a total of 400 malware families and the training set has 323 families.}
\added{
We set up this experiment by gradually increasing the ratio of the number of families in the fine-tuning data per the number of families in the training data from $0.01:1$ to $0.5:1$. The default ratio of fine-tuning data in our setting is $0.30:1$. For a lower ratio, we randomly deleted a number of families to match the desired ratio. For a higher ratio, we randomly took 1, -- the median family size in the dataset, samples per additional family. If the family has less than 1 samples, all the samples from it will be added to the fine-tuning set. The result is shown in Table~\ref{tab:family-ratio}.} 
\added{The results indicate that \method{} struggles when the number of malware families is too small, such as in the setting of 3 families at a ratio of $0.01:1$. However, our method demonstrates effective and stable performance starting from a family ratio of $0.1:1$, corresponding to 30 malware families. Furthermore, some families contain only a single sample, which enhances the feasibility of collecting malware samples for the fine-tuning set in real-world applications.}

\subsubsection{Analysis of local effect of backdoor attacks}
\begin{figure}[h!]
    \centering
    \includegraphics[width=0.75\linewidth]{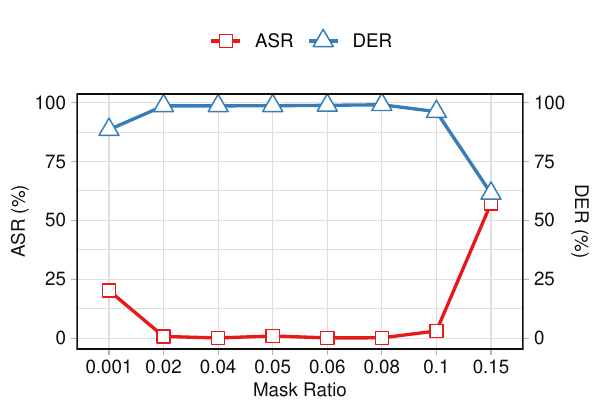}
    \caption{\added{Backdoor purification efficiency with increasing mask ratio $k$.}}
    \label{fig:mask-ratio}
\end{figure}

\added{In our experiments, the default mask ratio $k$ is set to $0.05$, which means top-$5$\% neurons will be considered as ``important'' for learning the dual objectives in Eqn.~\ref{eqn:align_loss}. The choice of 
$k$ is critical for achieving an optimal balance between backdoor mitigation and maintaining the model’s clean accuracy (C-Acc). If $k$ is set too high, it risks over-pruning neurons, reducing the model's ability to classify benign inputs accurately, thereby degrading C-Acc. Conversely, a small value of $k$ fails to sufficiently prune the backdoored neurons, leaving enough malicious pathways intact for the backdoor to remain effective, which increases the Attack Success Rate (ASR). To validate the effect of different mask ratios, we conducted experiments with $k$ values in the range $[0.001, 0.01, 0.02, 0.05, 0.1, 0.15]$. The results are presented in Fig.~\ref{fig:mask-ratio}. As presented, a small mask ratio (0.001) fails to remove the backdoor because it insufficiently prunes the backdoored neurons, leaving enough malicious pathways intact for the backdoor to remain effective. Conversely, a high mask ratio (0.15) negatively impacts both the DER and C-Acc by over-pruning, removing not only backdoored neurons but also essential neurons responsible for benign model behavior. This excessive pruning disrupts the model's ability to correctly classify benign inputs, resulting in degraded performance on both the backdoor and clean tasks.}

\added{The explanation for this local phenomenon is that backdoor attacks inherently require the adversary to strike a delicate balance between implanting a malicious trigger and preserving the network's overall benign performance. The adversary needs the model to maintain high performance on benign samples to avoid detection. As noted in recent works~\cite{gao2020backdoor,severi2021explanation}, if the backdoor were to affect the entire network, it would risk degrading the network's performance on benign samples, raising suspicion. Therefore, by design, the backdoor is typically restricted to a subset of neurons that respond only to specific triggers, while leaving the rest of the network largely unaffected. This localized impact confirms that backdoor attacks target specific neurons associated with adversarial triggers, rather than causing widespread disruption. Consequently, this sparsity allows the network to continue performing its benign task effectively, reinforcing the point that backdoor attacks do not result in global behavioral changes. This sparsity of backdoor effect is widely studied in related works~\cite{zhang2022neurotoxin,li2023reconstructive} and is part of a crucial insight into our approach to identifying and purifying these sparsely distributed neurons.}

\subsubsection{\added{Additional Results with CNN}}
\added{We incorporated an experiment with CNNs to strengthen our work, given that CNNs serve as fundamental building blocks in various architectures(e.g., ResNet/Inception/YOLO, etc.), and demonstrating PBP's effectiveness by reducing ASR from 99\% to 4\%. This performance is consistent across different adversary powers on PDR. 
In the following section with backdoor attacks on the CV domain, we further demonstrate that \method{} can work with more complicated structures such as ResNet and VGG.
}
\input{tabs/ndss_r2/CNN-model}

\added{The effectiveness of our method lies in three key aspects. First, by identifying backdoor neuron masks layer-by-layer, our approach ensures that the number of layers in the architecture does not affect performance, making it scalable to complex models. Second, we handle not only individual backdoor neurons but also their interactions with benign neurons through alternative optimization, allowing the method to adapt to varying inter-layer dependencies. Third, the method leverages the sparsity of backdoor neuron gradients, a phenomenon consistently observed across different architectures~\cite{zhang2022neurotoxin,li2024purifying}, ensuring robust backdoor mitigation without compromising clean accuracy.}

\subsection{\added{Broader Application}}
\label{appendix:cv-domain}
\begin{figure}[t!]
    \centering
    \begin{subfigure}{0.23\linewidth}
        \includegraphics[width=1.0\linewidth]{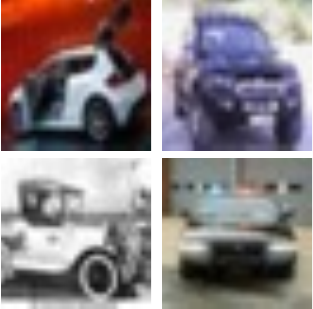}
        \caption{Original}
    \end{subfigure}\hfill
    \begin{subfigure}{0.23\linewidth}
        \includegraphics[width=1.0\linewidth]{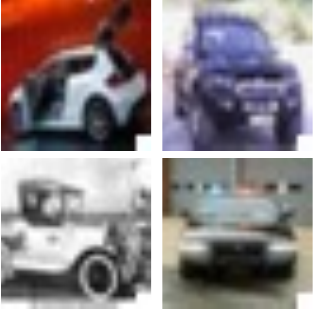}
        \caption{BadNet}
    \end{subfigure}\hfill
    \begin{subfigure}{0.23\linewidth}
        \includegraphics[width=1.0\linewidth]{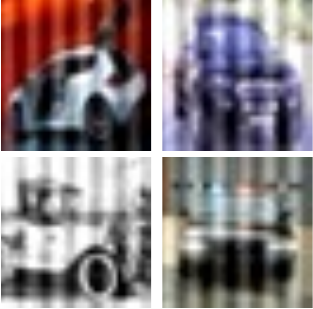}
        \caption{SIG}
    \end{subfigure}\hfill
    \begin{subfigure}{0.23\linewidth}
        \includegraphics[width=1.0\linewidth]{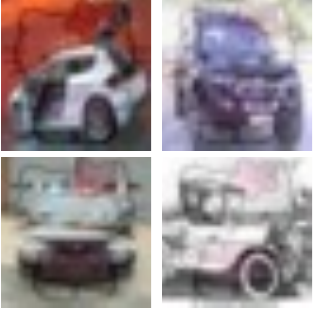}
        \caption{Blended}
    \end{subfigure}
    \caption{\added{Example images of backdoored samples from CIFAR-10 dataset with 3 attacks.}}
    \label{fig:cv-backdoor}
\end{figure}

\added{We extend the experiments on CV, which is beyond malware classification tasks to further demonstrate the potential efficacy of \method{}. We selected three mainstream backdoor attack methods including BadNet~\cite{gu2019badnets}, SIG~\cite{barni2019new}, and Blended~\cite{chen2017targeted}. The illustrative examples of these attack strategies are presented in Fig.~\ref{fig:cv-backdoor}. 
We selected FST as the baseline to compare since this method showed its SOTA performance on backdoor purification within the CV domain. 
The implementation is inherited from the benchmark provided by Min et al.~\cite{min2024towards}. For each attack strategy, we vary both the PDR, which corresponds to the adversary power, and model architecture (i.e., ResNet-18 and VGG19\_BN).}
\added{
From the results in Table~\ref{tab:cv-app-resnet} and Table~\ref{tab:cv-app-vgg}, \method{} achieves comparable or even better performance than FST in many settings. Specifically, \method{} successfully mitigates
ASR from 100\% to as low as under 5\%.}
\added{With VGG19\_BN model architecture in Table~\ref{tab:cv-app-vgg}, FST shows the difference in reducing ASR between BadNet and Blended attacks, i.e., ASR with 0.5\% PDR with BadNet is 2\% but the corresponding number with Blended is 28.19\%. The reason for this can be the different manners of these two attacks, where the BadNet adds a small pattern to the corner of images while Blended embeds the trigger to the whole image. As a result, shifting the model parameters far from the original one is not enough to address thoroughly the relationships between backdoored and benign neurons caused by these attacks. Meanwhile, \method{} can purify the backdoors better in this case, resulting in lower ASR.}
\added{Under ResNet-18 (Table~\ref{tab:cv-app-resnet}), FST shows a more stable performance, which aligns with the phenomenon observed in the original paper. As noted by the authors, one possible reason is that the classifier of VGG19-BN contains more than one layer which is slightly different from our previously used structure ResNet-18. In conclusion, our method is shown to be more stable with comparable performance in CV applications.}

\added{
The primary distinction between the backdoor purification task in CV and the malware classification problem lies in backdoor insertion and feature extraction. In CV applications, the backdoor is inserted in such a way that the model learns representations of backdoored samples in the hidden space, making them resemble the targeted class. Consequently, methods that shift the representation learning mechanism of the model, such as FST, tend to perform effectively.
Conversely, in malware classification, features of malware samples can be extracted using third-party tools, allowing adversaries to utilize these extracted features to craft poisoned samples. This approach differs significantly from the CV domain, where the goal is to manipulate the model’s representation of these features directly. Instead of forcing the model to adjust its internal representations for specific features, malware adversaries can leverage pre-extracted features to create malicious samples, highlighting the unique challenges in addressing backdoor attacks in this domain.}
\added{To this end, the performance of \method{} in an extended scenario demonstrates its potential application in various domains and its broader impacts.} 

\input{tabs/ndss_r2/cv_purification}
\input{tabs/ndss_r2/cv-purification-vgg}
\clearpage

%% file: tabs/training_params.tex
\begin{table}[tbh]
\centering
\caption{Training and Fine-tuning Configurations.}
\label{tab:main-params}
\resizebox{1.0\linewidth}{!}{
\begin{tabular}{@{}llccccc@{}}
\toprule
\multicolumn{2}{c}{Dataset}           & \begin{tabular}[c]{@{}c@{}}Train \\ batch size\end{tabular} & \begin{tabular}[c]{@{}c@{}}Test \\ batch size\end{tabular} & \begin{tabular}[c]{@{}c@{}}Optimizer\\ Learning rate\end{tabular} & Epochs & \begin{tabular}[c]{@{}c@{}}Model \\ architecture\end{tabular}  \\ \midrule
\multirow{2}{*}{EMBER}    & Train     & 512                                                         & 512                                                        & \begin{tabular}[c]{@{}c@{}}Adam\\ $0.001$\end{tabular}            & 5      & \begin{tabular}[c]{@{}c@{}}MLP\\ (4000/2000/1000)\end{tabular} \\
                          & Fine-tune & 512                                                         & 512                                                        & $0.004$                                                           & 10     & \begin{tabular}[c]{@{}c@{}}MLP\\ (4000/2000/1000)\end{tabular} \\ \cmidrule(l){2-7} 
\multirow{2}{*}{AndroZoo} & Train     & 256                                                         & 512                                                        & \begin{tabular}[c]{@{}c@{}}Adam\\ $0.001$\end{tabular}            & 20     & \begin{tabular}[c]{@{}c@{}}MLP\\ (4000/2000/1000)\end{tabular} \\
                          & Fine-tune & 256                                                         & 512                                                        & $0.001$                                                           & 10     & \begin{tabular}[c]{@{}c@{}}MLP\\ (4000/2000/1000)\end{tabular} \\ \bottomrule
\end{tabular}
}
\end{table}

%% file: tabs/model_acc.tex
\begin{table}
\centering
\caption{Backdoored model evaluation before fine-tuning on both datasets.}
\label{tab:model-acc}
\resizebox{0.98\columnwidth}{!}{
\begin{tabular}{@{}rlccccc@{}}
\toprule
Model                     & \begin{tabular}[c]{@{}l@{}}\replaced{PDR}{Poisoning \\ Rate}\end{tabular} & Accuracy & ASR     & F1-Score & Precision & Recall  \\ \midrule
\multicolumn{7}{l}{EMBER}                                                                                                                   \\ \cmidrule(lr){1-7}
\multirow{4}{*}{\rotatebox{90}{Backdoor}} & 0.005                                                     & 98.94\%  & 97.72\% & 98.94\%  & 99.04\%   & 98.85\% \\
                          & 0.01                                                      & 98.98\%  & 98.93\% & 98.97\%  & 99.05\%   & 98.89\% \\
                          & 0.02                                                      & 99.07\%  & 99.55\% & 99.07\%  & 99.20\%   & 98.93\% \\
                          & 0.05                                                      & 98.97\%  & 99.67\% & 98.97\%  & 99.12\%   & 98.82\% \\
\multicolumn{2}{l}{Clean}                                                             & 98.99\%  & 70.19\% & 98.98\%  & 99.18\%   & 98.79\% \\ \cmidrule(lr){1-7}
\multicolumn{7}{l}{AndroZoo}                                                                                                                \\ \cmidrule(lr){1-7}
\multirow{4}{*}{\rotatebox{90}{Backdoor}} & 0.005                                                     & 98.53\%  & 82.91\% & 90.72\%  & 92.20\%   & 89.29\% \\
                          & 0.01                                                      & 98.48\%  & 99.90\% & 90.14\%  & 94.35\%   & 86.28\% \\
                          & 0.02                                                      & 98.58\%  & 99.45\% & 90.97\%  & 92.86\%   & 89.16\% \\
                          & 0.05                                                      & 98.59\%  & 99.72\% & 90.90\%  & 94.29\%   & 87.74\% \\
\multicolumn{2}{l}{Clean}                                                             & 98.52\%  & 0.01\%  & 90.61\%  & 92.78\%   & 88.54\% \\ \bottomrule
\end{tabular}}
\end{table}

%% file: tabs/ft_jigsaw_0.01.tex
\begin{table}[h]
\centering
\caption{Comparison of different methods under fine-tuning size = 0.01 for AndroZoo dataset.}
\label{tab:jigsaw-full-0.01}
\resizebox{0.98\linewidth}{!}{
\begin{tabular}{@{}rl|ccccccc@{}}
\toprule
PDR         & Metrics & Pretrained & FT    & FT-Init & FE-Tuning & LP    & FST   & Ours  \\ \midrule
\multirow{2}{*}{0.005} & C-Acc   & 98.53      & 98.59 & 97.73   & 84.14     & 84.14 & 98.55 & 91.29 \\
                       & ASR     & 82.91      & 79.88 & 69.52   & 29.96     & 29.96 & 51.98 & 2.90  \\ \cmidrule{1-9}
\multirow{2}{*}{0.01}  & C-Acc   & 98.61      & 98.65 & 97.81   & 72.14     & 98.61 & 98.66 & 92.85 \\
                       & ASR     & 86.57      & 83.81 & 72.14   & 81.88     & 85.85 & 88.37 & 0.73  \\ \cmidrule{1-9}
\multirow{2}{*}{0.02}  & C-Acc   & 98.59      & 98.65 & 98.22   & 83.12     & 98.59 & 98.58 & 94.03 \\
                       & ASR     & 99.00      & 99.41 & 47.22   & 1.84      & 99.38 & 69.45 & 4.00  \\ \cmidrule{1-9}
\multirow{2}{*}{0.05}  & C-Acc   & 98.69      & 98.65 & 97.82   & 77.66     & 98.63 & 98.62 & 95.74 \\
                       & ASR     & 99.00      & 100.0 & 99.83   & 97.62     & 100.0 & 100.0 & 0.79  \\ \bottomrule
\end{tabular}
}
\end{table}

%% file: tabs/ft_jigsaw_0.02.tex
\begin{table}[h]
\centering
\caption{Comparison of different methods under fine-tuning size = 0.02 for AndroZoo dataset.}
\label{tab:jigsaw-full-0.02}
\resizebox{0.98\linewidth}{!}{
\begin{tabular}{@{}rl|ccccccc@{}}
\toprule
PDR         & Metrics & Pretrained & FT    & FT-Init & FE-Tuning & LP    & FST   & Ours  \\ \midrule
\multirow{2}{*}{0.005} & C-Acc   & 98.67      & 98.65 & 98.67   & 97.83     & 98.69 & 98.68 & 94.57 \\
                       & ASR     & 96.34      & 98.65 & 98.67   & 97.83     & 98.69 & 98.68 & 18.40 \\ \cmidrule{1-9}
\multirow{2}{*}{0.01}  & C-Acc   & 98.56      & 98.62 & 98.62   & 98.07     & 98.59 & 98.70 & 93.55 \\
                       & ASR     & 99.90      & 99.93 & 99.90   & 19.43     & 99.93 & 99.93 & 7.21  \\ \cmidrule{1-9}
\multirow{2}{*}{0.02}  & C-Acc   & 98.67      & 98.61 & 98.58   & 97.84     & 98.67 & 98.65 & 94.45 \\
                       & ASR     & 97.62      & 98.14 & 35.28   & 0.82      & 98.86 & 71.83 & 0.21  \\ \cmidrule{1-9}
\multirow{2}{*}{0.05}  & C-Acc   & 98.68      & 98.60 & 98.49   & 97.87     & 98.60 & 98.67 & 96.51 \\
                       & ASR     & 99.97      & 100.0 & 0.00    & 0.00      & 100.0 & 100.0 & 0.73  \\ \bottomrule
\end{tabular}
}
\end{table}

%% file: tabs/ft_jigsaw_0.05.tex
\begin{table}[hbt!]
\centering
\caption{Comparison of different methods under fine-tuning size = 0.05 for AndroZoo dataset.}
\label{tab:jigsaw-full-0.05}
\resizebox{0.98\linewidth}{!}{
\begin{tabular}{@{}rl|ccccccc@{}}
\toprule
PDR        & Metrics & Pretrained & FT    & FT-Init & FE-Tuning & LP    & FST   & Ours  \\ \midrule
\multirow{2}{*}{0.005} & C-Acc   & 98.62      & 98.66 & 98.62   & 98.57     & 98.59 & 98.64 & 96.51 \\ 
                       & ASR     & 85.40      & 85.36 & 87.68   & 80.08     & 82.40 & 91.09 & 10.25 \\ \cmidrule{1-9}
\multirow{2}{*}{0.01}  & C-Acc   & 98.62      & 98.65 & 98.67   & 98.66     & 98.61 & 98.70 & 96.32 \\ 
                       & ASR     & 100.0      & 100.0 & 96.65   & 99.31     & 100.0 & 99.93 & 13.77 \\ \cmidrule{1-9}
\multirow{2}{*}{0.02}  & C-Acc   & 98.48      & 98.67 & 98.64   & 98.55     & 98.56 & 98.66 & 96.29 \\
                       & ASR     & 80.42      & 80.81 & 89.75   & 69.04     & 94.58 & 0.31  & 14.88 \\ \cmidrule{1-9}
\multirow{2}{*}{0.05}  & C-Acc   & 98.69      & 98.63 & 98.61   & 98.51     & 98.62 & 98.58 & 96.36 \\
                       & ASR     & 99.79      & 100.0 & 0.00    & 0.14      & 100.0 & 95.72 & 8.21  \\ \bottomrule
\end{tabular}}
\end{table}

%% file: tabs/ndss_r2/overlapping.tex
\begin{table}[t]
\centering
\caption{\added{\method{}'s efficacy with different overlapping ratios of the fine-tuning dataset with the original training dataset.}}
\label{tab:overlap-apg}
\resizebox{1.0\linewidth}{!}{
\begin{tabular}{@{}c|rrc|ccc@{}}
\toprule
\multicolumn{1}{l}{\multirow{2}{*}{\makecell{Overlapping \\Fraction}}} & \multicolumn{3}{c}{AndroZoo}                    & \multicolumn{3}{c}{EMBER} \\ \cmidrule(lr){2-4}\cmidrule(lr){5-7} 
\multicolumn{1}{l}{}                                   & \multicolumn{1}{c}{C-Acc ($\uparrow$)} & \multicolumn{1}{c}{ASR ($\downarrow$)} & DER ($\uparrow$) & C-Acc ($\uparrow$)         & ASR  ($\downarrow$)   & DER ($\uparrow$)     \\ \midrule
0.0                                                    
& 96.86                  & 0.89   & 98.55                
& 96.41       & 17.58  & 89.64     \\
0.2                                                    
& 96.79                  & 0.03     & 98.95              
& 96.32       & 17.42  & 89.67     \\
0.4                                                    
& 94.98                  & 0.03     & 98.04              
& 96.14       & 12.86    & 91.86   \\
0.6                                                    
& 94.55                  & 0.03      & 97.83             
& 96.44       & 15.20    & 92.12   \\
0.8                                                    
& 96.42                  & 0.03    & 98.76               
& 96.44       & 15.84    & 90.52   \\
1.0                                                    
& 95.92                  & 0.03      & 98.51             
& 96.47       & 14.47    & 91.12   \\ \cmidrule{1-7}
\makecell{Backdoored}                                       
& 98.59                  & 99.72      & --            
& 98.99       & 99.43 & --\\  
\bottomrule
\end{tabular}}
\end{table}

%% file: tabs/ndss_r2/class_ratio.tex
\begin{table}[h!]
\centering
\caption{\added{\method{}'s efficacy with different positive per negative class ratios with both datasets.}}
\label{tab:class-ratio}
\resizebox{1.0\linewidth}{!}{
\begin{tabular}{@{}cccc|cccc@{}}
\toprule
\multicolumn{1}{c}{\multirow{2}{*}{\makecell{Class \\Ratio}}} & \multicolumn{3}{c|}{AndroZoo}                         & \multicolumn{1}{c}{\multirow{2}{*}{\makecell{Class \\Ratio}}} & \multicolumn{3}{c}{EMBER} \\ \cmidrule(lr){2-4} \cmidrule(l){6-8} 
\multicolumn{1}{l}{}                             & \multicolumn{1}{c}{C-Acc ($\uparrow$)}    & \multicolumn{1}{c}{ASR ($\downarrow$)}  & DER ($\uparrow$) & \multicolumn{1}{l}{}                             
& C-Acc ($\uparrow$)          
& ASR ($\downarrow$)
& DER ($\uparrow$)
\\ \midrule
0.01                                             & 96.12                     & 49.15       & 74.04             
& 0.10                                              
& 83.21       & 35.02   & 74.32     \\
0.04                                             & 96.92 & 0.14 & 98.96
& 0.20                                              & 94.02       & 21.31      & 86.58 \\
0.08                                             & 96.86                 & 0.89                & 98.55         
& 0.40                                              & 95.81       & 25.92      & 85.17 \\
0.10                                             & 96.90                      & 0.27         & 98.88             
& 0.60                                              & 95.87       & 29.03      & 85.20 \\
0.12                                             & 97.53                     & 0.00              & 99.16         
& 0.80                                              & 96.93       & 20.79      & 88.29 \\
0.15                                             & 97.26                     & 0.07              & 99.33       
& 1.00                                              & 96.41       & 17.58       & 89.64 \\ \midrule
Backdoored                                       & 98.59                     & 99.72        & --            & Backdoored                                       & 98.99       & 99.43  & --     \\ \bottomrule
\end{tabular}}
\end{table}

%% file: tabs/ndss_r2/family_ratio.tex
\captionof{table}{\added{\method{}'s efficacy with different fine-tuning family ratios with AndroZoo dataset.}}
\label{tab:family-ratio}
\resizebox{1.0\linewidth}{!}{
\begin{tabular}{@{}crrr@{}}
\toprule
\multicolumn{1}{l}{Family Ratio} & \multicolumn{1}{c}{C-Acc ($\uparrow$)} & \multicolumn{1}{c}{ASR ($\downarrow$)} & \multicolumn{1}{c}{DER ($\uparrow$)} \\ \midrule
0.01                             & 91.95                  & 100.0    & 46.54                \\
0.10                             & 96.27                  & 0.62     & 98.39              \\
0.20                             & 96.92                  & 1.34     & 98.36              \\
0.30                             & 95.31                  & 0.03     & 98.21              \\
0.40                             & 95.98                  & 0.00 & 98.56                      \\
0.50                             & 95.99                  & 0.00 & 98.56                      \\ \midrule
Backdoored                       & 98.59                  & 99.72    & --              \\ \bottomrule
\end{tabular}}

%% file: tabs/ndss_r2/CNN-model.tex
\begin{table}[tbh]
\centering
\caption{\added{Performance of PBP under CNN model with varied PDR.}}
\begin{tabular}{@{}lcccccccc@{}}
\toprule
\multirow{2}{*}{Models} & \multicolumn{2}{c}{0.005} & \multicolumn{2}{c}{0.01} & \multicolumn{2}{c}{0.02} & \multicolumn{2}{c}{0.05} \\ \cmidrule(lr){2-3} \cmidrule(lr){4-5} \cmidrule(lr){6-7}  \cmidrule(lr){8-9} 
                        & C-Acc       & ASR         & C-Acc       & ASR        & C-Acc       & ASR        & C-Acc       & ASR        \\ \cmidrule{1-9}
Pretrained              & 99.07       & 66.79       & 99.24       & 94.06      & 99.11       & 97.74      & 99.12       & 99.97      \\
Ours                    & 97.84       & 1.59        & 97.87       & 4.70       & 97.76       & 3.45       & 97.55       & 4.00       \\ \bottomrule
\end{tabular}
\end{table}

%% file: tabs/ndss_r2/cv_purification.tex
\begin{table}[tbh!]
\centering
\caption{\added{Defense results under various poisoning rates. The experiments are conducted on the CIFAR-10 dataset
with ResNet-18. All the metrics are measured in percentage (\%).}}
\label{tab:cv-app-resnet}
\resizebox{0.98\linewidth}{!}{
\begin{tabular}{@{}cl|rr|rr|rr@{}}
\toprule
\multicolumn{1}{l}{\multirow{2}{*}{PDR}} & \multirow{2}{*}{Model}         & \multicolumn{2}{c}{BadNet}                          & \multicolumn{2}{c}{SIG}                             & \multicolumn{2}{c}{Blended}                         \\ \cmidrule(lr){3-4} \cmidrule(lr){5-6} \cmidrule(lr){7-8} 
\multicolumn{1}{l}{}                     &                                & \multicolumn{1}{l}{C-Acc} & \multicolumn{1}{l}{ASR} & \multicolumn{1}{l}{C-Acc} & \multicolumn{1}{l}{ASR} & \multicolumn{1}{l}{C-Acc} & \multicolumn{1}{l}{ASR} \\ \midrule
\multirow{3}{*}{0.005}                   & \multicolumn{1}{c}{No-defense} & 94.84                     & 92.25                   & 94.79                     & 88.38                   & 94.58                     & 94.88                   \\
                                         & FST                            & 92.47                     & 0.56                    & 92.26                     & 1.99                    & 91.77                     & 1.56                    \\
                                         & PBP                            & 92.23                     & 2.17                    & 91.94                     & 0.07                    & 91.37                     & 3.50                     \\ \midrule
\multirow{3}{*}{0.01}                    & \multicolumn{1}{c}{No-defense} & 94.42                     & 92.78                   & 94.56                     & 88.69                   & 94.86                     & 97.88                   \\
                                         & FST                            & 92.05                     & 1.04                    & 92.82                     & 0.23                    & 92.24                     & 5.48                    \\
                                         & PBP                            & 91.31                     & 1.73                    & 92.10                     & 5.88                    & 92.00                     & 3.37                    \\ \midrule
\multirow{3}{*}{0.02}                    & \multicolumn{1}{c}{No-defense} & 94.20                     & 94.00                   & 94.63                     & 92.97                   & 94.25                     & 98.72                   \\
                                         & FST                            & 91.61                     & 0.74                    & 92.29                     & 0.72                    & 91.83                     & 4.08                    \\
                                         & PBP                            & 91.15                     & 1.10                    & 92.00                     & 8.48                    & 91.25                     & 4.90                    \\ \midrule
\multirow{3}{*}{0.05}                    & \multicolumn{1}{c}{No-defense} & 93.80                     & 96.62                   & 94.46                     & 98.29                   & 94.85                     & 99.67                   \\
                                         & FST                            & 91.25                     & 1.23                    & 92.16                     & 0.07                    & 92.38                     & 11.19                   \\
                                         & PBP                            & 91.59                     & 1.17                    & 91.42                     & 0.50                     & 91.64                     & 4.71                    \\ \bottomrule
\end{tabular}}
\end{table}

%% file: tabs/ndss_r2/cv-purification-vgg.tex
\begin{table}[h!]
\centering
\caption{\added{Defense results under various poisoning rates. The experiments are conducted on the CIFAR-10 dataset
with VGG19\_BN. All the metrics are measured in percentage (\%).}}
\label{tab:cv-app-vgg}
\resizebox{0.98\linewidth}{!}{
\begin{tabular}{@{}cl|rr|rr|rr@{}}
\toprule
\multicolumn{1}{l}{\multirow{2}{*}{PDR}} & \multirow{2}{*}{Model}         & \multicolumn{2}{c}{BadNet}                          & \multicolumn{2}{c}{SIG}                             & \multicolumn{2}{c}{Blended}                         \\ \cmidrule(lr){3-4} \cmidrule(lr){5-6} \cmidrule(lr){7-8} 
\multicolumn{1}{l}{}                     &                                & \multicolumn{1}{l}{C-Acc} & \multicolumn{1}{l}{ASR} & \multicolumn{1}{l}{C-Acc} & \multicolumn{1}{l}{ASR} & \multicolumn{1}{l}{C-Acc} & \multicolumn{1}{l}{ASR} \\ \midrule
\multirow{3}{*}{0.005}                   & \multicolumn{1}{c}{No-defense} & 93.22                     & 83.89                   & 92.23                     & 76.95                   & 92.62                     & 97.89                   \\
                                         & FST                            & 88.49                     & 2.02                    & 87.29                     & 17.14                   & 88.79                     & 28.19                   \\
                                         & PBP                            & 88.97                     & 2.44                    & 86.47                     & 0.82                    & 87.25                     & 10.32                   \\ \midrule
\multirow{3}{*}{0.01}                     & \multicolumn{1}{c}{No-defense} & 93.17                     & 87.12                   & 91.47                     & 80.48                   & 92.35                     & 95.47                   \\
                                         & FST                            & 89.04                     & 1.53                    & 87.01                     & 13.12                   & 88.67                     & 29.10                    \\
                                         & PBP                            & 88.90                      & 2.00                       & 86.27                     & 4.02                    & 88.70                      & 9.40                     \\ \midrule
\multirow{3}{*}{0.02}                    & \multicolumn{1}{c}{No-defense} & 92.51                     & 90.39                   & 91.68                     & 88.60                    & 93.07                     & 98.54                   \\
         & FST                            & 88.23                     & 2.13                    & 87.00                        & 6.18                    & 88.94                     & 24.75                   \\
         & PBP                            & 89.26                     & 2.41                    & 86.11                     & 1.83                    & 88.73                     & 5.21                                  \\ \midrule
\multirow{3}{*}{0.05}                    
& \multicolumn{1}{c}{No-defense} 
& 92.52                     & 94.30                    & 93.20                      & 93.77                   & 93.11                     & 99.44                   \\
 & FST                            & 89.10                      & 2.61                    & 88.65                     & 8.73                    & 89.81                     & 23.99                   \\
 & PBP                            & 88.51                     & 3.03                    & 87.40                      & 0.65                    & 89.63                     & 4.63                    \\ 
                                         \bottomrule
\end{tabular}}
\end{table}